%% file: main.tex
\documentclass[review]{elsarticle}

\input{includes/preamble_packages.tex}

\input{includes/packages.tex}

\input{includes/preamble_symbols.tex}

\input{includes/shortcuts.tex}

\journal{Artificial Intelligence Journal}

\begin{document}

\begin{frontmatter}

\title{Monitoring of Perception Systems: \\ 
Deterministic, Probabilistic, and Learning-based \\ Fault Detection and Identification\tnoteref{t1}}

\author[mit]{Pasquale Antonante\corref{corrauthor}}
\cortext[corrauthor]{Corresponding author}
\ead{antonap@mit.edu}

\author[mit]{Heath Nilsen}
\ead{hnilsen@mit.edu}

\author[mit]{Luca Carlone}
\ead{lcarlone@mit.edu}

\address[mit]{Massachusetts Institute of Technology, 77 Massachusetts Ave, Cambridge, MA 02139}

\tnotetext[t1]{This work was partially funded by the NSF CAREER award “Certifiable Perception for Autonomous Cyber-Physical Systems”}

\begin{abstract}
\input{sections/abstract.tex}
\end{abstract}


\end{frontmatter}

\input{sections/introduction.tex}

\input{sections/related_work}
\input{sections/problem_formulation.tex}
	\input{sections/subsections/problem_formulation.tex}
	\input{sections/subsections/fault_detection_vs_identification.tex}

\input{sections/approach_formulation}
	\input{sections/subsections/diagnostic_tests.tex}
	\input{sections/subsections/diagnostic_graph.tex}
\input{sections/algos_for_identification.tex}
	\input{sections/subsections/inference_on_deterministic.tex}

\input{sections/subsections/inference_on_factor_graphs.tex}

	\input{sections/subsections/graph_neural_networks.tex}
\input{sections/foundamental_limits.tex}
	\input{sections/subsections/deterministic_diagnosability}
	\input{sections/subsections/probabilistic_diagnosability}
\input{sections/experiments.tex}

\input{sections/conclusions.tex}



\bibliography{references/refs.bib, references/myRefs.bib}

\end{document}

%% file: includes/preamble_packages.tex
\usepackage{comment}
\usepackage[binary-units]{siunitx}
\usepackage{relsize}
\usepackage{ifthen}
\usepackage[colorinlistoftodos]{todonotes}

\usepackage[caption=false]{subfig}

\usepackage[vlined,ruled,linesnumbered]{algorithm2e}
\usepackage{graphics} 
\usepackage{rotating}
\usepackage{color}
\usepackage{enumerate}
\usepackage[T1]{fontenc}
\usepackage{psfrag}
\usepackage{epsfig} 
\usepackage{booktabs}
\usepackage{graphicx,url}
\usepackage{multirow}
\usepackage{array}
\usepackage{latexsym}
\usepackage{amsfonts}
\usepackage{amsmath}
\usepackage{amssymb}
\usepackage{xstring}
\usepackage[noend]{algorithmic}
\usepackage{multirow}
\usepackage{xcolor}
\usepackage{prettyref}
\usepackage{flexisym}
\usepackage{bigdelim}
\usepackage{breqn} 
\usepackage{listings}

\usepackage{enumitem}
\usepackage{xspace}
\usepackage{bm}
\graphicspath{{./figures/}}
\usepackage{tikz}
\usetikzlibrary{matrix,calc}

\usepackage{mdwlist}

\makecompactlist{itemize}{stditemize}


%% file: includes/packages.tex
\usepackage{hyperref}
\hypersetup{
    linktoc=all,
}
\usepackage[export]{adjustbox}  
\usepackage{soul}
\usepackage{amsthm} 
\usepackage{amsmath}
\usepackage{nicefrac}
\usepackage[capitalize]{cleveref}
\usepackage{dsfont}
\usepackage{longtable}
\usepackage{multirow}

\usepackage{tabularx}

\usepackage{floatrow}
\usepackage{blindtext}
\usepackage{makecell}

%% file: includes/preamble_symbols.tex
\newrefformat{prob}{Problem\,\ref{#1}}
\newrefformat{def}{Definition\,\ref{#1}}
\newrefformat{sec}{Section\,\ref{#1}}
\newrefformat{sub}{Section\,\ref{#1}}
\newrefformat{prop}{Proposition\,\ref{#1}}
\newrefformat{app}{Appendix\,\ref{#1}}
\newrefformat{alg}{Algorithm\,\ref{#1}}
\newrefformat{cor}{Corollary\,\ref{#1}}
\newrefformat{thm}{Theorem\,\ref{#1}}
\newrefformat{lem}{Lemma\,\ref{#1}}
\newrefformat{fig}{Fig.\,\ref{#1}}
\newrefformat{tab}{Table\,\ref{#1}}

\newtheorem{theorem}{Theorem}

\newtheorem{corollary}[theorem]{Corollary}

\newtheorem{lemma}[theorem]{Lemma}

\newtheorem{definition}[theorem]{Definition}

\newtheorem{remark}[theorem]{Remark}

\newcommand{\cf}{\emph{cf.}\xspace}

\newcommand{\bdmath}{\begin{dmath}}
\newcommand{\edmath}{\end{dmath}}
\newcommand{\beq}{\begin{equation}}
\newcommand{\eeq}{\end{equation}}
\newcommand{\bdm}{\begin{displaymath}}
\newcommand{\edm}{\end{displaymath}}
\newcommand{\bea}{\begin{eqnarray}}
\newcommand{\eea}{\end{eqnarray}}
\newcommand{\beal}{\beq \begin{array}{ll}}
\newcommand{\eeal}{\end{array} \eeq}
\newcommand{\beas}{\begin{eqnarray*}}
\newcommand{\eeas}{\end{eqnarray*}}
\newcommand{\ba}{\begin{array}}
\newcommand{\ea}{\end{array}}
\newcommand{\bit}{\begin{itemize}}
\newcommand{\eit}{\end{itemize}}
\newcommand{\ben}{\begin{enumerate}}
\newcommand{\een}{\end{enumerate}}

\newcommand{\calA}{{\cal A}}
\newcommand{\calB}{{\cal B}}
\newcommand{\calC}{{\cal C}}
\newcommand{\calD}{{\cal D}}
\newcommand{\calE}{{\cal E}}
\newcommand{\calF}{{\cal F}}

\newcommand{\calL}{{\cal L}}
\newcommand{\calM}{{\cal M}}
\newcommand{\calN}{{\cal N}}
\newcommand{\calO}{{\cal O}}

\newcommand{\calR}{{\cal R}}
\newcommand{\calS}{{\cal S}}
\newcommand{\calT}{{\cal T}}

\newcommand{\calV}{{\cal V}}
\newcommand{\calW}{{\cal W}}
\newcommand{\calX}{{\cal X}}
\newcommand{\calY}{{\cal Y}}

\newcommand{\etal}{\emph{et~al.}\xspace}
\newcommand{\setal}{~\emph{et~al.}\xspace}
\newcommand{\eg}{\emph{e.g.,}\xspace}
\newcommand{\ie}{\emph{i.e.,}\xspace}
\newcommand{\myParagraph}[1]{{\bf #1.}\xspace}

\newcommand{\M}[1]{{\bm #1}} 
\renewcommand{\boldsymbol}[1]{{\bm #1}}

\newcommand{\hide}[1]{}

\newcommand{\hiddenText}{{\color{gray} hidden text.}}
\newcommand{\hideWithText}[1]{\hiddenText}

\newcommand{\Natural}[1]{ { {\mathbb N}^{#1} } }

\DeclareMathOperator*{\argmax}{arg\,max}

\newcommand{\tran}{^{\mathsf{T}}}

\newcommand{\eye}{{\mathbf I}}

\newcommand{\opt}{^{\star}}

\newcommand{\at}[1]{^{(#1)}}

\newcommand{\setdef}[2]{ \{#1 \; {:} \; #2 \} }

\newcommand{\MA}{\M{A}}

\newcommand{\MD}{\M{D}}
\newcommand{\ME}{\M{E}}

\newcommand{\MP}{\M{P}}

\newcommand{\MW}{\M{W}}

\newcommand{\va}{\boldsymbol{a}}

\newcommand{\ve}{\boldsymbol{e}}
\newcommand{\vf}{\boldsymbol{f}}

\newcommand{\vxx}{\boldsymbol{x}}

\newcommand{\vzz}{\boldsymbol{z}}

\newcommand{\blue}[1]{{\color{blue}#1}}

\newcommand{\linkToPdf}[1]{\href{#1}{\blue{(pdf)}}}
\newcommand{\linkToPpt}[1]{\href{#1}{\blue{(ppt)}}}
\newcommand{\linkToCode}[1]{\href{#1}{\blue{(code)}}}
\newcommand{\linkToWeb}[1]{\href{#1}{\blue{(web)}}}
\newcommand{\linkToVideo}[1]{\href{#1}{\blue{(video)}}}
\newcommand{\linkToMedia}[1]{\href{#1}{\blue{(media)}}}
\newcommand{\award}[1]{\xspace} 

\newcommand{\pos}{\boldsymbol{p}}

%% file: includes/shortcuts.tex
\input{includes/redeclare_utils.tex} 

\newcommand{\subscript}[2]{$#1 _ #2$} 
\newlist{assumptions}{enumerate}{10}
\setlist[assumptions]{label=(\subscript{A}{{\arabic*}})}

\newcommand{\module}{module\xspace}
\newcommand{\modules}{{\module}s\xspace}
\newcommand{\dtest}{diagnostic test\xspace}
\newcommand{\dtests}{diagnostic tests\xspace}
\newcommand{\DTests}{Diagnostic Tests\xspace}
\newcommand{\dgraph}{diagnostic graph\xspace}
\newcommand{\DGraph}{Diagnostic Graph\xspace}
\newcommand{\dgraphs}{diagnostic graphs\xspace}
\newcommand{\DGraphs}{Diagnostic Graphs\xspace}

\newcommand{\fmode}{failure mode\xspace}
\newcommand{\fmodes}{{\fmode}s\xspace}
\newcommand{\pass}{PASS\xspace} 
\newcommand{\fail}{FAIL\xspace}
\newcommand{\mathpass}{\mathrm{\pass}} 
\newcommand{\mathfail}{\mathrm{\fail}}
\newcommand{\factive}{ACTIVE\xspace} 
\newcommand{\finactive}{INACTIVE\xspace}
\newcommand{\mathfactive}{\mathrm{\factive}} 

\newcommand{\ltrue}{\mathrm{TRUE}} 
\newcommand{\btrue}{1} 
\newcommand{\bfalse}{0} 
\newcommand{\dkappa}{\kappa} 
\newcommand{\dgamma}{\gamma} 
\newcommand{\dprob}{p} 
\newcommand{\kdiagnosability}{$\dkappa$-diagnosability\xspace}
\newcommand{\kdiagnosable}{$\dkappa$-diagnosable\xspace}
\newcommand{\sys}{\calS} 
\newcommand{\dsys}{\calD} 

\newcommand{\moduleset}{\calM} 
\newcommand{\modulevar}{m} 
\newcommand{\outputset}{\calO} 
\newcommand{\outputvar}{o} 
\newcommand{\failureset}{\{1,\ldots,\nrfmode\}} 
\newcommand{\failure}{f} 
\newcommand{\testset}{\{1,\ldots,\nrtests\}} 
\newcommand{\test}{t} 
\newcommand{\dataset}{\calW}

\newcommand{\faults}{\vf} 
\newcommand{\faultsover}[1]{\faults_{#1}} 
\newcommand{\syndrome}{\vzz} 
\newcommand{\syndromevar}{z} 
\newcommand{\faultsdistr}{\mathcal{F}} 

\newcommand{\varset}{\calV} 
\newcommand{\relset}{\calR} 
\newcommand{\rel}{\factorvar} 
\newcommand{\relsetTest}{\relset_{\mathrm{test}}} 
\newcommand{\relsetPrior}{\relset_{\mathrm{prior}}} 
\newcommand{\relation}{relation\xspace}
\newcommand{\relations}{{\relation}s\xspace}
\newcommand{\edgeset}{\calE}

\newcommand{\nrfmode}{{N_f}}
\newcommand{\nrtests}{{N_t}}

\newcommand{\probDetect}[1]{p_{d,#1}}
\newcommand{\probFalseAlarm}[1]{p_{a,#1}}
\newcommand{\activeFailureModesSet}{\calA}

\newcommand{\marginal}{\mu}
\newcommand{\partitionfcn}{Z} 
\newcommand{\fgvarset}{\calV} 
\newcommand{\fgedgeset}{\calE} 
\newcommand{\factorset}{\Phi} 
\newcommand{\factorvar}{\phi} 

\DeclareMathOperator{\scope}{scope} 
\DeclareMathOperator{\ind}{\mathds{1}} 
\DeclareMathOperator{\syngen}{syndrome}
\RedeclareMathOperator{\pos}{pos} 
\DeclareMathOperator{\cls}{cls} 
\DeclareMathOperator{\neighbors}{\calN}
\DeclareMathOperator{\Ex}{\mathbb{E}} 

\DeclareMathOperator{\readout}{\mathrm{READOUT}} 

\newcommand{\Reals}[1]{ { {\mathbb R}^{#1} } }

\DeclareMathOperator{\update}{update}
\DeclareMathOperator{\aggregate}{aggregate}

\newcommand{\loss}{\calL}

\newcommand{\sst}{\;\mid\;} 

\newcommand{\MAP}{MAP\xspace}
\newcommand{\Scope}{\mathbb{S}}
\newcommand{\sqat}[1]{^{[#1]}}
\newcommand{\blank}{{\cdot}}

\newcommand{\fst}[1]{{%
  \colorlet{foo}{green!33}%
  \sethlcolor{foo}\hl{#1}%
}}
\newcommand{\snd}[1]{{%
  \colorlet{foo}{yellow!80}%
  \sethlcolor{foo}\hl{#1}%
}}

\newcommand{\lobs}{\calA}
\newcommand{\robs}{\calB}
\newcommand{\matches}{\calC}
\newcommand{\lgsvl}{LGSVL\xspace}

\newcommand{\omitted}[1]{}

\newcommand{\WeakerOR}{Weaker-OR\xspace} 

\pgfmathtruncatemacro\distance{1}
\newfloatcommand{capbtabbox}{table}[][\FBwidth]

\definecolor{FaultColor}{rgb}{0.94,0.25,0.15}
\definecolor{FaultFreeColor}{rgb}{0.43, 0.76, 0.46}

\newcommand{\repourl}{https://github.com/MIT-SPARK/PerceptionMonitoring}
\newcommand{\videourl}{https://www.mit.edu/~antonap/videos/AIJ22PerceptionMonitoring.mp4}

%% file: sections/abstract.tex

This paper investigates runtime monitoring of perception systems.
Perception is a critical component of high-integrity applications of robotics and autonomous systems, such as self-driving cars. 
In these applications, failure of perception systems may put human life at risk, and a broad adoption of these technologies requires the development of methodologies to guarantee and monitor safe operation.
Despite the paramount importance of perception, currently there is no formal approach for system-level perception monitoring. 
In this paper, we  formalize the problem of runtime fault detection and identification in perception systems and 
 present a framework to model diagnostic information using a \emph{\dgraph}. 
We then provide a set of deterministic, probabilistic, and learning-based algorithms that use \dgraphs to perform fault detection and identification.  
Moreover, we investigate fundamental limits and provide deterministic and probabilistic guarantees 
on the fault detection and identification results. 
We conclude the paper with an extensive experimental evaluation, which recreates several realistic failure modes in 
the \lgsvl open-source autonomous driving simulator, and applies the proposed system monitors to a state-of-the-art autonomous driving software stack 
(Baidu's Apollo Auto). The results show that the proposed system monitors outperform baselines, have the potential of preventing accidents in realistic autonomous driving scenarios, and incur a negligible computational overhead.

%% file: sections/introduction.tex

\section{Introduction}


The number of Autonomous Vehicles (AVs) on our roads is increasing rapidly, with major players in the space already offering autonomous rides to the public~\cite{BusinessInsider22-WaytmoToSF}.
Self-driving cars promise a deep transformation of personal mobility and have the potential to improve safety, 
efficiency (\eg commute time, fuel), and induce a paradigm shift in how entire cities are designed~\cite{Silberg12wp-selfDriving}.
One key factor that drives the adoption of such technology is the capability of ensuring 
and monitoring safe operation. 
Consider Uber's fatal self-driving crash~\cite{ntsbuber} in 2018: the report from the National Transportation Safety Board states that ``inadequate safety culture'' contributed to the fatal collision between the autonomous vehicle and the pedestrian. 
In a recent survey~\cite{aaa2022}, the American Automobile Association (AAA) reports that vehicles with autonomous driving features consistently failed to avoid crashes with other cars or bicycles.
An analysis by Business Insider~\cite{BusinessInsider22-AVIncresedCollisions} found that the number of accidents 
involving AVs surged in 2021. 
This is a clear sign that the industry needs a sound methodology, embedded in the design process, to guarantee 
safety and build public trust.

Safe operation requires AVs to correctly understand their surroundings, in order to avoid unsafe behaviors.
In particular, AVs rely on onboard \emph{perception systems} to provide situation awareness and inform the onboard decision-making and control systems.
The perception system uses sensor data and prior knowledge (\eg high-definition maps) to create an internal representation of the surrounding environment, 
 including estimates for the positions and velocities of other vehicles and pedestrians, or the presence of traffic signs and traffic lights. 
Modern perception systems use both data-driven and classical methods. 
While classical methods are well-rooted in signal processing and estimation theory and have been extensively studied in robotics and computer vision, 
they may still have unexpected failure modes in practice,
 \eg local convergence in the Iterative Closest Point for 3D object pose estimation~\cite{Yang20arxiv-teaser} or 
 premature termination of robust estimation techniques as RANSAC~\cite{Fischler81}, among many other examples.
The use of data-driven methods further exacerbates the problem of ensuring correctness of the perception outputs, since 
current neural network architectures are  still prone to creating
 unexpected and often unpredictable failure modes~\cite{Salay17arxiv-mlISO26262}.

Ensuring and monitoring the correct operation of the perception system of an AV is a major challenge.
Industry heavily relies on simulation and testing to provide evidence of safety.
Although there is an increasing interest in the area of safety certification and runtime monitoring, the literature lacks a system-level framework to organize and reason over the diagnostic information available at runtime for the purpose of detecting and identifying potential perception-system failures.
Reliable runtime monitoring would enable the vehicle to have a better understanding of the conditions it operates in, and would give it enough notice to take adequate actions to preserve safety (\ie switch to fail-safe mode or hand over the control to a human operator) in case of severe failures. 
In this paper, we use the term ``failure'' (or ``fault'') in a general sense, to also denote failures of the \emph{intended functionality}~\cite{iso21448, AptivSafetyAV}.
For instance, a neural network can execute correctly (\eg without errors in the implementation or in the hardware running the network) but can still fail to produce a correct prediction for out-of-distribution inputs. 
Then, \emph{fault detection} is the problem of detecting 
the \emph{presence} of a fault in the system, while \emph{fault identification} is the problem of inferring which 
components of the system are faulty. The latter is particularly important since (i) not every fault has the same severity, hence understanding which component is failing may lead to different responses, (ii) a designer can use 
fault statistics to decide to focus research and development efforts on certain components, and (iii) a regulator can use information about specific faults to trace the steps or even determine responsibilities after an accident. 

Most of the existing literature (which we review more extensively in \cref{sec:relatedWork}) has focused on detecting failures of specific modules or specific algorithms, like localization~\cite{Jing22tits-GPSintegrity, Hafez20ral-integrityMonitoring}, semantic segmentation~\cite{Besnier21cvf-triggeringFailures}, or obstacle detection~\cite{Besnier21cvf-triggeringFailures}.
These methodologies often use a white-box approach (the monitor knows how the monitored algorithm works to some extent), 
and are sometimes computationally expensive to run~\cite{Besnier21cvf-triggeringFailures}. 
However, the literature still lacks a framework for system-level monitoring of perception systems, which is able to detect and identify failures in complex systems involving both classical and data-driven (possibly asynchronous, multi-modal)\footnote{Modern perception systems rely on data from multiple sensors and are implemented 
in multi-threaded architectures, where each algorithm may be executed at a different rate. 
} perception algorithms. 

\myParagraph{Contribution} This paper addresses this gap and provides methodologies for runtime monitoring 
(in particular, fault detection and identification) of complex perception systems.
Our first contribution is to formalize the problem (\cref{sec:problem_formulation}) and to present a framework (\cref{sec:approach_formulation}) to organize heterogeneous diagnostic tests 
of a perception system into a graphical model, the \emph{\dgraph}. 
In particular, we present different mathematical models (including both deterministic and probabilistic models) to describe common \dtests. Then, we introduce the concept of \dgraph, 
and extend it to capture asynchronous information over time (leading to \emph{temporal} \dgraphs).
 Our framework adopts a black-box approach, in that it remains agnostic to the inner workings of the perception algorithms, and only focuses on collecting results from \dtests that check the validity of their outputs.

Our second contribution (\cref{sec:algorithms}) is a set of algorithms that use \dgraphs to perform fault detection and identification. 
For the deterministic case, we provide optimization-based methods that find the smallest set of faults that explain the test results. For the probabilistic case, we transform a \dgraph into a factor graph and perform inference to find the set of faulty modules. Finally, we propose a learning-based approach  based on graph neural networks that learns to predict failures in a  \dgraph.

Our third contribution (\cref{sec:foundamental_limits}) is to investigate fundamental limits and provide deterministic and probabilistic guarantees 
on the fault detection and identification results. In the deterministic case, we draw connections between perception system monitoring and the literature on diagnosability in multiprocessor systems, and in particular 
 the PMC 
 model~\cite{Preparata67tec-diagnosability}. This allows us to establish formal guarantees on the maximum number of faults that can be uniquely identified in a given perception system, leading to the notion of \emph{diagnosability}.{\footnote{As discussed in~\cref{sec:foundamental_limits}, \emph{diagnosability} is related to the level of redundancy within the system and provides a quantitative measure of robustness.}} 
 In the probabilistic case, we develop Probably Approximate Correctly (PAC) bounds on the expected number of mistakes
 our runtime monitors will make.

Finally, we show that our framework is effective in detecting and identifying faults in a real-world perception pipeline for  obstacle detection (\cref{sec:experimental_section}). In particular, we perform experiments using a realistic open-source autonomous driving simulator (the \lgsvl Simulator~\cite{lgsvl-sim}) and a state-of-the-art autonomous driving software stack 
(Baidu's Apollo Auto~\cite{apollo-auto}). 
Our experiments show that 
(i) some of our algorithms outperform common baselines in terms of accuracy, 
(ii) they allow detecting failures and provide enough notice to stop the vehicle before an accident occurs in realistic scenarios,
and (iii) their runtime is typically below five milliseconds, incurring a negligible overhead in practice.
A video showcasing the execution of the proposed runtime monitors can be found at \url{\videourl}.
\\We have also released an open-source version of our code at \url{\repourl}.

%% file: sections/related_work.tex
\section{Related Work}\label{sec:relatedWork}

This section reviews related work on runtime monitoring and AV safety assurance, spanning
 both industrial practice (\cref{sec:stateOfPractice}) 
and academic research (\cref{sec:stateOfArt}).

\subsection{State of Practice}\label{sec:stateOfPractice}

The automotive industry currently uses four classes of methods to claim the safety of an AV~\cite{Shalev-Shwartz17arxiv-safeDriving}, namely: miles driven, simulation, scenario-based testing, and disengagement. 
Each of these methods has well-known limitations. 
The \emph{miles driven} approach is based on the statistical argument that if the probability of crashes per mile is lower in autonomous vehicles than for humans, then AVs are safer; however, such an analysis would require an impractical amount (\ie\!billions) of miles to produce statistically-significant results~\cite{Kalra16tra-selfDriving,Shalev-Shwartz17arxiv-safeDriving}.\footnote{Moreover, the analysis should cover all representative driving conditions (\eg driving on a highway is easier than driving in urban environment) and should be repeated at every software update, quickly becoming impractical.} 
The same approach can be made more scalable through \emph{simulation}, but unfortunately creating a life-like simulator is an open problem, for some aspects even more challenging than self-driving itself. 
\emph{Scenario-based} testing is based on the idea that if we can enumerate all the possible driving scenarios that could occur, then we can simply expose the AV (via simulation, closed-track testing, or on-road testing) to all these scenarios and, as a result, be confident that the AV will only make sound decisions. 
However, enumerating all possible corner cases (and perceptual conditions) is a daunting task.
Finally, \emph{disengagement} is defined as the moment when a human safety driver has to intervene in order to prevent a hazardous situation.
However, while less frequent disengagements indicate an improvement of the AV behavior, they do not give evidence of the system safety.

An established methodology to ensure safety is to develop a \emph{standard} that every manufacturer has to comply with.
In the automotive industry, the standard ISO 26262~\cite{iso26262} is a risk-based safety standard that applies to electronic systems in production vehicles.
A key issue is that ISO 26262 mostly focuses on electronic systems rather than algorithmic aspects, hence it does not readily apply to fully autonomous vehicles~\cite{Koopman16sae-autonomousVehicleTesting}.
The recent ISO 21448~\cite{iso21448}, which extends the scope of ISO 26262 to cover autonomous vehicles functionality, primarily considers mitigating risks due to unexpected operating conditions, and provides high-level considerations on best-practice for the development life-cycle.
Both ISO 26262 and ISO/PAS 21448 are designed for self-driving vehicles supervised by a human~\cite{Concas21book-validationFrameworksForAV}.
Koopman and Wagner~\cite{Koopman19safecomp-UL4600} propose a standard called UL 4600~\cite{ul4600} specifically designed for high-level autonomy (levels 4 and 5).
This standard focuses on ensuring that a comprehensive safety case is created, but it is technology-agnostic, meaning that it requires evidence of system safety without prescribing the use of any specific approach or technology to achieve it.

\subsection{State of the Art}\label{sec:stateOfArt}

Related work tries to tackle the problem of safety assurance using different strategies.
{\bf Formal methods}~\cite{Ingrand19irc-verificationTrends,Desai17icrv-verification,Hoxha16aaaiws-temporalLogic,Vasile17icra-motionPlanning, Dathathri17arxiv-synthesis,Ghosh16hscc-formalMethods,Li11fmmc-formalMethods,Li14tacas-formalMethods,Kloetzer08tac-temporalLogic}
have been recently used as a tool to study safety of autonomous systems.
These approaches have been successful for decision systems, such as obstacle avoidance~\cite{Mitsch17ijrr-verificationObstacleAvoidance}, road rule compliance~\cite{Roohi18arxiv-selfDrivingVerificationBenchmark}, high-level decision-making~\cite{Cardoso2020nasa-curiosityVerification}, and control~\cite{Jha17ar-SafeAutonomy, Pasqualetti13tac-CPSsecurity}, where the specifications are usually model-based and have well-defined semantics~\cite{Foughali18formalise-verificationRobots}.
However, they are challenging to apply to perception systems, due to the complexity of modeling the physical environment~\cite{Seshia16arxiv-verifiedAutonomy}, and the trade-off between evidence for certification and tractability of the model~\cite{Luckcuck19csur-surveyFormalMethods}.
One common approach is finding an example where the system fails (\ie \emph{falsification}).
Current approaches~\cite{Dreossi19arxiv-verifai,Fremont19ieee-scenic,Leahy19ijrr-formalMethods} consider high-level abstractions of perception~\cite{Shalev-Shwartz17arxiv-safeDriving,Balakrishnan19date-perceptionVerification,Dokhanchi18rv-perceptionVerification} or rely on simulation to assert the true state of the world~\cite{Dreossi19arxiv-verifai,Fremont19ieee-scenic,Dreossi17arxiv-testingDNN}. 
Other approaches focus on adversarial attacks for neural-network-based object detection~\cite{Cao19sigsac-adversarialAttackLidar, Boloor20jsa-attackingVision,Delecki22arxiv-testingPerception};
these methods derive bounds on the magnitude of the perturbation that induces incorrect detection result, and are typically used off-line~\cite{Akhtar18ieee-adversarialAttacks}.

Previous works on {\bf runtime fault detection and identification} focused on components of the perception system~\cite{Rahman21-runtimeMonitoring}.
Miller\setal~\cite{Miller22arxiv-falseNegativeObjectDetectors} propose a framework for quantifying false negatives in object detection.
For semantic segmentation, Besnier\setal~\cite{Besnier21cvf-triggeringFailures} propose an out-of-distribution detection mechanism, while
Rahman\setal~\cite{Rahman22ral-FSNet} propose a failure detection framework to identify pixel-level misclassifications.
Lambert and Hays~\cite{lambert21neurips-trustButVerify} propose cross-modality fusion algorithm to detect changes in high-definition map.
Liu and Park~\cite{Liu21tdsc-detectingPerceptionError} propose a methodology to analyze the consistency between camera image data and LiDAR data to detect perception errors.
Sharma \etal~\cite{Sharma21uai-scod} propose a framework for equipping any trained deep network with a task-relevant epistemic uncertainty estimate.
Several GPS/RTK integrity monitors have been proposed~\cite{Jing22tits-GPSintegrity, Hafez20ral-integrityMonitoring} to detect localization errors.
Another line of works leverages spatio-temporal information to detect failures. 
You\setal~\cite{You21-temporalCheckLiDAR} use spatio-temporal information from motion prediction to verify 3D object detection results.
Balakrishnan\setal~\cite{Balakrishnan19date-perceptionVerification, Balakrishnan21-percemon} propose the Timed Quality Temporal Logic (TQTL) to reason about desiderable spatio-temporal properties of a perception algorithm.

Kang\setal~\cite{Kang18nips-ModelAF} use model assertions, which similarly place a logical constraint on the output of a module to detect anomalies.
Fault-tolerant architectures~\cite{navarro20arxiv-hero} have been also proposed to detect and potentially recover from a faulty state, but these efforts mostly focus on implementing watchdogs and monitors for specific modules, rather than providing tools for system-level analysis and monitoring. 

{\bf Fault identification and anomaly detection} have been extensively studied {\bf in other areas of engineering}.
Bayesian networks, Hidden Markov Models~\cite{Cai17trii-bayesianNetFaultDiagnosis,Abdollahi16-PGMforFaultDiagnosis}, and deep learning~\cite{Lei20mssp-MLforFaultDiagnosis} have been used to enable fault identification, but mainly in industrial systems instrumented to detect component failures.  
Graph-neural networks have been used for anomaly detection (see~\cite{Ma21tkde-GNNAnomalyDetection} for a comprehensive survey).
In this context, ``anomaly detection is the data mining process that aims to identify the unusual patterns that deviate from the majorities in a dataset''~\cite{Ma21tkde-GNNAnomalyDetection}.
In order to detect anomalies, objects (\ie nodes, edges, or sub-graphs) are usually represented by features that provide valuable information for anomaly detection, and when a feature considerably differs from the others (or the training data), the object is classified as anomalous. 
%
De Kleer and Williams~\cite{DeKleer87-diagnosingMultipleFaults} propose a methodology to detect failures by comparing observations with a predicted output.
The dissimilarities are then used to search for potential failures that explain the measurements.
The work assumes the availability of a model that predicts the behavior of the system, and ---after collecting intermediate results of each component--- it searches for the smallest set of failing components that explains the wrong measurements.
Preparata, Metze, and Chien~\cite{Preparata67tec-diagnosability} study the problem of fault diagnosis in multi-processor systems, introducing the concept of diagnosability; their work is then extended by subsequent works~\cite{Hakimi74tc-diagnosability, Bhat82acm-diagnosability, Dahbura88cc-diagnosability}.
Sampath\setal~\cite{Sampath95trac-desDiagnosability} propose the concept of diagnosability for discrete-event systems~\cite{Zaytoon2013arc-desDiagnosability, Tuxi22deds-desDiagnosability}.
The system is modeled as a finite-state machine, and is said to be diagnosable if and only if a fault can be detected after a finite number of events.

The present paper extends this literature in several directions. First, we take a black-box approach and remain agnostic to the inner workings of the perception system we aim to monitor (relaxing assumptions in related work~\cite{DeKleer87-diagnosingMultipleFaults}). Second, we develop a fault identification framework that 
reasons over the consistency of heterogeneous and potentially asynchronous perception modules (going beyond the homogeneous, synchronous, and deterministic framework in~\cite{Preparata67tec-diagnosability}). 
Third, the framework is applicable to complex real-world perception systems (not necessarily modeled as discrete-event systems~\cite{Zaytoon2013arc-desDiagnosability, Tuxi22deds-desDiagnosability}). The present paper also extends our previous work on perception-system monitoring~\cite{Antonante21iros-perSysMonitoring2}, which only focuses on the deterministic case and considers a simplified model.

%% file: sections/problem_formulation.tex

\section{Problem Statement: Fault Detection and Identification \\ in Perception Systems}\label{sec:problem_formulation}

%% file: sections/subsections/problem_formulation.tex

\subsection{Perception System: Modules and Outputs}\label{sec:problem_statement}

A perception system $\sys$ comprises a finite set of interconnected \emph{\modules} $\moduleset=\{\modulevar_1,\modulevar_2,\ldots \modulevar_{|\moduleset|}\}$; for instance, the perception system of a self-driving car 
may include modules for lane detection, camera-based object detection, LiDAR-based motion estimation, 
ego-vehicle localization, etc.
Each \module~$\modulevar\in\moduleset$ produces an \emph{output}. 
For instance, the lane detection module may produce an estimate of the 3D location of the lane boundaries, while the 
pedestrian detection module may produce an estimate of the pose and velocity of pedestrians in the surroundings.
Some of these outputs provide inputs for other perception \modules, 
while other are the outputs of the perception system and feed into other systems (\eg to planning and control).
The set of modules' outputs are disjoint (\ie each output is produced by a single module), and the set of all outputs is denoted by $\outputset$.
%
We model the perception system 
as a graph of modules and outputs.
\begin{definition}[Perception System]
  A perception system $\sys$ is a directed graph $\sys=(\moduleset\cup\outputset, \calE)$, where
  the set of nodes $\moduleset\cup\outputset$ describes modules and outputs in the system, while the set of edges $\calE$
  describes which module produces or consumes a certain output. In particular,
  an edge $(\modulevar_i, \outputvar_j)\in \calE$ with $\modulevar_i\in\moduleset$ and $\outputvar_j\in\outputset$ models the fact that \module~$\modulevar_i$ produces output~$\outputvar_j$.
  Similarly, and edge $(\outputvar_j, \modulevar_i)\in \calE$ with $\outputvar_j\in\outputset$ and $\modulevar_i\in\moduleset$ models the fact that  \module~$\modulevar_i$ uses output~$\outputvar_j$.
\end{definition}

  We treat each module as a \emph{black-box} and remain agnostic to the algorithms they implement. 
  This allows our framework to generalize to complex perception systems, possibly including a combination of classical and data-driven methods.

\begin{figure}[!ht]
  \centering
  \includegraphics[width=\textwidth]{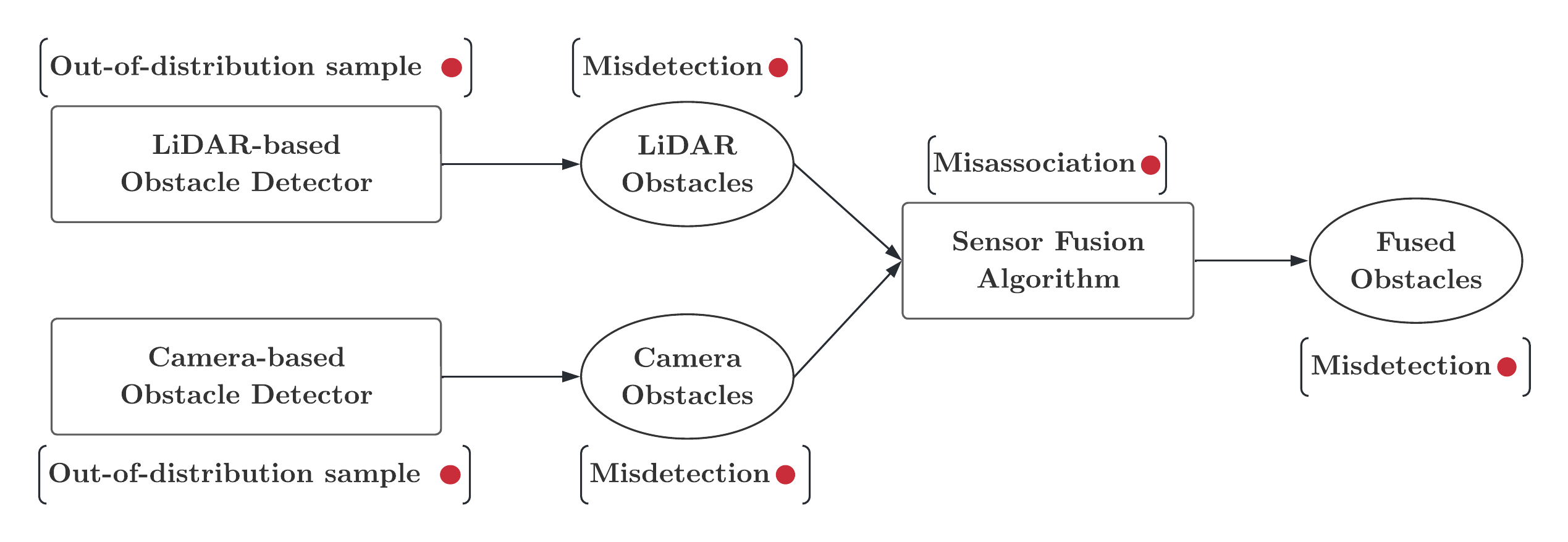}\vspace{-5mm}
  \caption{Perception system including 3 modules (rectangles) and 3 outputs (circles). 
  Modules are connected by edges describing which module produces or consumes a given output.
  The failure modes of each module (resp. output) are represented by red dots.
  \label{fig:runningExample}} 
\end{figure}

While we will consider more complex examples of perception systems in the experimental section, 
\cref{fig:runningExample} shows a simple example of perception system to ground the discussion. 
  The system comprises three modules: a \emph{LiDAR-based} obstacle detector, a \emph{camera-based} obstacle detector, and a \emph{sensor fusion} module.
  Both the LiDAR-based and the camera-based obstacle detectors generate a set of obstacles detected in the environment,  namely, the \emph{LiDAR obstacles} and \emph{camera obstacles}.
  The sensor fusion algorithm combines the two sets of obstacles to produce a new set of objects, called \emph{fused obstacles}.

\begin{remark}[Modules vs. Outputs]\label{rmk:modules_vs_outputs}
Our system model treats modules and outputs as separate nodes.
This is convenient for two reasons. 
First, fault identification at the modules and outputs may serve different purposes: 
 output fault identification is more useful at runtime to identify unreliable information from the perception system and prevent accidents; module fault identification is typically more informative for designers and regulators.
Second, in practical applications we can rarely \emph{measure} if a module is failing (indeed developing algorithms that can ``self-diagnose'' their failures is an active area of research, see work on \emph{certifiable algorithms}~\cite{Yang21arxiv-certifiablePerception}). 
On the other hand, we can directly measure the outputs of the modules and develop diagnostic tests 
to check if an output is plausible and consistent with other outputs in the system. 
\end{remark}

%% file: sections/subsections/fault_detection_vs_identification.tex

\subsection{Fault Detection and Fault Identification}\label{sec:fault_detection_vs_identification}

Each \module~in $\sys$ might fail at some point, jeopardizing the system performance or even its safety.
In particular, each \module~$\modulevar\in\moduleset$ is assumed to have a 
set of \fmodes. 
While the list of failures can include any software and hardware failures, 
  in this paper we particularly focus on failures of the intended functionality. For example, a neural-network-based camera-based object detection module might experience the \fmode ``out-of-distribution sample'' when it processes an input image, which indicates that while the module's code executed successfully, the resulting detection is expected to be incorrect. 

  Similarly, each output $\outputvar\in\outputset$ 
  has an associated set of \fmodes. 
For instance, the output of the camera-based object detector might experience a ``mis-detection'' \fmode if it fails to detect an object, or a ``mis-classification'' \fmode if the object is detected but misclassified. A module's \fmode typically causes a failure in one of its outputs.
 Examples of failure modes are given in~\cref{fig:runningExample}. 
  For each module and output, the figure lists a potential failure mode: for instance, the LiDAR-based obstacle detection output may fail if it misdetects an obstacle, while the sensor fusion module may fail it it incorrectly associates the input obstacles.

\begin{definition}[Failure Modes]
At each time instant, 
the $i$-th \fmode $\failure_i \in \{\mathrm{\finactive}, \mathrm{\factive}\}\cong\{0,1\}$ is either \factive~(also $1$) if such failure is occurring, or \finactive~(also $0$). 
A module or an output is \emph{failing} if at least one of its \fmodes~is \factive. 
If we stack the status (ACTIVE/INACTIVE) of all failure modes into a single binary vector, the \emph{fault state vector} $\faults \in 
\{0,1\}^\nrfmode$ (where $\nrfmode$ is the number of failure modes), then 
$\faults$ is all zeros if there are no faults, or has entries equal to ones for the active failure modes.
\end{definition}

The goal of this paper is then to address the following problems:
\begin{description}
  \item[Fault Detection] decide whether the system is working in nominal conditions or whether a fault has occurred (\ie infer if there is at least an active \fmode in $\faults$);
  \item[Fault Identification] identify the specific \fmode~the system is experiencing (\ie infer which \fmode is active in $\faults$).
\end{description}

Fault detection is the easiest between the two problems, as it only requires specifying the presence of at least a fault, without specifying which modules or outputs are incorrect. 
Mathematically, this reduces to identifying whether the  unknown vector $\faults$ has at least an entry equal to $1$.
Fault identification goes one step further by explicitly indicating the set of active \fmodes.
 Mathematically, this reduces to identifying exactly which entries of the  unknown vector $\faults$ 
 are equal to $1$.
 Identifying which module is faulty is particularly important to 
 inform regulators (\eg to trace the steps that that led to
an accident caused by an autonomous vehicle) and system designers (\eg to 
highlight modules that are likely to fail and require further development). 
Moreover, not all faults are equally problematic: for instance, a failure in localizing a car in the opposite lane of a divided highway is less consequential that failing to detect a pedestrian in front of the car.
 Note that solving fault identification implies a solution for fault detection (\ie whenever we
 declare one or more modules to be faulty, we essentially also detected there is a failure), hence 
 in the rest of this paper we focus on the design of a monitoring system for fault identification.
 
 \begin{remark}[Assumptions and Terms of Use]
We assume that the potential \fmodes of the system are known to the system designer.
In practice, these can be discovered using  
some form of hazard analysis, such as Failure Modes and Effects Analysis (FMEA)~\cite{Yang17sae-safetyCases} or Fault tree analysis (FTA)~\cite{Yan16scor-reliabilityModeling}.
Moreover, one can always add a generic ``unknown failure mode'' to the list of failure modes of a module or output, 
hence this assumption is not restrictive. 
We also remark that our monitoring system's objective is to diagnose potential failures, while it does not prescribe what are the actions that need to be taken in response to each failure (\eg whether to stop the car, provide a warning to the passenger, etc.), which is failure and system-dependent. Investigation how to respond to or mitigate failures is left to future work.
\end{remark}

%% file: sections/approach_formulation.tex
\section{Modeling Fault Identification with Diagnostic Graphs}\label{sec:approach_formulation}


This section develops a framework to model fault identification problems in 
perception systems. In the previous section we have discussed how the goal is to identify the set 
of active failure modes associated to modules and outputs in a system. 
Here we introduce the concept of \emph{\dgraphs} to study fault identification: \dgraph will allow developing fault identification algorithms (\cref{sec:algorithms}) and understanding fundamental limits (\cref{sec:foundamental_limits}).

The intuition is that in a perception system we can perform a number of \emph{\dtests} that check the validity of the output of certain modules. For instance, we can compare the outputs of different modules to ensure they are consistent (\eg compare the obstacles detected by the LiDAR-based obstacle detection against the camera-based obstacle detection), 
or inspect that 
the output a certain module respects certain requirements 
(\eg the vision-based ego-motion module is tracking a sufficient number of features). 
Then, we can model these checks as edges in a bipartite graph, the \emph{\dgraph},  which can be
 used for fault identification. 
In the following, we formalize the notions of \dtests and \dgraphs.

%% file: sections/subsections/diagnostic_tests.tex
\subsection{\DTests}\label{sec:dtests}


In our fault identification framework, 
the system is equipped with a set of \emph{\dtests} that can (possibly unreliably) provide diagnostic information about the state of a subset of \fmodes.
Each \dtest is a function $\test:\Scope\to\{\mathpass,\mathfail\}$, where $\Scope\subseteq \{1,\ldots,\nrfmode\}$ is 
a subset of the failure modes that the test is checking, called the \emph{scope} of the test,
and the test returns a value $\syndromevar\in \{\mathpass,\mathfail\}\cong\{0,1\}$, 
called the \emph{outcome} of the test.
A \dtest returns \pass (also denoted with $0$) if there is no active \fmode in its scope, \fail (also denoted with $1$) otherwise.
In general, tests can be \emph{unreliable}, meaning that they can both fail to detect active failures or incorrectly detect failures as active (\ie false alarms).

\begin{figure}[!ht]
  \centering
  \begin{floatrow}
    \ffigbox[\FBwidth]{
      \includegraphics[width=3cm]{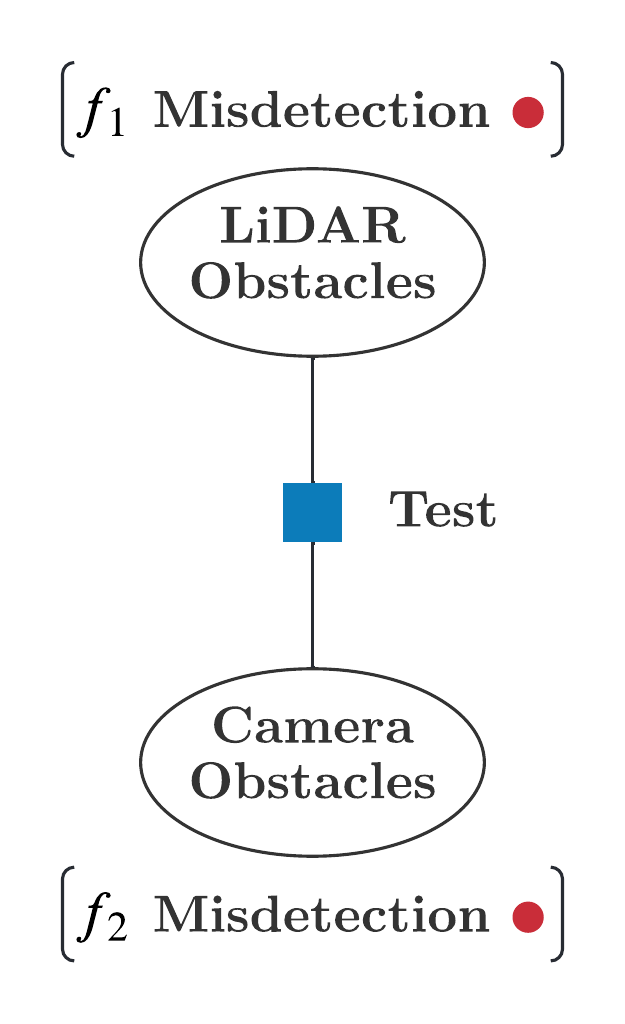} 
      }{
        \caption{A test comparing two outputs, LiDAR Obstacles and Camera Obstacles}
        \label{fig:runningExample-test}
    }
    \capbtabbox{
      \smaller{\begin{tabular}{c|c||c|c}
        \multicolumn{2}{c||}{Scope} & \multicolumn{2}{c}{ {Test outcome} $\syndromevar$} \\ \hline
          $\failure_1$ & $\failure_2$ & OR & Noisy-OR \\ \hline
          $0$ & $0$ & $\quad0\quad$ & 
          $\left\{\begin{array}{ll} 0 \text{ with prob. } (1 - p_{a,1}) (1 - p_{a,2}) \\ 1 \text{ with prob. } p_{a,1} + p_{a,2} - p_{a,1}p_{a,2} \end{array}\right.$\\
          \hline
          $0$ & $1$ & $1$ & 
          $\left\{\begin{array}{ll} 0 \text{ with prob. } (1 - p_{a,1}) (1 - p_{d,2}) \\ 1 \text{ with prob. } p_{a,1} + p_{d,2} - p_{a,1}p_{d,2} \end{array}\right.$\\
          \hline
          $1$ & $0$ & $1$ & 
          $\left\{\begin{array}{ll} 0 \text{ with prob. } (1 - p_{d,1}) (1 - p_{a,2}) \\ 1 \text{ with prob. }  p_{d,1} + p_{a,2} - p_{d,1}p_{a,2} \end{array}\right.$\\
          \hline
          $1$ & $1$ & $1$ & 
          $\left\{\begin{array}{ll} 0 \text{ with prob. } (1 - p_{d,1}) (1 - p_{d,2}) \\ 1 \text{ with prob. }  p_{d,1} + p_{d,2} - p_{d,1}p_{d,2} \end{array}\right.$ 
      \end{tabular}}
    }{
      \caption{Table of possible outcomes for the Deterministic OR and the probabilistic Noisy-OR version of a test with scope $\failure_1$ and $\failure_2$.}
      \label{tab:tfactor_1}
    }
    \end{floatrow}
\end{figure}

While in the experimental section we will describe more complex tests (and provide an open-source framework\footnote{Code available at \url{\repourl}.}
to easily code new tests), it is instructive to consider a simple test between the outputs of the LiDAR-based
obstacle detection and the camera-based obstacle detection in~\cref{fig:runningExample-test}.
The test in~\cref{fig:runningExample-test} compares the two sets of objects detected by the two detectors;
whenever an inconsistency arises, the test returns \fail.
However, if both detectors are subject to the same failure, \eg they both misdetect an obstacle, the test might still pass, thus exhibiting unreliable behavior. 
We remark that a single test does not suffice for fault identification: for instance, 
if the test in~\cref{fig:runningExample-test} fails, we can only conclude that one of the two detectors had a failure (or that the test was a false alarm); therefore, we typically need to collect a number of tests and 
perform some inference process to draw conclusions about which modules failed. 
The collection of the outcomes of multiple \dtests is called a \emph{syndrome}.

\begin{definition}[Syndrome]
Assuming we have $\nrtests$ \dtests, the vector collecting the test outcomes 
$\syndrome\in\{\mathpass,\mathfail\}^\nrtests$ is called a \emph{syndrome}. 
\end{definition}

In the following, we describe how to mathematically model the relation between the \fmodes and the test outcomes;
this will be instrumental in solving the inverse problem of identifying the \fmode from a given syndrome.
We provide a deterministic and a probabilistic model for the tests below. 

\myParagraph{Deterministic Tests}
Deterministic \dtests encode the set of possible test outcomes, by establishing a deterministic relation between \fmodes in the test's scope and the test outcome. We discuss potential models for 
deterministic diagnostic test below.

Ideally we would like the test to return \fail if and only if at least one of the \fmodes in its scope is active. 
This leads to the definition of a ``Deterministic OR'' test.
\begin{definition}[Deterministic OR]
  A \dtest $\test(\faultsover{\scope(\test)})$ is a deterministic OR if its test outcome $\syndromevar$ is
  \begin{equation}\label{eq:deterministic_or}
    \syndromevar = \begin{cases}
      \mathpass & \text{if } \|\faultsover{\scope(\test)}\|_1 = 0 \\
      \mathfail & \text{otherwise}
    \end{cases}
  \end{equation}
\end{definition}
This kind of tests can be hard to implement in practice. 
For example, imagine a \dtest that compares the output of two object classifiers: if one of them produces a wrong label, 
it is easy to detect there is a failure; however, if both classifiers are trained on similar data and both report the incorrect label there is no way to detect the failure.
In this case, the test outcome is unreliable. The following definition introduces a type of unreliable test.
\begin{definition}[Deterministic Weak-OR]
  A test $\test(\faultsover{\scope(\test)})$ is a deterministic Weak-OR if its test outcome $\syndromevar$ is
  \begin{equation}\label{eq:deterministic_weak_or}
    \syndromevar = \begin{cases}
      \mathpass & \text{if } 0 < \|\faultsover{\scope(\test)}\|_1 < |\scope(\test)| \\
      \mathpass \text{ or } \mathfail & \text{if } \|\faultsover{\scope(\test)}\|_1 = |\scope(\test)| \\
      \mathfail & \text{otherwise}
    \end{cases}
  \end{equation}
\end{definition}
This kind of test is consistent with the tests used in~\cite{Antonante20tr-perSysMonitoring2}. 
Intuitively, a ``Deterministic Weak-OR'' may return \pass even if all failure modes are active, since 
the test might fail to detect an inconsistency if all faults are consistent with each others
(again, think about two object classifiers failing in the same way).
Even though the Weak-OR test may pass or fail when all failure modes are active, 
its outcome remains deterministic. 

Finally, an even weaker type of deterministic test is what we call the Deterministic \WeakerOR (this is the easiest test to implement in practice).
\begin{definition}[Deterministic \WeakerOR]
  A \dtest $\test(\faultsover{\scope(\test)})$ is a Deterministic \WeakerOR if its test outcome $\syndromevar$ is
  \begin{equation}\label{eq:deterministic_blind_or}
    \syndromevar = \begin{cases}
      \mathpass \text{ or } \mathfail & \text{if } \|\faultsover{\scope(\test)}\|_1 > 0 \\
      \mathpass & \text{if } \|\faultsover{\scope(\test)}\|_1 = 0 \\
    \end{cases}
  \end{equation}
\end{definition}
In other words, the tested is designed to pass in nominal conditions (\ie when no failure mode is active), but 
it can have arbitrary outcomes otherwise.

The types of deterministic tests presented above are not the only possible deterministic tests. 
Other examples include, for instance, \dtests that fail to detect specific sets of failure modes. 
Deterministic tests can be designed using formal methods tools or certifiable perception algorithms~\cite{Yang20neurips-certifiablePerception,Yang20cvpr-shapeStar,Yang19rss-teaser},\footnote{
Certifiable perception algorithms are a class of model-based perception algorithms that provide a soundness certificate at runtime, allowing one to directly measure the presence (or absence) of certain failure modes, see~\cite{Yang19rss-teaser,Yang21arxiv-certifiablePerception}.}  see also~\cref{rmk:test_from_literature} below.

\myParagraph{Probabilistic Tests} 
Deterministic tests might not capture the complexity of real world \dtests.
Most practical tests are likely to incorrectly detect faults (\ie produce false positive) or fail to detect faults (\ie produce false negatives) with some probability.
For this reason, in this paper, we also allow for an arbitrary probabilistic relationship between test outcomes and \fmodes~in the test scope.

A simple-yet-expressive way to formalize a probabilistic test is to use what we call a  ``Noisy-OR'' model.
In particular, the Noisy-OR model represents the probability of a \dtest outcome as a conditional probability distribution over the failure modes in its scope $\Pr(\syndromevar\mid\faultsover{\scope(\test)})$ as defined below.

\begin{definition}[Noisy-OR]
A \dtest $\test(\faultsover{\scope(\test)})$ is a probabilistic Noisy-OR if its test outcome $\syndromevar$ follows
\begin{align}\label{eq:noisy_or}
  \Pr(\syndromevar=\mathpass\mid\faultsover{\scope(\test)}) =& \prod_{i\in\scope(\test)} \Pr(\syndromevar=\mathpass\mid\failure_i)
\end{align}
where $\Pr(\syndromevar\mid\failure_i)$ denotes the conditional probability of the test outcome (\pass/\fail) 
conditioned on the status (\factive/\finactive) of the failure mode $\failure_i$. Clearly, 
$ \Pr(\syndromevar=\mathfail\mid\faultsover{\scope(\test)}) = 1 - \Pr(\syndromevar=\mathpass\mid\faultsover{\scope(\test)})$.
\end{definition}

Now suppose each test has a probability $\probDetect{i}$ of correctly identifying failure $\failure_i$ (detection probability), and a probability $\probFalseAlarm{i}$ of false alarm for $\failure_i$. 
Exploiting the fact that $\failure_i\in\{0,1\}$, we can write~\cref{eq:noisy_or} as:
\begin{equation}\label{eq:noisy_or_prob_pass}
  \Pr(\syndromevar=\mathpass\mid\faultsover{\scope(\test)}) = \prod_{i\in\scope(\test)}(1-\probDetect{i})^{\failure_i} (1-\probFalseAlarm{i})^{1-\failure_i}
\end{equation}

An example of probabilistic test outcome is given in~\cref{tab:tfactor_1}.


Similarly to the deterministic case, the Noisy-OR model is not the only possible model. 
However,~\cref{sec:experimental_section} shows that this model is particularly effective in modeling fault identification problems in practice.
In~\cref{sec:algorithms}, we discuss how to learn the probabilities involved in probabilistic tests  (\ie 
$\probDetect{i}$ and  $\probFalseAlarm{i}$ in~\cref{eq:noisy_or_prob_pass})
given a training dataset, 
and how to use the test outcomes to infer the most likely failure modes. 
Towards that goal, we need to group \dtests into a suitable graph structure, called a \emph{\dgraph}, which 
we present in the following section. 
We conclude this section with a remark. 

\begin{remark}[From \dtests to fault identification] \label{rmk:test_from_literature}
The \dtests we introduced in this section are not dissimilar from the typical 
diagnostic tests or watchdogs considered in prior work or used by practitioners. 
Our goal here is to formalize these tests and use the test outcomes to infer the most likely set of system-wide failures. In this sense, our fault identification framework is designed to capitalize on (rather than replace) existing diagnostic tools used in practice.  
  For example the detection mechanism proposed by Liu and Park~\cite{Liu21tdsc-detectingPerceptionError}, which is based on the idea of projecting the 3D LiDAR points onto camera images, and then checking whether objects detected from LiDAR and images match each other, can be formulated as a \dtests with the camera and LiDAR misdetection in its scope, such that the test outcome is the output of the algorithm in~\cite{Liu21tdsc-detectingPerceptionError}.
  Also, out-of-distribution detection based on epistemic uncertainty, \eg~\cite{Sharma21uai-scod}, can be formulated as a \dtests with the module's ``out-of-distribution sample'' \fmode in its scope, such that the test outcome is \fail if the estimated uncertainty is above a threshold.
\end{remark}


%% file: sections/subsections/diagnostic_graph.tex
\subsection{\DGraph}

A \dgraph is a structure defined over a perception system and has the goal of describing the 
\dtests (as well as more general relations among failure modes) and their scope. We provide a formal definition below.

\begin{definition}[\DGraph]
A \emph{\dgraph} is a bipartite graph $\dsys=(\varset,\relset,\edgeset)$ where the nodes are 
split into \emph{variable nodes} $\varset$, corresponding to the \fmodes in the system,
and \emph{relation nodes} ${\relset}$, where each relation 
$\rel_k(\faults)\in\relset$ 
is a function 
over a subset of \fmodes $\faults$.
Then an edge in $\edgeset$ exists between a \fmode $\failure_i\in\varset$ and a \relation $\rel_k \in \relset$, 
 if $\failure_i$ is in the scope of the \relation $\rel_k$ (\ie if the variable $\failure_i$ appears in the function
 $\rel_k$). 
\end{definition}

Relations capture constraints among the variables induced by the test outcomes or from prior knowledge we might have about the \fmodes. We describe the two main types of relations below and for each we describe their implementation in the deterministic and probabilistic case.

\begin{definition}[Test-driven Relations]
A \emph{test-driven relation} $\rel_k$ describes whether ---for a test $\test_k$--- a given set of \fmode assignments might have produced a certain test outcome $\syndromevar_k$. 
More formally, for a deterministic test $\test_k$, a test-driven relation is a boolean function:
\begin{equation}\label{eq:testRelations-deterministic}
 \rel_k(\faults) = \rel(\faultsover{\scope({\test_k})}; \syndromevar_k) = 
\ind\left[ \syndromevar_k = \test\left(\faultsover{\scope({\test_k})}\right) \right]
\end{equation}
where $\ind$ is the indicator function that returns $1$ if the condition is satisfied or $0$ otherwise.
The function~\cref{eq:testRelations-deterministic} checks if an assignment of \fmodes $\faults$ may have produced the test outcome $\syndromevar_k$ 
and where the notation $\rel_k(\faults) = \rel(\faultsover{\scope({\test_k})}; \syndromevar_k)$
clarifies that the function $\rel_k$ only involves a subset of \fmodes $\faultsover{\scope({\test_k})}$ (the ones in the scope of test $\test_k$) and depends on the (given) test outcome $\syndromevar_k$.
Similarly, for a probabilistic test $\test_k$, a test-driven relation
is a real-valued function:
\begin{equation}
\rel_k(\faults) = \rel(\faultsover{\scope({\test_k})}; \syndromevar_k) =
 \Pr(\syndromevar_k | \faultsover{\scope(\test_k)})
\end{equation}
which returns the likelihood of the test outcome $\syndromevar_k$ 
 given an assignment $\faults$.
\end{definition}

\begin{definition}[A Priori Relations]
An \emph{a priori relation} describes whether a given set of \fmodes is plausible, considering a priori knowledge about the system.
More formally, in the deterministic case, an a priori relation is a boolean function $\rel_k(\faults)$ that
returns $\btrue$ if the assignment of $\faults$ is plausible or $\bfalse$ otherwise.
Similarly, in the probabilistic case, an a priori relation
is a real-valued function $\rel_k(\faults)$ that returns the likelihood of a  given assignment $\faults$.
\end{definition}

In the following we will denote the set of Test-driven Relations as $\relsetTest$ while the set of A Priori Relations as $\relsetPrior$.
Therefore, $\relset = \relsetTest \cup \relsetPrior$. 

\begin{figure}[!ht]
  \centering
  \vspace{-3mm}\includegraphics[width=0.7\textwidth]{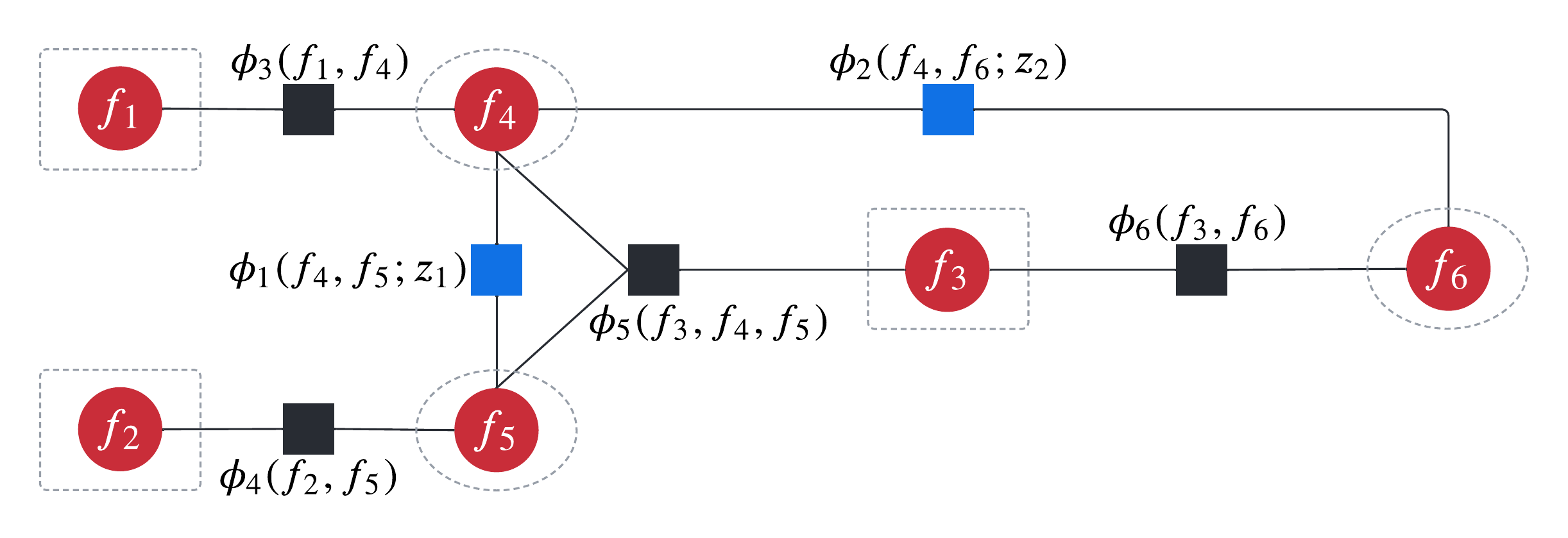}\vspace{-3mm}
  \caption{
    A \dgraph for the perception system example in~\cref{fig:runningExample}. Red circles represent variable nodes  
    (\fmodes) while squares represent \relations.
    Test-driven Relations are shown in blue, while a priori relations are shown in black.
   \label{fig:runningExample-faults}}
\end{figure}

While we have provided several examples of \dtests in the previous section, 
we now provide examples of a priori relations. 
For instance, in the deterministic case, some \fmodes of a module can be mutually exclusive (\eg ``too many outliers'', ``not enough features'' in the Lidar-based ego-motion estimation) or one can imply another (\eg if a module is experiencing an ``out-of-distribution sample'' \fmode, then its outputs will have at least an active failure mode). 
Not all relations are deterministic, for example in~\cref{fig:runningExample-faults}, the \fmodes of the sensor fusion algorithm may have a complex probabilistic relationship with the failure modes of the lidar and camera obstacles failure modes. 
Note that the main difference between test-driven relations and a priori relations is that the 
former provides a measurable test outcome, while the latter relies on a priori knowledge about the system
(\ie no outcome is measured).

We elucidate on the notion of \dgraph with two examples below.
\myParagraph{Example 1: Multi-sensor Obstacle Detection}
Consider the perception system in \cref{fig:runningExample}.
  We can associate a \dgraph to the system where the variable nodes of the \dgraph are the \fmodes of  modules and outputs in the system. The \dgraph, shown in~\cref{fig:runningExample-faults}, also includes two \dtests and
    a priori relations encoding input/output relationship between modules and outputs. 
  Each \dtest compares a pair of outputted obstacles, namely LiDAR obstacles and camera obstacles (with failures $\failure_4$ and $\failure_5$), and
  camera obstacles and fused obstacles (with failures $\failure_4$ and $\failure_6$).

\begin{figure}[!ht]
  \centering
  \vspace{-3mm}\includegraphics[width=0.7\textwidth]{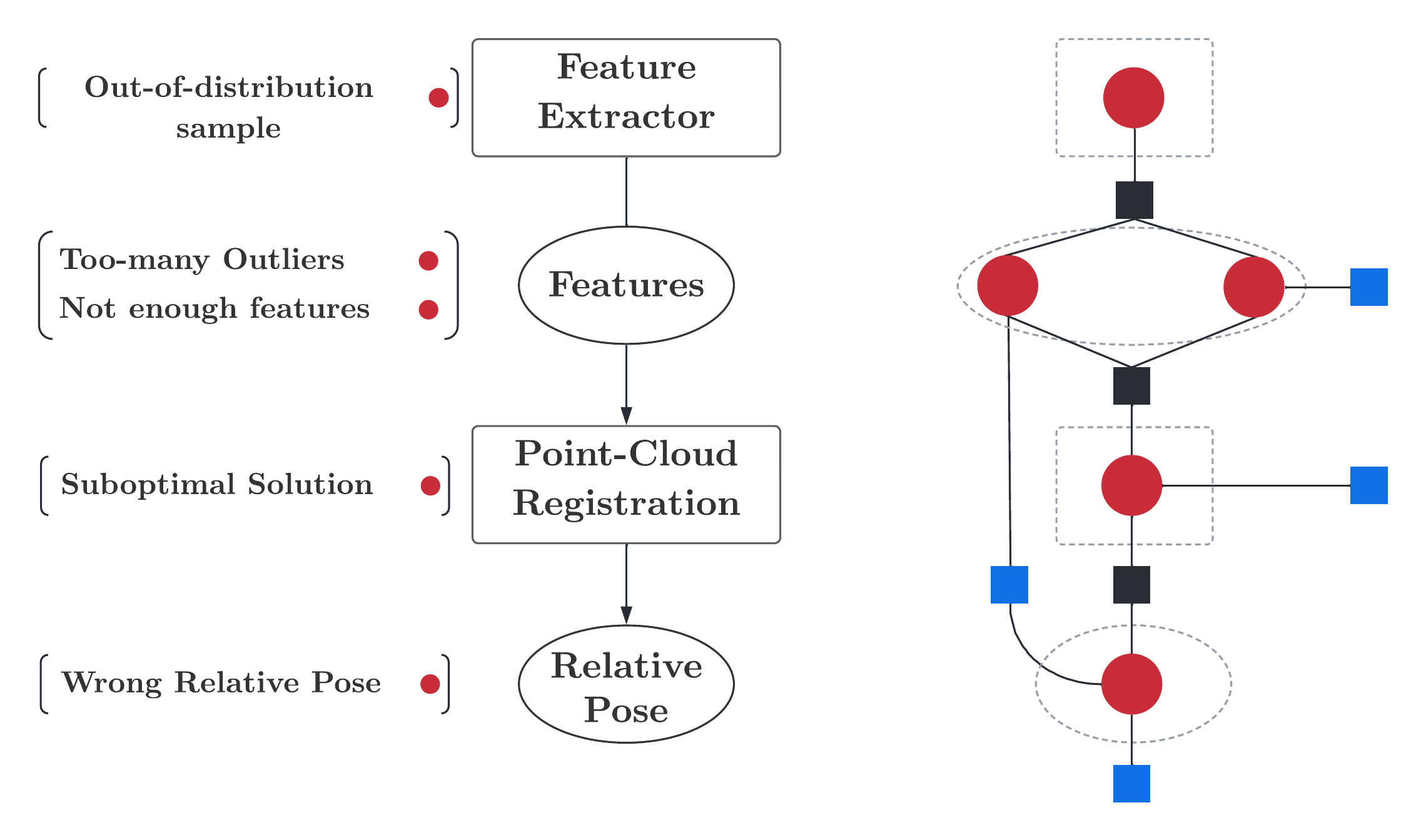}\vspace{-3mm}
  \caption{
    \emph{(Left)} Example of the LiDAR-based ego-motion estimation system $\sys$. 
    The system is composed by two modules (rectangles), each producing one output (circles).
    \emph{(Right)} The corresponding \dgraph, where red circles represent variable nodes  
    (\fmodes) while squares represent \relations (test-driven Relations in blue, a priori relations  in black).
  }
  \label{fig:example-ego-motion}
\end{figure}

\myParagraph{Example 2: LiDAR-based Ego-motion Estimation} 
We provide a second example that also includes singleton \dtests (having a single \fmode in their scope) and includes explicit tests over modules. 
The example consists of a LiDAR-based odometry system that computes the relative motion between consecutive LiDAR scans 
using feature-based registration, see~\eg~\cite{Cadena16tro-SLAMsurvey,Yang20tro-teaser}. 
The system $\sys$ comprises two \modules, a \emph{feature extraction} \module and a \emph{point-cloud registration} \module, as depicted in~\cref{fig:example-ego-motion}(left).
The feature extraction \module extracts 3D point \emph{features} from input LiDAR data, while the point-cloud registration \module uses the features to 
estimate the relative pose between two consecutive LiDAR scans.
Suppose that the feature extraction \module is based on a deep neural network and that it can experience an  ``out-of-distribution sample'' failure, which causes the corresponding output to potentially experience 
``too-many outliers'' or ``few features'' failures.
Similarly, the \module~\emph{point-cloud registration} can experience the failure ``suboptimal solution'', which leads its outputs, the relative pose, to experience a  ``wrong relative pose'' failure. 
\cref{fig:example-ego-motion}(right) shows a \dgraph for the system.
The system is equipped with three \dtests.
A \dtest detects if the \fmode ``few features'' is active by checking the cardinality of the feature set.
If the point-cloud registration \module~is a certifiable algorithm~\cite{Yang21arxiv-certifiablePerception}, 
 we can attach a \dtest to the point-cloud registration module that uses the module's certificate to detect if the module is experiencing a ``suboptimal solution'' failure.
Another \dtest detects if the relative pose is wrong by checking that the relative pose does not exceed some meaningful threshold given the vehicle dynamics.
Finally, another test checks if under the computed relative pose, the feature extractor has ``too many outliers''.
This can be achieved by 
counting the number of  features that are correctly aligned after applying the estimated relative pose.
{The \dgraph also contains a priori relations encoding constraints on the input/output relationships.}

\subsubsection{Temporal \DGraph}
So far, we have considered a \dgraph as a representation of the diagnostic information available at a specific instant of time (\eg the examples above include tests and relations involving the behavior of modules and outputs at a certain time instant).
However, perception systems evolve over time, and considering the temporal dimension offers further opportunities for 
fault identification, \eg by monitoring  the temporal evolution of the outputs. 

\begin{figure}[htbp]
  \centering
  \includegraphics[width=0.8\textwidth]{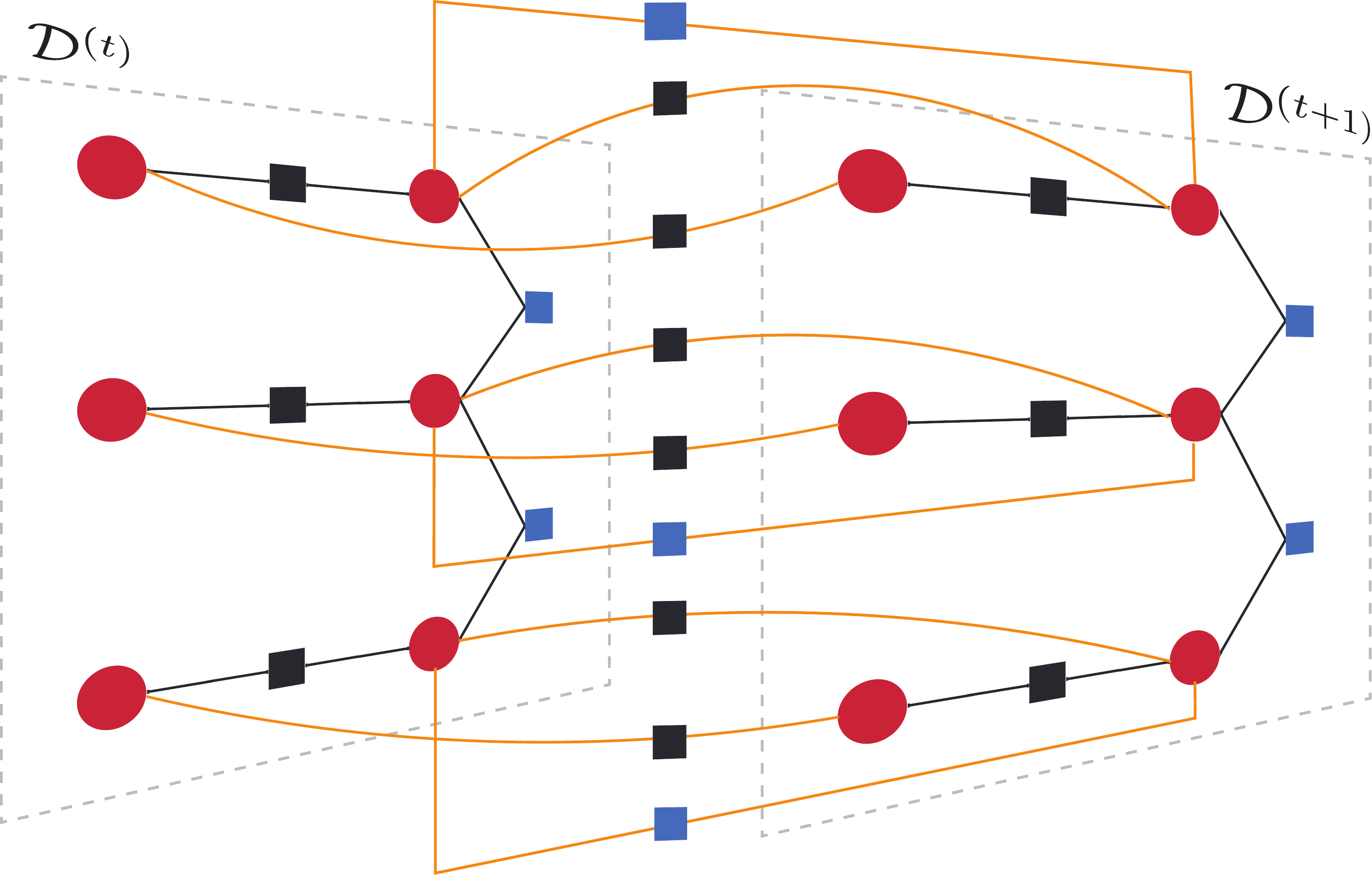}
  \caption{
    Example of Temporal \DGraph composed by two identical sub-graphs. 
    We added temporal \relations (both test-driven and a priori) between the two sub-graphs.
  \label{fig:example_temporal_dgraph}} 
\end{figure}

Suppose we have a collection of \dgraphs $\calT=\{\dsys\at{t},\ldots,\dsys\at{t+K}\}$, collected over and interval of time.
We could think of \emph{stacking} these \dgraphs, into a new \emph{temporal} \dgraph $\dsys\sqat{K}$.
The temporal graph preserves the \fmode, \relations and edges of each sub-graph $\dsys\at{k}\in\calT$.
However, since $\dsys\sqat{K}$ includes outputs produced at multiple time instants,  
we can also augment the graph to include temporal \dtests and temporal relationships. 
For example, we might check that an obstacle does not disappear from the scene (unless it goes out of the sensor field of view), or that the pose of the ego-vehicle does not change too much over time.
As we will see, the use of temporal \dgraph leads to slightly improved fault identification performance. 
An example of temporal \dgraph is given in~\cref{fig:example_temporal_dgraph}.

The algorithms and results presented in the rest of this paper apply to both regular and temporal \dgraph, unless specified otherwise.

\begin{remark}[Temporal \DTests] 
  Temporal \dtests are used to monitor the evolution of the system over time.
  For example the Timed Quality Temporal Logic in~\cite{Dokhanchi18rv-perceptionVerification} can be implemented with a temporal \dtest that spans multiple $\dsys\at{t}$'s.
  More specifically, the test example considered in~\cite{Dokhanchi18rv-perceptionVerification} requires that ``At every time step, for all the objects in the frame, if the object class is \emph{cyclist} with probability more than $0.7$, then in the next $5$ frames the same object should still be classified as a cyclist with probability more than $0.6$''. This  can be modeled as a \dtest that spans $5$ \dgraphs and that returns \fail if the predicate is false.
\end{remark}

%% file: sections/algos_for_identification.tex
\section{Algorithms for Fault Identification}
\label{sec:algorithms}

This section  shows how to perform fault identification over a \dgraph. In particular, we present algorithms to
  infer  which \fmodes are active, given a syndrome.
We study fault identification with deterministic tests in \cref{sec:inference_deterministic} and then extend it to the probabilistic case in \cref{sec:inference_on_fg}. Finally, we present a graph-neural-network approach for fault identification in \cref{sec:graph_neural_networks}. 

%% file: sections/subsections/inference_on_deterministic.tex

\subsection{Inference in the Deterministic Model}\label{sec:inference_deterministic}

In the deterministic case, our inference algorithm looks for the smallest set of active failure 
modes that explains a given syndrome. In \cref{sec:foundamental_limits}, we will show that 
such approach is guaranteed to correctly identify the faults as long as the tests 
provide a sufficient level of redundancy, an insight we will formalize through the notion of ``diagnosability''.

Looking for the smallest set of active failures that explains the test outcomes (and more generally, the relations) 
in a \dgraph can be formulated as the following optimization problem (given a syndrome $\syndrome$):
\begin{equation}\label{eq:deterministic_fault_identification}
\tag{D-FI}
  \begin{aligned}
    & \underset{\faults\in\{0,1\}^\nrfmode}{\textrm{minimize}} & & \|\faults\|_1 & \\
    & \textrm{subject to}
    & & \rel_k(\faultsover{\scope(\test_k)};\syndromevar_k)=1, & i = 1, \ldots, \nrtests,\\
    &&& \rel_j(\faults) = 1, &  j = 1, \ldots, N_r,
  \end{aligned}
\end{equation}
where $\rel_k(\faultsover{\scope(\test_k)};\syndromevar_k)$ are the $\nrtests$ test-driven relations 
in the \dgraph, while $\rel_j(\faults)$ are the $N_r$ a priori relations in the graph.
In words, \cref{eq:deterministic_fault_identification} looks for binary decisions (\factive/\finactive)  
 for the \fmodes $\faults$, and looks for the smallest set of faults (the objective $\|\faults\|_1$ counts the number of \factive~\fmodes) 
 such that the faults satisfy the relations in the \dgraph. 
 \cref{eq:deterministic_fault_identification} is our \emph{Deterministic Fault Identification} algorithm.

The optimization in \cref{eq:deterministic_fault_identification} can be solved using standard computational tools from Integer Programming~\cite{Wolsey2020-integerProgramming} or 
Constraint Satisfaction Programming~\cite{Rossi2006book-CSP}. 
While integer programming is better suited to find the solution to the minimization problem, constraint programming also allows finding all the solutions in the feasible set.
The choice between the two depends on the application and the expression for the relations.
In our experiments, we solve it using Integer Programming. We remark that while Integer Programming is NP complete, our problems typically only involve tens to hundreds of failure modes, and can be solved efficiently in practice.

\omitted{
\myParagraph{Implementation and Generalizations}
While minimizing the number of active \fmodes is the most natural thing to do, there are cases where we might still be interested in all the set of possible failure modes, excluding the trivial ones.
For example, imagine a system that, among several other \fmodes, has one where the camera experience an `image overexposure` \fmode, meaning that the image is too bright to be used for ego-motion estimation.
At the same time, the GPS has a `signal loss' \fmode, meaning there is an interruption of the GPS signal caused by tall buildings.
In this case, we might be interested in knowing if, given the observed syndrome, the two failures may be active at the same time, because in such cases the localization reliability is too low and the vehicle must take precautionary actions.
}

The model presented above is generic and valid for any deterministic test and a priori relations.
In the following, we provide an example to ground the discussion and show how to instantiate the optimization problem in practice.

\myParagraph{Example 3: Deterministic Inference with \WeakerOR and Module-Output Relations}
We consider a \dgraph with Deterministic \WeakerOR tests. Moreover, as a priori relations, we assume that 
whenever the output of a module has a failure, then also the module itself must have at least an active failure mode.
 This is also the setup we use in our experiments in~\cref{sec:experimental_section}.

In \WeakerOR \dtests, the \pass outcome is unreliable, meaning that if a test returns \pass it might have $0$ or more \fmodes active in its scope. 
However, when it the test returns \fail, we know there must be at least one \fmode active.
This can be easily enforced in the optimization by imposing 
the constraint:
\begin{equation*} 
  \|\faultsover{\scope(\test_i)}\|_1 \geq 1 \qquad
  \forall\test_i\in\testset\text{ such that }\syndromevar_i = \mathrm{\fail},
\end{equation*}
We then have to enforce the relation that if an output has an active failure mode,
then the module that produced it
must have at least one active \fmode as well.
Towards this goal, let $\calF(\outputvar_i)\subseteq\failureset$ be the set of \fmodes associated to outputs 
of module $\modulevar_i$ and  $\calF(\modulevar_i)$ be the set of \fmodes associated to $\modulevar_i$; 
then the a priori relation can be enforced via the constraint:
\begin{align*}
  \|\faultsover{\calF(\modulevar_i)}\|_1 &\geq \frac{1}{|\calF(\outputvar_i)|} \|\faultsover{\calF(\outputvar_i)}\|_1
\end{align*}
Intuitively, when there is no active failure in the outputs (\ie $\|\faultsover{\calF(\outputvar_i)}\|_1 = 0$) the constraint is trivially satisfied, while when there is at least an output failure (\ie $\|\faultsover{\calF(\outputvar_i)}\|_1 > 0$) then $\|\faultsover{\calF(\modulevar_j)}\|_1$ is forced to be at least $1$.
The resulting optimization problem finally becomes:
\begin{equation}\label{eq:optimization_weakerOR}
  \begin{aligned}
    & \underset{\faults\in\{0,1\}^\nrfmode}{\textrm{minimize}} & & \|\faults\|_1 \\
    & \textrm{subject to}
    & & \|\faultsover{\scope(\test_i)}\|_1 \geq 1 \qquad
  \forall\test_i\in\testset\text{ such that }\syndromevar_i = \mathrm{\fail}, \\
    &&& \|\faultsover{\calF(\modulevar_i)}\|_1 \geq \frac{1}{|\calF(\outputvar_i)|} \|\faultsover{\calF(\outputvar_i)}\|_1 \qquad \forall \modulevar_i \in \moduleset.
  \end{aligned}
\end{equation}

%% file: sections/subsections/inference_on_factor_graphs.tex
\subsection{Inference in the Probabilistic Model}\label{sec:inference_on_fg}

%
This section shows how to use the formalism of factor graphs to find 
the most likely active failure modes that explain a given syndrome in a \dgraph with probabilistic tests.


Factor graphs are a powerful class of probabilistic graphical models.
Probabilistic graphical models allow describing relationships between multiple variables using a concise language.
In particular, they describe joint or conditional distributions over a set of unknown variables and a set of known observations, and can be used to infer the values of the unknown variables.
In this work we limit ourselves to factor graphs over discrete (binary) variables.
We start from the definition of a factor graph.
\begin{definition}[Factor Graph]
  A factor graph is a bipartite graph $F=(\fgvarset,\factorset,\fgedgeset)$ consisting of a set $\fgvarset$ of variable nodes, a set $\factorset$ of factor nodes, and a set $\fgedgeset\subseteq\fgvarset\times\factorset$ of edges having one endpoint at a variable node and the other at a factor node.
  Let $\neighbors(\factorvar)$ the set of variables to which a factor node $\factorvar$ is connected, then, the factor graph defines a family of distributions that factorize according to
  \begin{equation}\label{eq:factor_graph}
  \marginal(\faults\mid\syndrome) = \frac{1}{\partitionfcn} \prod_{\factorvar\in\factorset} \factorvar(\faults_{\neighbors(\factorvar)};\syndrome)
  \end{equation}
  where the normalization factor $\partitionfcn$, also known as the \emph{partition function}, 
  ensures that $\marginal(\faults)$ is a valid distribution:\footnote{The notation $\sum_\faults$ means ``sum over all possible values of $\faults$.''}
  \begin{equation}\label{eq:factor_graph_partition_function}
  \partitionfcn(\syndrome) = \sum_{\faults} \prod_{\factorvar\in\factorset} \factorvar(\faults_{\neighbors(\factorvar)};\syndrome)
  \end{equation}
    The notation $\factorvar(\faults_{\neighbors(\factorvar)};\syndrome)$ emphasizes the fact that each factor is a function of a subset $\faults_{\neighbors(\factorvar)}$ of the \fmodes $\faults$, for given observed $\syndrome$.
\end{definition}

The factor graph $F$ and the \dgraph $\dsys$ have a similar structure.
In fact we can choose the set of variables $\fgvarset$ in the factor graph to be the same as the set of variables in the \dgraph, namely the set of \fmodes.
Then, we can choose the set of factors $\factorset$ to be the \relations $\relset$ of $\dsys$, and the set of edges to be the same.
Therefore, for a given \dgraph $\dsys$, it is easy to devise the corresponding factor graph as:
\begin{equation}\label{eq:dsys_factor_graph}
  \marginal(\faults \mid \syndrome) = \frac{1}{\partitionfcn} 
    \prod_{\rel_k\in\relsetTest} \rel_k(\faults_{\scope(\test_k)};\syndromevar_k)
    \prod_{\rel_j\in\relsetPrior} \rel_j(\faultsover{\neighbors(\rel_j)}) 
\end{equation}
where we have simply observed that the probability distributions induced by the relations 
in the \dgraph naturally factorize into factors, each one corresponding to a (test-driven 
or a priori) relation in the \dgraph.

\myParagraph{Maximum a Posteriori Inference}
Given a factor graph, a natural question to ask is what is the most likely assignment of variables that maximizes the probability distribution induced by the factor graph (\eg in our case, this is the most likely set of faults in the system).
This leads to the concept of \emph{maximum a posteriori} (\MAP) inference, 
which ---given a factor graph and a syndrome $\syndrome$--- looks for the most likely variables $\faults\opt$, that maximize the
 posterior distribution:
\begin{equation*}
\tag{FG-FI}
  \faults\opt = \argmax_{\faults \in \{0,1\}^\nrfmode}\; \marginal(\faults\mid\syndrome)
\end{equation*}
Computing a \MAP estimate is known to be NP-hard for general factor graphs~\cite{shimony94ai-MAPisNPHard}, therefore it is common to use approximate methods. 
In our experiments we used belief propagation(Sec. 3 in~\cite{Nowozin11-structuredLearning}) to solve the MAP inference, which finds the optimal solution for tree-structured factor graphs, and is known to empirically return good approximations for the \MAP estimate in general factor graphs.

\myParagraph{Learning the Factor Graph Parameters}
While in the deterministic case we know the expression of the relations $\rel_k$, 
in the probabilistic case the probabilistic tests might depend on unknown parameters, \cf 
the expression in \cref{eq:noisy_or_prob_pass} that requires specifying 
the parameters $\probDetect{i}$ (probability that a fault is not detected)
and $\probFalseAlarm{i}$ (probability of a false alarm).
There are several paradigms to learn the factor graph parameters. 
In our experiments we use a method called \emph{structured support vector machine (SSVM)} or \emph{maximum margin learning} (Sec. 19.7 in~\cite{Murphy12book-MLProbabilisticPerspective}). 

%% file: sections/subsections/graph_neural_networks.tex

\subsection{Graph Neural Networks for Fault Identification}\label{sec:graph_neural_networks}

The factor graph framework introduced in the previous section learns the factor graph 
parameters from training data, and then performs maximum a posteriori inference at runtime 
for fault identification. In this section, we propose a learning-based framework that is also trained 
on a dataset, but then learns 
directly how to predict which \fmode is active at runtime.
In particular, we use \emph{Graph Neural Networks} (GNN) to learn to identify active faults in a \dgraph.

GNNs provide a general framework for learning using graph-structured data, 
and have empirically achieved state-of-the-art performance in many tasks such as node classification, link prediction, and graph classification~\cite{Hamilton20book-graphRepresentationLearning}.
The fault identification problem considered 
in this paper can be phrased as a \emph{node classification} problem.
In node classification, given a undirected graph $G=(\calV,\calE)$ where each node $i\in\calV$ has an (unknown) label $y_i$, the objective is to learn a representation vector $\ve_i$ of node $i$ such that label $y_i$ can be predicted as a function of the node embeddings $\ve_i$.

In the following, we recall common GNN architectures (\cref{sec:gnn}) 
and then we discuss how to transform our \dgraph into a structure that can be fed to a GNN 
to predict active faults (\cref{sec:dgraph_to_gnn}).


\subsubsection{Graph Neural Network Preliminaries}\label{sec:gnn}
A GNN is an extension of recurrent neural networks that operates on graph-structured data. 
GNNs are based on the concept of \emph{neural message passing} in which real-valued vector messages are exchanged between nodes of a graph ---not dissimilarly to the belief propagation we used in \cref{sec:inference_on_fg}--- but were the messages (and node updates) are built using differentiable functions  encoded as neural networks. 
To understand the basic idea of neural message passing consider an undirected graph $G=(\calV,\calE)$. 
 At the beginning, each node is assigned a feature vector $\ve_i\at{0}$ for each $i\in \calV$.
Then, during each message-passing iteration $k=1,2,\ldots$, the embedding $\ve_i\at{k}$  is updated by aggregating the embeddings of node $i$'s neighborhood $\neighbors(i)$
\begin{align}\label{eq:neural_message_passing}
  \ve_i\at{k+1} &= \update \left(\ve_i\at{k}, \aggregate \left( \left\{\ve_j\at{k} \sst j\in\neighbors(i) \right\} \right) \right) \\
   &= \update \left(\ve_i\at{k}, \va_i\at{k}\right) 
\end{align}
Where $\aggregate(\blank)$ and $\update(\blank)$ are two learned differentiable functions (\ie neural networks).
{At each iteration $k$, the $\aggregate(\blank)$ function takes the embeddings of node $i$'s neighbors and generates a \emph{message} $\va_i\at{k}$.}
Then, the $\update(\blank)$ function combines the message with the previous embedding of node $i$, generating the new  embedding of node $i$.
The final embedding is obtained by running the neural message passing for $K$ iterations.
Finally, the node label is predicted by a learned differentiable function of the node embeddings:
\beq\label{eq:gnn_prediction}
\tag{GNN-FI}
y_i = \readout(\ve_i\at{K}) 
\eeq

The literature on GNN offers a number of potential choices for the $\update(\blank)$ and $\aggregate(\blank)$ functions. We review four popular choices below.


\myParagraph{Graph Convolutional Networks (GCNs)}
One of the most popular graph neural network architectures is the graph convolutional network (GCN)~\cite{Kipf17iclr-gcn}.
The GCN model {implements the update and aggregate function as:}
\begin{equation}\label{eq:aggregate_gcn}
  \ve_i\at{k+1} = \sigma\left( \MW\at{k+1} \sum_{j\in\neighbors(i)\cup\{i\}} \frac{\ve_j\at{k}}{\sqrt{|\neighbors(i)||\neighbors(j)|}} \right)
\end{equation}
{where $\MW\at{k+1}$ is a trainable weight matrix and $\sigma(\cdot)$ is a nonlinear activation function.}
Note that \cref{eq:aggregate_gcn} can also be written in a matrix form
\begin{equation*}
  \ME\at{k+1} = \sigma \left( \hat{\MP} \ME\at{k} \MW\at{k+1} \right)
\end{equation*}
where $\hat{\MP} = \hat{\MD}^{-\frac{1}{2}} (\MA + \eye) \hat{\MD}^{-\frac{1}{2}}$, the matrix $\MA$ is the adjacency matrix of the original graph, and $\hat{\MD}$ is its diagonal degree matrix.

\myParagraph{Graph Convolutional Network via Initial residual and Identity mapping (GCNII)}
The GCN is affected by the \emph{over-smoothing} problem~\cite{li18aaai-GCNoversmoothing}, 
where 
after several iterations of GNN message passing, the nodes' embeddings become very similar to each another;
over-smoothing prevents the use of deeper GNN models, which in turn prevents the GNN from leveraging longer-term dependencies in the graph.
To solve this problem, Chen\setal~\cite{chen20icml-GCN2Conv} propose the GCNII, 
where the update of the embedding vectors becomes:
\begin{equation}\label{eq:aggregate_gcnii}
  \ME\at{k+1} = \sigma \left(
    \left( (1-\alpha_k)\hat{\MP}\ME\at{k} + \alpha_k\ME\at{0} \right)
    \left( (1-\beta_k)\eye+\beta_k\MW\at{k} \right)
  \right)
\end{equation}
and where $\alpha_k$ and $\beta_k$ are two hyper-parameters.
GCNII improves on the basic GCN by adding a smoothed representation $\hat{\MP}\ME\at{k}$ with an initial residual connection to the first layer $\ME\at{0}$, and adds an identity mapping to the {$k$-th weight matrix $\MW\at{k}$}.

\myParagraph{Graph Sample and Aggregate (GraphSAGE)}
GraphSAGE is another approach for node classification~\cite{Hamilton17nips-GraphSage}.
The aggregate function takes the form
\begin{equation}\label{eq:aggregate_graphsage}
  \va_i\at{k+1} = \sigma \left( \MW \cdot g\left( \setdef{\ve_j\at{k} }{ j\in\neighbors(i)\cup\{i\}} \right) \right)
\end{equation}
where $g(\cdot)$ is an aggregator function like the element-wise mean or max pooling.
Then, the update function is a function over the concatenation of the old embedding and the message $\va_i\at{k}$:
\begin{equation}\label{eq:update_graphsage}
  \ve_i\at{k+1} = \sigma\left( \MW [\ve_i\at{k}, \va_i\at{k+1}] \right)
\end{equation}

\myParagraph{Graph Isomorphism Network (GIN)}
The Graph Isomorphism Network (GIN)~\cite{Xu19iclr-gin}
is defined by the following aggregation function
\begin{equation}\label{eq:aggregate_gin}
  \va_i\at{k+1} = (1+\epsilon\at{k+1})\ve_i\at{k} + \sum_{j\in\neighbors(i)} \ve_j\at{k}
\end{equation}
where $\epsilon\at{k}$ is a trainable (or fixed) parameter.
The update function in GIN is 
\begin{equation}\label{eq:update_gin}
  \ve_i\at{k+1} = \zeta\at{k+1}(\va_i\at{k+1})
\end{equation}
where $\zeta(\blank)$ is also a neural network. 

\subsubsection{From \DGraphs to Graph Neural Networks} \label{sec:dgraph_to_gnn}

In order to apply GNNs to our \dgraph $\dsys$, we need to convert $\dsys=(\varset_\dsys, \relset_\dsys, \edgeset_\dsys)$ into an undirected graph $G=(\calV_G,\calE_G)$.
Towards this goal, we take the set of nodes $\calV_G$ to be \emph{both} the set of \fmodes and \dtest outcomes. Note that we add the \dtest outcomes as nodes in the graph since this allows attaching the 
test outcomes as features to these nodes.
For each test $\test_k$ we form a clique\footnote{A clique is a subset of vertices of an undirected graph such that every two distinct vertices in the clique share an edge.} involving the set of nodes in the test's scope and the variable corresponding to the test $\syndromevar_k$, namely the set $\scope(\test_k)\cup\{\syndromevar_k\}$.
We then form another clique  for each a priori relation $\rel_j\in\relsetPrior$ using the set of \fmodes $\neighbors(\rel_k)$ connected to $\rel_j$. 
For example if we have a factor $\rel(\failure_1,\failure_2;\syndromevar_2)$ we add the following (undirected) edges to $\calE_G$: $(\failure_1,\failure_2)$, $(\failure_1,\syndromevar_2)$, $(\failure_2,\syndromevar_2)$.
%
We attach a feature vector to each node in the graph. For the test nodes, we use a one-hot encoding 
describing the test outcome as node feature. For the module nodes, we use the failure probability (computed from the training data) as node features.
{We provide more details on the node features in~\cref{sec:experimental_section}. }

\begin{figure}[htbp]
  \centering
  \includegraphics[width=0.5\textwidth]{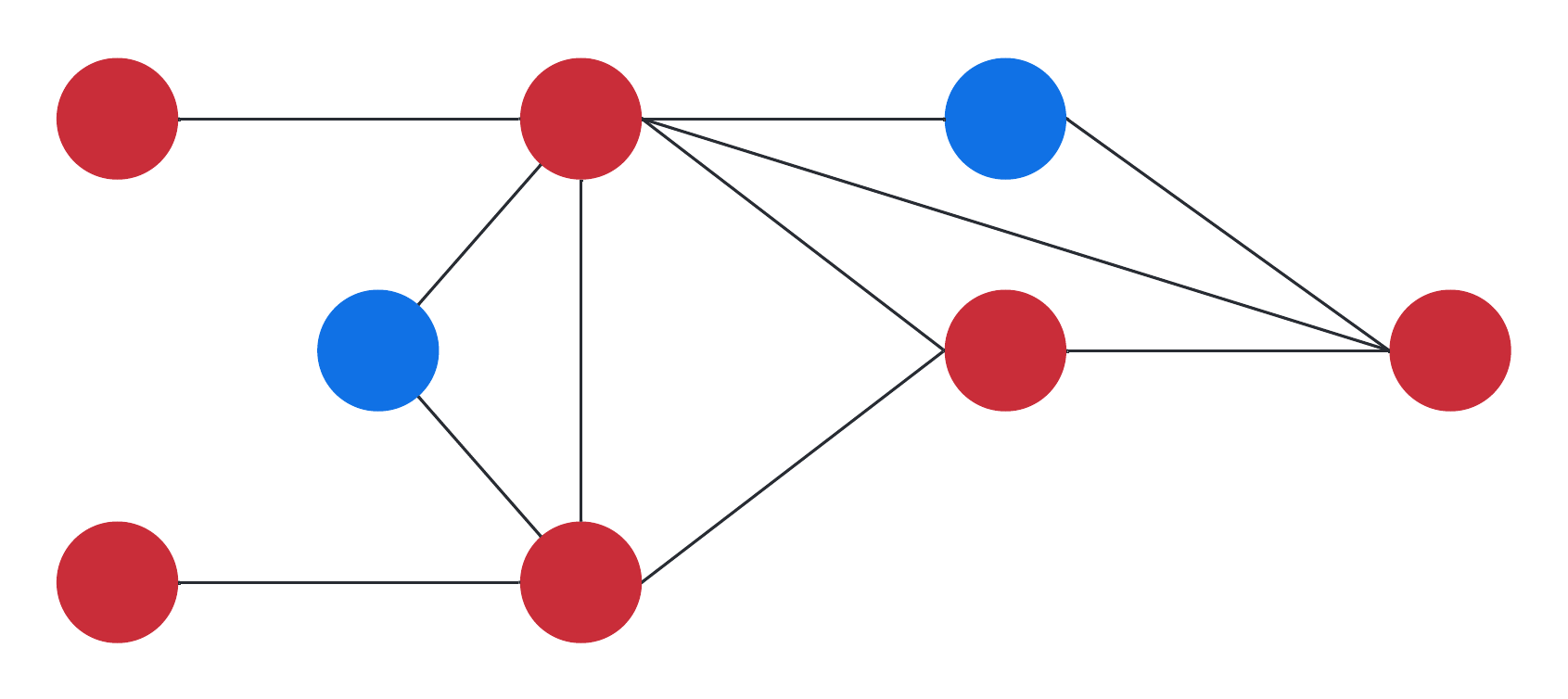}
  \caption{
    Example of conversion of the \dgraph in \cref{fig:runningExample-faults} into an undirected graph.
  }
  \label{fig:example_dsys_to_graph}
\end{figure}

\myParagraph{Learning to Identify Active Faults} 
In order to train the GNN to identify active faults, we use a supervised learning approach.
In particular, we use a softmax classification function and negative log-likelihood training loss, 
which is available in standard libraries, such as PyTorch~\cite{Paszke19neurips-pytorch}.

%% file: sections/foundamental_limits.tex

\section{Fundamental Limits}\label{sec:foundamental_limits}

Given a \dgraph~it is natural to ask if there is 
a maximum number of \fmodes~that can be correctly identified as active. 
In order words, for a given system, can we guarantee that our algorithms are able to correctly
identify all faults? Under which conditions? 
We answer these questions in this section, where we introduce the 
concept of \emph{diagnosability}. 
We discuss the deterministic case (\ie where the tests are assumed to be unreliable deterministic tests) in~\cref{sec:deterministicDiagnosability}. 
Then we obtain more general guarantees for the probabilistic case {(which also apply to our learning-based algorithms)} in~\cref{sec:probabilisticDiagnosability}. 

%% file: sections/subsections/deterministic_diagnosability.tex

\subsection{Deterministic Diagnosability}\label{sec:deterministicDiagnosability}

In this section, we assume \dgraphs with deterministic \relations and present theoretical results 
on the maximum number of faults that can be correctly identified. 
Towards this goal, we borrow and extend results from 
 fault identification in multi-processor systems~\cite{Preparata67tec-diagnosability}, which
 were partially presented in our previous work~\cite{Antonante20tr-perSysMonitoring2}.  
 In particular,~\cref{th:diagnosability} and~\cref{thm:characterization-deterministic-diagnosability} below
 are a direct application of results in~\cite{Preparata67tec-diagnosability}, while the others are our extensions.
%
%

We start with the definition of deterministic diagnosability. 

\begin{definition}[\kdiagnosability~\cite{Preparata67tec-diagnosability,Antonante20tr-perSysMonitoring2}]\label{def:diagnosability}
A \dgraph $\dsys$ is \emph{\kdiagnosable} if, given any syndrome, all active \fmodes can be correctly identified, provided that the number of active \fmodes in the system does not exceed $\dkappa$.
\end{definition}

The idea behind \kdiagnosability is that the number of failures that can be correctly identified is an intrinsic
property of a system and its \dgraph, and somehow it measures if the system has enough redundancy to unambiguously identify the cause of certain failures.

\myParagraph{Example 4: Multi-sensor Obstacle Detection (\cref{fig:runningExample} and~\cref{fig:runningExample-faults})}
Consider the example in \cref{fig:runningExample} and 
assume that an output fails if and only if the module producing it fails. 
Also assume that the sensor fusion algorithm does not necessarily fail if its inputs are wrong (thus removing $\rel_5(\failure_3,\failure_4,\failure_5)$, or setting it to be always $\ltrue$).
If both \dtests behave like Deterministic ORs, and they both return \fail, we would not know if the state of the \fmode $(\failure_1, \failure_2, \failure_3, \failure_4, \failure_5, \failure_6)$ was 
$(0,1,0,0,1,0)$, $(0,1,1,0,1,1)$, $(1,0,1,1,0,1)$, $(1,1,0,1,1,0)$ or $(1,1,1,1,1,1)$.
In fact, all these failures would generate the same syndrome (\fail,\fail). 
However, if we impose that the maximum number of active \fmode is $2$ (i.e., $\dkappa=2$), the number of feasible candidates drops to only one, namely $(0,1,0,0,1,0)$.
In other words, if we have at most two failures in the system, the two tests would allow us to uniquely identify which failure mode is active without any doubt.


After defining the notion of diagnosability in~\cref{def:diagnosability}, we are left with the question: 
can we develop an algorithm to compute the diagnosability of a certain \dgraph?
It has been noted in~\cite{Sengupta92-diagnosability} that a system is \kdiagnosable if the set of possible syndromes uniquely encodes the set of active \fmodes. Such observation is formalized by the following lemma.

\begin{lemma}[Diagnosability and Syndromes]\label{th:diagnosability}
Let $\syngen(\activeFailureModesSet)$ be the set of all possible syndromes produced by a set of active \fmodes $\activeFailureModesSet\subseteq \{1,\ldots,\nrfmode\}$.
A diagnostic graph $\dsys$ is \kdiagnosable if and only if, given any $\activeFailureModesSet_1,\activeFailureModesSet_2\subseteq \{1,\ldots,\nrfmode\}$, such that $|\activeFailureModesSet_1|,|\activeFailureModesSet_2|\leq \dkappa$ (with $\activeFailureModesSet_1 \neq \activeFailureModesSet_2$), we have $\syngen(\activeFailureModesSet_1)\cap\syngen(\activeFailureModesSet_2)=\emptyset$.
\begin{proof}
We prove ``\kdiagnosability $\Rightarrow \syngen(\activeFailureModesSet_1)\cap\syngen(\activeFailureModesSet_2)=\emptyset$'' and its reverse implication below.
In both, we define $\calX=\{\activeFailureModesSet\subseteq \{1,\ldots,\nrfmode\} \mid |\activeFailureModesSet|\leq\dkappa\}$ to be the set of subsets of $\{1,\ldots,\nrfmode\} $ of cardinality no larger than $\dkappa$.
\begin{description}[leftmargin=0cm]
  \item[$\Rightarrow$]
    Suppose $\dsys$ is \kdiagnosable. 
    Suppose by contradiction that there exists a syndrome $\syndrome$ such that $\syndrome\in\syngen(\activeFailureModesSet_1)\cap\syngen(\activeFailureModesSet_2)$, with $\activeFailureModesSet_1,\activeFailureModesSet_2\in\calX$ and $\activeFailureModesSet_1\neq \activeFailureModesSet_2$.
    Since $\syndrome\in\syngen(\activeFailureModesSet_1)$ and $\syndrome\in\syngen(\activeFailureModesSet_2)$, we are unable to say if the syndrome $\syndrome$ is produced by the set of active \fmodes~is $\activeFailureModesSet_1$ or $\activeFailureModesSet_2$, contradicting the definition of \kdiagnosability of $\dsys$.
  \item[$\Leftarrow$]
    Call $\calY = \bigcup_{\activeFailureModesSet\in\calX} \syngen(\activeFailureModesSet)$ the set of all possible syndromes assuming there are less than $\dkappa$ active \fmodes.
    From the assumptions we know that any two $\activeFailureModesSet_1,\activeFailureModesSet_2\in\calX$ have $\syngen(\activeFailureModesSet_1)\cap\syngen(\activeFailureModesSet_2)=\emptyset$, which means that we can uniquely map a syndrome to any set $\activeFailureModesSet$. 
    This is exactly the definition of \kdiagnosability.
      \vspace{-7mm}
  \end{description}
\end{proof}
\end{lemma}

The lemma intuitively establishes that for a \kdiagnosable system, two different sets of $\dkappa$ faults must produce 
different syndromes, such that for any given syndrome,
there is no ambiguity on which set of active \fmodes generated it, and we can perform fault identification without any mistake. 

\cref{th:diagnosability} suggests an algorithmic way to check if a \dgraph~is \kdiagnosable, which however 
requires checking every subset of failure modes of cardinality up to $\dkappa$ (and their syndromes).
In the following, we refine the result, showing that, under technical assumptions, one can directly compute the diagnosability by only looking at the 
topology of the \dgraph.
\begin{theorem}[Characterization of \kdiagnosability\cite{Hakimi74tc-diagnosability}]\label{thm:characterization-deterministic-diagnosability}
Let 
\begin{equation*}
  H(\failure) \doteq \{\test \mid \test\in\testset, \failure\in\scope(\test)\}
\end{equation*}
be the set of tests involving a failure mode $\failure$, and let 
\begin{equation*}
  \Gamma(\failure) \doteq \bigcup_{\test \in H(\failure)}\scope(t) \setminus \{\failure\}
\end{equation*}
be the set of failure modes that share a test with $f$. 
Also define $\Gamma(X) \doteq \bigcup_{\failure\in X} \Gamma(\failure)\setminus X$ the extension of $\Gamma$ to a set of \fmodes.
Now assume that all tests follow the Deterministic Weak-OR model and have scope of cardinality~$2$.
Then $\dsys$ is \kdiagnosable if all the following conditions are satisfied:
\begin{enumerate}[label=\roman*.]
  \item $\dkappa\leq{(\nrfmode-1)}/{2}$
  \item $\dkappa\leq\min_{i\in\failureset} |H(\failure_i)|$
  \item for each $q\in\Natural{}$ with $0 \leq q < \kappa$, and each $X\subset\failureset$ with $|X|=\nrfmode-2\kappa+q$ we have $|\Gamma(X)| > q$
\end{enumerate}
  \begin{proof}
    The assumption on the cardinality allows us to transform our general \dgraph into an undirected graph akin to the one used in~\cite{Hakimi74tc-diagnosability,Antonante21iros-perSysMonitoring2}. Then,
    the conditions \textit{(i)}, \textit{(ii)} and \textit{(iii)} are a straightforward application of Theorem 2 in~\cite{Hakimi74tc-diagnosability} to the resulting graph.  
  \end{proof}
\end{theorem}

\cref{thm:characterization-deterministic-diagnosability} also shows that the diagnosability of a system depends on the amount of redundancy in the systems and how well the tests are able to capture it.
The connection is particularly visible in condition \textit{(iii)}:
for each set of possible set $X$ of active \fmodes (of appropriate size), there must be a sufficient number the tests, that ---using information coming from different modules/outputs--- give an opinion on the state of the \fmodes in~$X$. 

Let us now move our attention to temporal \dgraphs.
Denote with $\dkappa(\dsys)$ the maximum value of $\dkappa$ for the \dgraph~$\dsys$.  
Then the following result characterizes the diagnosability of temporal \dgraphs. 
\begin{theorem}[Diagnosability in Temporal \DGraphs]\label{thm:kappa_of_composition}
  Let $\dsys\sqat{K}$ a temporal \dgraph built by stacking a set of $K$ regular \dgraphs $\dsys\at{1},\ldots,\dsys\at{K}$.
  Then $\dkappa(\dsys\sqat{K})\geq\min_{i\in\{1,\ldots,K\}}\dkappa(\dsys\at{i})$.
  \begin{proof}
    Let $\syndrome$ be a syndrome for the temporal \dgraph $\dsys\sqat{K}$, generated by a set of active failure mode $\activeFailureModesSet$, such that $|\activeFailureModesSet|=m\leq\min_{i\in\{1,\ldots,K\}}\dkappa(\dsys\at{i})$.
    Clearly, each element of $\activeFailureModesSet$ is a variable node of one of the regular \dgraphs $\dsys\at{1},\ldots,\dsys\at{K}$ that compose $\dsys\sqat{K}$, therefore we can split $\activeFailureModesSet$ into the variables nodes of each regular \dgraph, obtaining $\activeFailureModesSet\at{1},\ldots,\activeFailureModesSet\at{K}$ (these sets are non-overlapping and are such that $\cup_{i=1}^K \activeFailureModesSet\at{i} = \activeFailureModesSet$).
    Similarly, we can project the syndrome $\syndrome$ into $K$ sub-syndromes $\syndrome\at{1},\ldots,\syndrome\at{K}$ each containing only the test outcomes of  the corresponding regular \dgraphs (notice that doing the projection we lose the temporal tests, if any).
    By construction $|\activeFailureModesSet\at{i}|\;\leq m$ for each $i=1,\ldots,K$.
    From the assumption, we know that each sub-graph $\dsys\at{i}$ is $m$-diagnosable. 
    Therefore, each sub-graph $\dsys\at{i}$ will be able to correctly identify the set of active failure modes $\activeFailureModesSet\at{i}$ from the syndrome $\syndrome\at{i}$.
    This means that $\dsys\sqat{K}$ is at least $m$-diagnosable, concluding the proof.
  \end{proof}
\end{theorem}
As an immediate result we have the following corollary, which characterizes the diagnosability of ``homogeneous'' 
temporal \dgraph, obtained by stacking multiple identical \dgraphs over time.
\begin{corollary}[Diagnosability in Homogeneous Temporal \DGraphs]
  The diagnosability of the composition of identical \dgraph~is {monotonically increasing.}
\end{corollary}

This means that by stacking \dgraphs~over time, we have the opportunity to increase the diagnosability, without 
any risk of harming it.


%% file: sections/subsections/probabilistic_diagnosability.tex

\subsection{Probabilistic Diagnosability}\label{sec:probabilisticDiagnosability}

The deterministic notion of \kdiagnosability~introduced in the previous section imposes a strong condition on $\dsys$, as it requires that any syndrome unequivocally encodes all possible configurations of \fmodes.
When the tests are probabilistic, such a condition becomes too stringent: 
intuitively, 
since with some probability each test can produce different outcomes it is unlikely that~\cref{th:diagnosability} 
will be satisfied for any $\dkappa > 0$. In other words, \kdiagnosability deals with the worst case 
over all possible test outcomes, which becomes too conservative when every outcome is possible (with some probability).
For this reason, in this section, we extend the definition of diagnosability to deal with 
the case where the \dgraph includes probabilistic tests.

Towards defining a probabilistic notion of diagnosability, we introduce the 
\emph{Hamming distance} $h(\faults,\faults^\prime)$ between two binary vector $\faults$ and $\faults^\prime$ as follows:
\begin{equation}\label{eq:hamming_distance}
  h(\faults,\faults^\prime) = \sum_{i=1}^l \ind[\failure_i\neq \failure_i^\prime]
\end{equation}
where $l$ is the length of $\faults$, and $\ind$ is the indicator function.
Assuming that $\faults$ is the binary vector describing the active failures in the system, and that 
$\faults^\prime$ is an estimated vector of the fault states, the Hamming distance simply counts the number 
of \emph{mis-identified faults}.
We are now ready to introduce the following probabilistic definition of diagnosability.
\begin{definition}[(Probably Approximately Correct) PAC-Diagnosability]\label{def:pac-diagnosability}
Consider a fault identification algorithm $\Psi_\dsys$ applied to a \dgraph $\dsys$.
The \dgraph~$\dsys$ is $(\dgamma,p)$-PAC-diagnosable under $\Psi_\dsys$, if, for some $1\leq\dgamma\leq\nrfmode$
\begin{equation}
  \Pr_{(\syndrome, \faults)\sim\faultsdistr}
    \left[h\left(\Psi_\dsys(\syndrome), \faults \right)\leq \dgamma \right]\geq \dprob
\end{equation}
where $\faultsdistr$ is the joint distribution of potential failures and test outcomes.
\end{definition}
This definition simply says that a given fault identification algorithm applied to the \dgraph $\dsys$ is $(\dgamma,p)$-PAC-diagnosable if it expected to make less than $\dgamma$ mistakes with probability at least $\dprob$. 
We observe that~\cref{def:pac-diagnosability} depends on the \dgraph, but also on the fault identification algorithm.

Clearly, since the outcome of the tests is a random variable, so is the Hamming distance 
$h(\Psi_\dsys(\syndrome), \faults )$. Therefore, we can define its expected
value as:
\begin{equation*}
h_\faultsdistr(\Psi_\dsys) = \Ex_{(\syndrome, \faults)\sim\faultsdistr} \left[h(\Psi_\dsys(\syndrome), \faults)\right]
\end{equation*}
This quantity is the number of mistakes that the fault identification algorithm $\Psi_\dsys$ is expected to make.
Let us suppose we have a dataset $\dataset$ of i.i.d. samples of the underlying faults distribution $\faultsdistr$.
Let 
\begin{equation}\label{eq:empirical_hamming_loss}
  \hat{h}_\dataset(\Psi_\dsys) = \frac{1}{|\dataset|}\sum_{(\syndrome, \faults)\in \dataset} h(\Psi_\dsys(\syndrome),\faults)
\end{equation}
be the empirical number of mistakes the fault identification algorithm $\Psi_\dsys$ makes on $\dataset$. 
For instance, if we are given a (labeled) dataset $\dataset$ describing the system execution, with the  
corresponding ground truth failure modes states $\faults$, we can test our algorithm $\Psi_\dsys$ 
and calculate the empirical number of mistakes $\hat{h}_\dataset(\Psi_\dsys)$ it makes. 
Then, we can use the following result to bound the expected number of mistakes our algorithm will make in expectation over all future scenarios. 

\begin{theorem}[Fault Identification Error Bound]\label{thm:pac-diagnosability-bound}
  Consider a dataset $\dataset$ of i.i.d. samples of the underlying faults distribution $\faultsdistr$, and a fault identification algorithm $\Psi_\dsys$ over $\dsys$. 
  Then, for any $\delta > 0$, the following inequality holds with probability at least 
  $1-\delta$:
  \begin{equation}\label{eq:pac-diagnosability-bound}
    h_\faultsdistr(\Psi_\dsys) \leq \hat{h}_\dataset(\Psi_\dsys) + {\nrfmode\sqrt{\frac{\log(2/\delta)}{2|\dataset|}}}
  \end{equation}
  \begin{proof}
    For each sample $(\syndrome\at{i},\faults\at{i})$ in $\dataset$, the result of each Hamming distance will less or equal than $\nrfmode$.
    From {the Hoeffding's inequality} we have that
    \begin{equation}\label{eq:hoeffding}
      \Pr\left[|h_\faultsdistr(\Psi_\dsys) - \hat{h}_\dataset(\Psi_\dsys) | \geq \epsilon \right] \leq 
      2\mathrm{exp}\left(-\frac{2\epsilon^2|\dataset|}{\nrfmode^2}\right)
    \end{equation}
    Setting the right-hand side of \cref{eq:hoeffding} to be equal to $\delta$ and solving for $\epsilon$ yields: 
      \begin{equation}\label{eq:epsilon}
    \epsilon = {\nrfmode\sqrt{\frac{\log(2/\delta)}{2|\dataset|}}}
     \end{equation}
    After setting the right-hand side to $\delta$,~\cref{eq:hoeffding} can be rewritten as: 
    \begin{equation}
    \label{eq:temp1}
    \Pr\left[|h_\faultsdistr(\Psi_\dsys) - \hat{h}_\dataset(\Psi_\dsys) | \leq \epsilon \right] \geq \delta 
     \end{equation}
     Combining~\eqref{eq:epsilon} and~\eqref{eq:temp1} and removing the absolute value we get:
     \begin{equation}
    \Pr\left[ h_\faultsdistr(\Psi_\dsys) - \hat{h}_\dataset(\Psi_\dsys) \leq {\nrfmode\sqrt{\frac{\log(2/\delta)}{2|\dataset|}}} \right] \geq \delta 
     \end{equation}
     from which the result easily follows.    
  \end{proof}
\end{theorem}

The previous result essentially says that the expected number of mistakes the algorithm $\Psi_\dsys$ makes  stays close to the empirical mean 
$\hat{h}_\dataset(\Psi_\dsys)$, and the distance from the empirical mean gets smaller when the training dataset gets larger (\ie for larger $|\dataset|$), but gets larger for larger number of \fmodes (\ie for larger $\nrfmode$).
The following corollary easily follows.

\begin{corollary}[Characterization of PAC-diagnosability]\label{thm:corollary-pac-diagnosability-bound}
For a given dataset $\dataset$ of i.i.d. samples of the underlying faults distribution $\faultsdistr$, and a fault identification algorithm $\Psi_\dsys$ over $\dsys$, the \dgraph~$\dsys$ is $(\dgamma,\dprob)$-PAC-diagnosable with $\dprob$ satisfying the following inequality:
\begin{equation}
  \dprob \geq 1- 2e^{-2\left(\frac{\dgamma-\hat{h}_\dataset}{\nrfmode}\right)^2|\dataset|}
\end{equation}
\begin{proof}
  Let $\dgamma=h_\faultsdistr(\Psi_\dsys)$ and $\dprob=1-\delta$, substituting into \cref{eq:pac-diagnosability-bound}, and solving for $\dprob$ yield the result. 
\end{proof}
\end{corollary}


\begin{remark}[Diagnosability over Subgraphs]
Given a \dgraph~$\dsys$, we might be interested in running fault identification algorithms over a subgraph $\bar{\dsys}\subseteq\dsys$.
Analyzing the diagnosability of certain subgraphs of $\dsys$ might suggest weaknesses of the perception pipeline.
For example the system might have sufficient redundancy to be able to correctly identify the faults in the obstacle detection subgraph with low errors and high confidence, but might lack of redundancy to detect and identify faults in the traffic light recognition.
\end{remark}

%% file: sections/experiments.tex

\section{Experimental Evaluation}\label{sec:experimental_section}

This section shows that \dgraphs are an effective model to detect and identify failures in complex perception systems. 
{In particular, we show that the proposed monitors (i) outperform baselines in terms of fault identification accuracy, (ii) allow detecting failures and provide enough notice to prevent accidents in 
realistic test scenarios, and (iii) run in milliseconds, adding minimal overhead.}
A video showcasing the execution of the proposed runtime monitors can be found at \url{\videourl}.

We test our runtime monitors in several scenarios, specifically designed to stress-test the perception system.
The scenarios are simulated using the \lgsvl Simulator~\cite{lgsvl-sim}, an open-source autonomous driving simulator.
The simulator also generates ground-truth data, \eg ground-truth obstacles and active failure modes,
 and seamlessly connects to the perception system through the Cyber RT Bridge interface~\cite{lgsvl-sim}. 
 %
We apply our monitors to a state-of-the-art perception system.
In particular, 
we use 
Baidu's Apollo Auto~\cite{apollo-auto} version 7~\cite{apollo-auto-repo}.
Baidu's Apollo is an open-source, sate-of-the-art, autonomous driving stack that includes all the relevant functionalities for level 4 autonomous driving.

\cref{sec:apollo} provides more details about Apollo Auto and its perception system. 
\cref{sec:apollo_dgraph} describes the \dtests we design for Apollo Auto's perception system.
\cref{sec:implementationDetails} discusses implementation details for the proposed monitors.
\cref{sec:scenarios} describes our test scenarios.
\cref{sec:results} provides quantitative fault detection and identification results, including an ablation study of the different GNN architectures.
\cref{sec:example_scenario} provides qualitative results and discussion for a key test scenario.

\input{sections/subsections/apollo.tex}
\input{sections/subsections/apollo_diagnostic_graph.tex}
\input{sections/subsections/inference_on_apollo_dgraph.tex}

\input{sections/subsections/scenarios.tex}
\input{sections/subsections/results.tex}

\input{sections/subsections/example_scenario.tex}

%% file: sections/subsections/apollo.tex

\subsection{Apollo Auto}
\label{sec:apollo}

Baidu's Apollo Auto~\cite{apollo-auto} uses a flexible and modularized architecture for the autonomy stack based on the sense-plan-act framework.
The stack includes seven subsystems:
(i) the \emph{localization subsystem} provides the pose of the ego vehicle;
(ii) the \emph{high-definition map} provides a high-resolution map of the environment, including lanes, stop signs, and traffic signs;
(iii) the \emph{perception subsystem} processes sensory information (together with the localization data) and creates a world model;
(iv)  the \emph{prediction subsystem} predicts future evolution of the world state;
(v) the \emph{motion planning subsystems} and  (vi) the \emph{routing subsystem} generate a feasible trajectory for the ego vehicle, and finally, (vii) the \emph{control subsystem} generates low-level control signals to move the vehicle.
In our experiments, we focus on the \emph{perception subsystem}, to which we apply our runtime monitors.
In the following, we briefly review the key aspects of the Apollo Auto perception system.

\subsubsection{Apollo Auto Perception System}
Apollo Auto's perception system is tasked with the detection and classification of obstacles and traffic lights.\footnote{Note that our monitors can be also applied to other perception-related subsystems, such as the localization and high-definition map subsystem, see~\cite{Antonante21iros-perSysMonitoring2} for an example.} 
The perception module is capable of using multiple cameras, radars, and LiDARs to recognize obstacles.
There is a submodule for each sensor modality, that independently detects, classifies, and tracks obstacles.
The results from each sub-module are then fused using a probabilistic sensor fusion algorithm.

\myParagraph{Obstacle Detection}
Obstacles such as cars, trucks, bicycles, are detected using an array of radars, LiDARs, and cameras.
Each obstacle is represented by a 3D bounding-box in the world frame, the class of the object, a confidence score, together with other sensor-specific information (\eg the velocity of the obstacle). 
Each sensor is processed as follows: 
\begin{description}
  \item[Camera:] The camera-based obstacle detection network is based on the monocular object detection SMOKE~\cite{liu20arxiv-smokeDetector} and trained on the Waymo Open Dataset~\cite{Sun20-waymoDataset}.
  The network predicts 2D and 3D information about each obstacle, and then a post-processing step predicts the 3D bounding box of each obstacle by minimizing the reprojection error of available templates for the predicted obstacle class;
  \item[LiDAR:] The LiDAR-based obstacle detection network, called Mask-Pillars is based on PointPillars~\cite{lang19cvpr-pointpillars}, but enhanced with a residual attention module to improve detection in case of occlusion;
  \item[Radar:] Apollo Auto uses directly the obstacles detections reported by the radar (assumed to have an embedded detector~\cite{ContiRadar}), that are post-processed to be transformed to the world frame.
\end{description}




\subsubsection{Vehicle Configuration}
The simulated vehicle is a Lincoln MKZ with one Velodyne VLS-128 LiDAR, one front-facing camera with a field-of-view of $50^{\circ}$, one front-facing telephoto camera (pointed $4^\circ$ upwards) for traffic light detection and recognition, one Continental ARS 408-21 front-facing radar, GPS, and IMU.

\input{sections/figures/vehicle.tex}

We ran the Baidu's Apollo AV stack on a computer with an Intel i9-9820X (\SI{4.1}{\giga\hertz}) processor, \SI{64}{\giga\byte} of memory and two NVIDIA GeForce RTX 2080Ti.
The simulator ran on a computer with 11th Generation Intel i7-11700F (\SI{4.8}{\giga\hertz}) processor, \SI{16}{\giga\byte} of memory, and an NVIDIA GeForce RTX 3060.
The two computers were connected using a Gigabit Ethernet cable.

%% file: sections/figures/vehicle.tex
\begin{figure}[!tbp]
  \begin{center}
  \subfloat 
  {
    \includegraphics[width=0.5\textwidth]{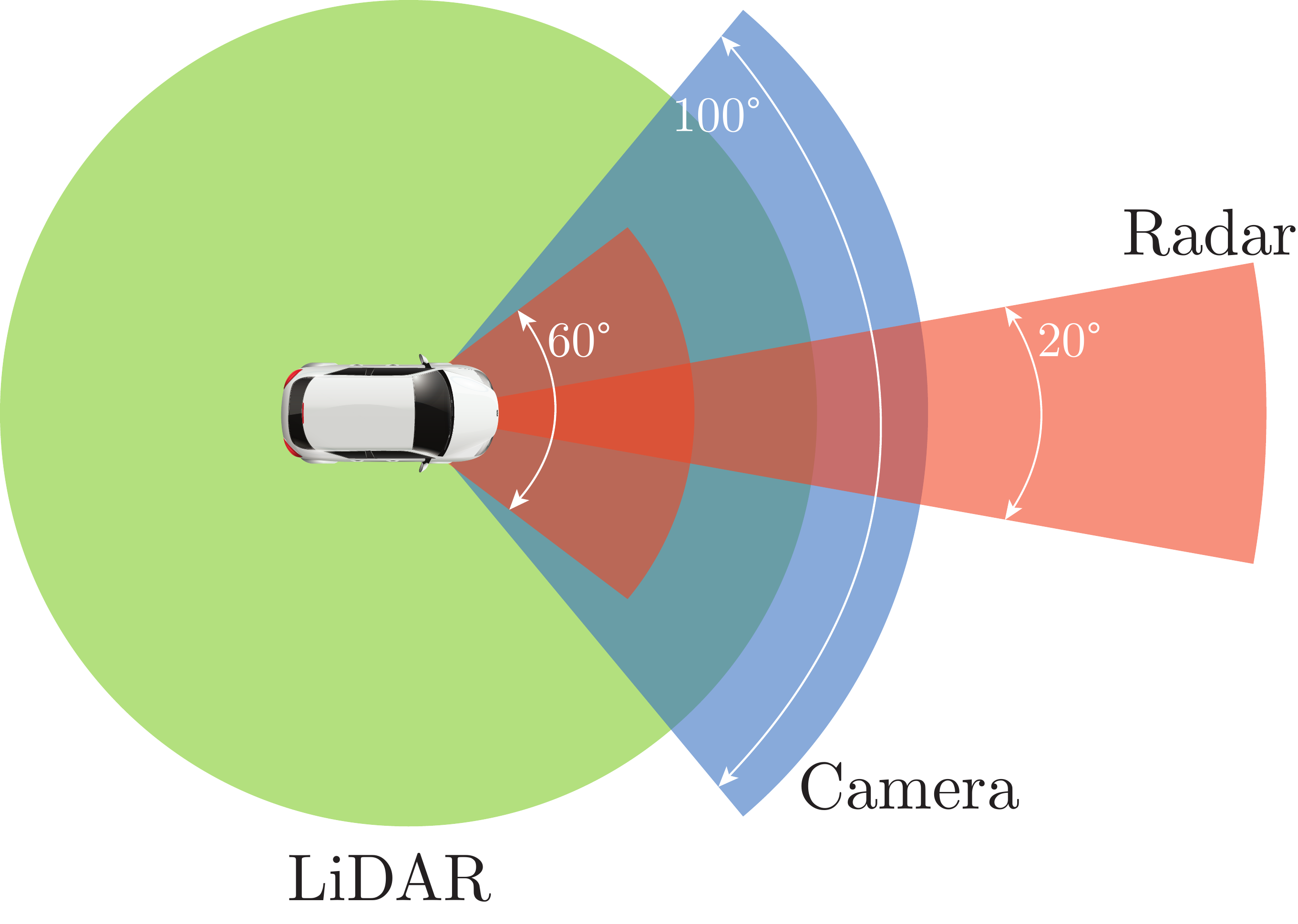}
  }~
  \subfloat 
  {
    \begin{tikzpicture}
      \node at (0,0) {\includegraphics[width=0.5\textwidth]{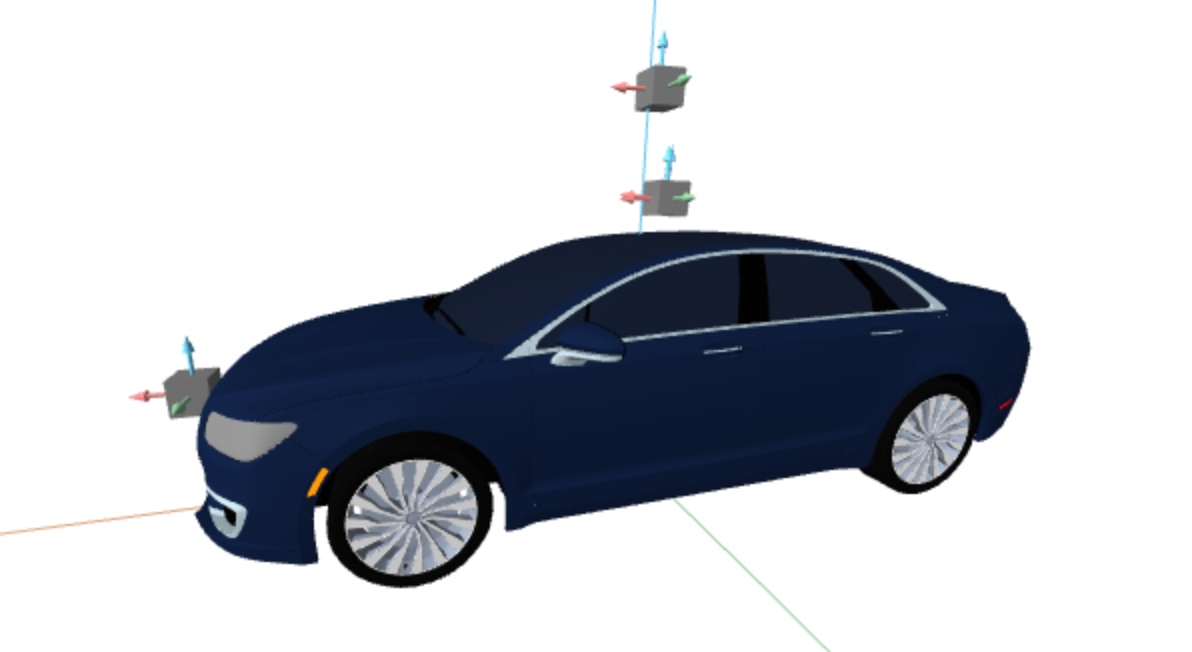}};
      \node[text width=3cm] at (2.1, 1.3) {LiDAR};
      \node[text width=3cm] at (2.1, 0.7) {Camera};
      \node[text width=3cm] at (-1, 0.3) {Radar};
    \end{tikzpicture}
  }
  \caption{
    Vehicle configuration and sensor field-of-view (FOV). LiDAR FOV is shown in green, 
  the camera FOV in blue and the radar FOV in orange. \label{fig:apollo_sensor_fov}
  \label{fig:apollo_sensor_fov}}
  \end{center}
\end{figure}

%% file: sections/subsections/apollo_diagnostic_graph.tex

\subsection{\DGraph}\label{sec:apollo_dgraph}

\begin{figure}[!ht]
  \centering
  \includegraphics[width=\textwidth]{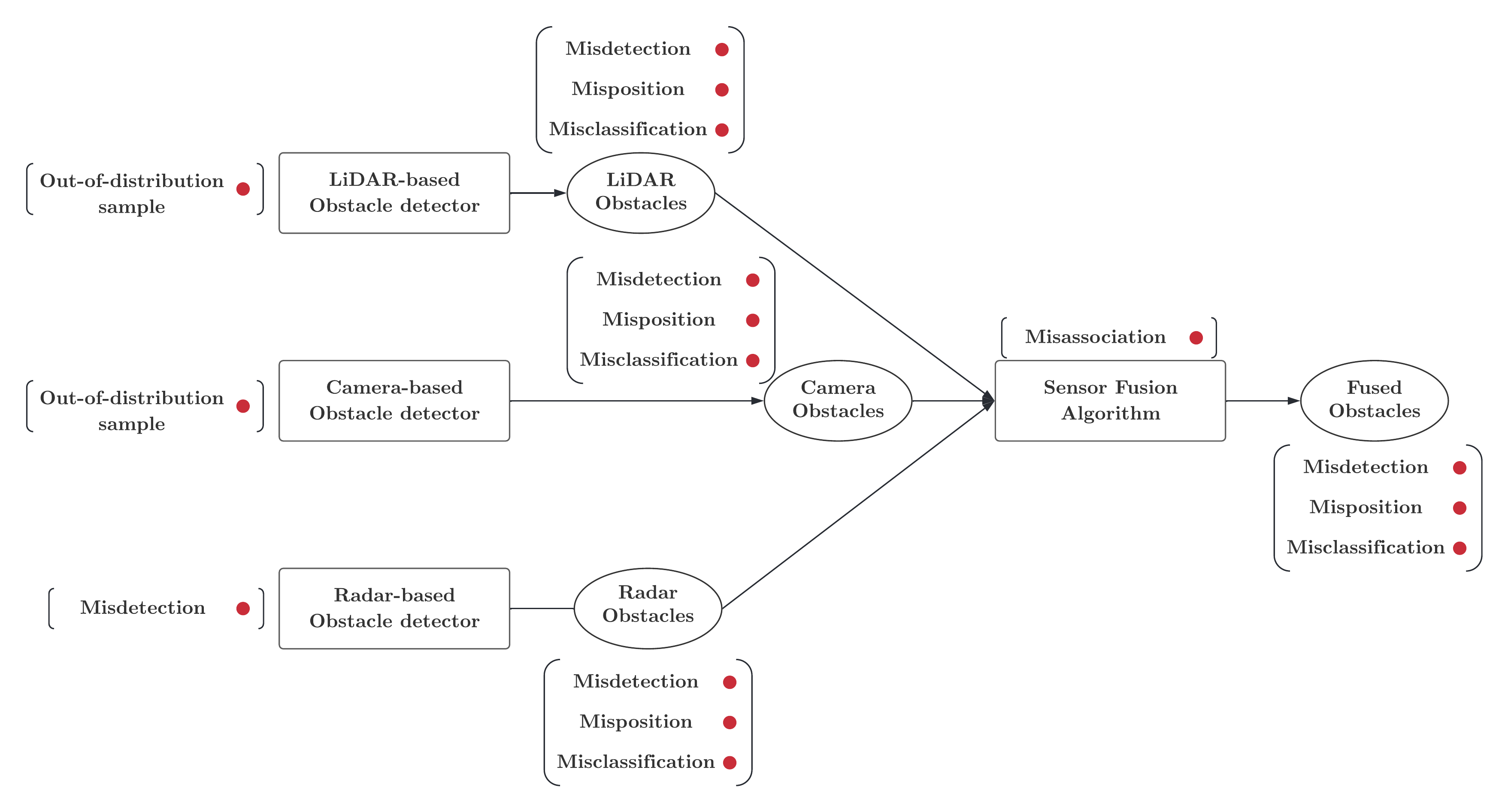}
  \caption{Perception system considered in our experiments. Modules are shown as rectangular blocks, outputs are shown as rounded boxes, while \fmodes are denoted with red dots.}
  \label{fig:apollo_diagnostic_graph}
\end{figure}

We focused our attention on the obstacle detection pipeline.
The system we aim to monitor, together with the failure modes considered, is shown in~\cref{fig:apollo_diagnostic_graph}
The system is composed of four modules: 
\begin{itemize}
  \item Lidar-based Obstacle detector, based on a deep learning algorithm, subject to \emph{out-of-distribution sample} \fmode;
  \item Camera-based Obstacle detector, based on a deep learning algorithm, subject to \emph{out-of-distribution sample} \fmode; 
  \item Radar-based Obstacle detector subject to \emph{misdetection} \fmode; 
  \item Sensor Fusion subject to \emph{misassociation} \fmode. 
\end{itemize}
Each module produces a set of detected obstacles.
We identified three \fmodes for each set of detected obstacles:
\begin{itemize}
  \item \emph{misdetection}: the module detected a ghost obstacle or is missing an obstacle in the scene;
  \item \emph{misposition}: the module detected the obstacle correctly, but its position is incorrect (\ie more than 2.5m error in our tests);
  \item \emph{misclassification}: the module detected the obstacle correctly but the obstacle's semantic class is incorrect.
\end{itemize}
We equipped the obstacle detection system with 18 \dtests.
For each pair of modules' outputs, namely (Lidar, Camera), (Radar, Camera), (Lidar, Sensor Fusion), (Radar, Sensor Fusion), (Lidar, Radar), and (Camera, Sensor Fusion), there is a test that compares the outputs to diagnose each of the output's \fmodes (\ie misdetection, misposition, and misclassification).
Intuitively, each test compares the two sets of obstacles coming from the corresponding modules, and if they are different, it reports if the inconsistency was due to a misdetection, misposition, or misclassification.
Moreover, we included a priori relation between every module and its output.
In particular, the modules are assumed to fail if their outputs have at least one active \fmode.
In the probabilistic \dgraph we also added an a priori relation for each module's \fmode, indicating the prior probability of that \fmode being active.

\subsubsection{\DTests}\label{sec:apollo_diagnostic_test}
We now describe the logic for the \dtests we implemented.
Consider two sets of synchronized detected obstacles\footnote{By synchronized we mean that the two outputs are produced at the same time instant.}, say $\lobs$ and $\robs$, produced by two modules, using some sensor data.
Let $\Omega$ be the region defined by the intersection of both sensor fields of view and a region of interest (\eg a region close to a drivable area\footnote{In our experiments, the region of interest is the area within $5$ meters  from a drivable lane.}).
Denote by $\lobs_\Omega$ and $\robs_\Omega$ the set of obstacles restricted to the region $\Omega$, namely $\lobs_\Omega\subseteq\lobs$ such that for each obstacle $o$ in $\lobs$, $o$ is in $\lobs_\Omega$ if and only if $o$ is inside the region defined by $\Omega$.
The same relation holds for $\robs_\Omega$.
Then the \dtest checking for misdetections is defined as follows:
\begin{equation*}
\test_{\mathrm{misdetection}} = 
\begin{cases}
  \mathrm{\fail} &\quad\mathrm{if } \left|\lobs_\Omega\right|\neq\left|\robs_\Omega\right|\\
  \mathrm{\pass} &\quad\mathrm{otherwise} \\
\end{cases}
\end{equation*}
Note that if the two sets of obstacles have a different cardinality ---when restricted to the area co-visible by both sensors--- it means that one of the two sets contains a ghost obstacle or one of the two sets is missing an obstacle.
From a single test, we are not able to say which of the two sets is experiencing the misdetection, but we know at least one output did.

Let us now move our attention to the misposition \fmode.
Let $\matches$ be the set of \emph{matched} obstacles, that is, a pair of obstacles $(l,r)$ ---with $l\in\lobs_\Omega$ and $r\in\robs_\Omega$--- is in $\matches$, if $l$ and $r$ represent the same obstacles.
A common approach for finding the set of matches is to select all the pairs that are closest to each other (\ie solving an assignment problem)\footnote{We matched obstacles using a generalization of the Hungarian algorithm~\cite{Crouse16taes-linearAssignment}, with the cost of each match being the Euclidean distance between obstacles. }.
The \dtest checking for mispositioned obstacles is defined as follows:
\begin{equation*}
\test_{\mathrm{misposition}} =
\begin{cases}
  \mathrm{\fail} &\quad\exists (l,r)\in\matches \mbox{ such that } |\pos(l)-\pos(r)|\geq\theta\\
  \mathrm{\pass} &\quad\mathrm{otherwise} \\
\end{cases}
\end{equation*}
where $\pos(\blank)$ is the position of an obstacle and $\theta$ is an error threshold, {chosen as $\theta = \SI{2.5}{\meter}$ in our experiments.}

Finally, the test checking for misclassified obstacles is defined as follows:
\begin{equation*}
\test_{\mathrm{misclassification}} =
\begin{cases}
  \mathrm{\fail} &\quad\exists (l,r)\in\matches \text{ such that } \cls(l)\neq\cls(r)\\
  \mathrm{\pass} &\quad\mathrm{otherwise} \\
\end{cases}
\end{equation*}
where $\cls(\blank)$ is the class of the obstacle, \ie the test fails if associated obstacles are assigned 
different semantic classes.

\subsubsection{Temporal \DGraph}

To build a temporal \dgraph we stack $2$ regular \dgraphs into a temporal \dgraph.
In the probabilistic case, 
each module \fmode is connected to its successive (in time) via a priori relationships, which represent the transition probability {between states in consecutive time steps}. 
No temporal a priori relations are added in the deterministic case.
We also added temporal tests.
{The logic of the tests presented in~\cref{sec:apollo_dgraph} is applicable to temporal tests with small changes.}
In temporal tests, the sets $\lobs$ and $\robs$ are not time-synchronized anymore (\eg they are obstacles detected by 
the same sensor at consecutive time stamps), therefore the position of each obstacle in each set must be adjusted for the distance the obstacle traveled between consecutive detections.
To use the tests described earlier in the temporal domain we used the following approach.
If the obstacle is equipped with an estimated velocity vector, since the time difference between detections is usually below \SI{30}{\milli\second}, we assume constant speed and integrate the speed over the time interval to find an approximate position of each obstacle.
When the velocity is not available, we use the average speed of an obstacle (for a given obstacle's class) and adapt the misposition threshold $\theta$ to account for the uncertainty.


%% file: sections/subsections/inference_on_apollo_dgraph.tex

\subsection{Fault Identification: Implementation Details}
\label{sec:implementationDetails}

\myParagraph{Deterministic Fault Identification}\label{sec:deterministic_model}
For the tests with the deterministic model, we assumed the \WeakerOR model for the \dtests as described in \cref{eq:deterministic_blind_or}.
We used this model for both the regular \dgraph and the temporal \dgraph, and solved the optimization problem in~\cref{eq:optimization_weakerOR} using Google OR-Tools~\cite{google-ortools} Integer Programming Solver.

\myParagraph{Probabilistic Fault Identification}
To perform probabilistic inference on the \dgraph, we transformed it into a factor graph and trained the potentials for each relation using the maximum margin learning algorithm described in~\cref{sec:inference_on_fg} on the training dataset.
We used the Hamming distance defined in \cref{eq:hamming_distance} as the loss function $\loss$.
We set the regularization parameter to $\lambda=10$; see~\cite{Nowozin11-structuredLearning}.\footnote{In our experiment we noticed that the performance of the learning algorithm are not sensitive to the choice of $\lambda$.}
For each \dgraph, we perform inference using 
 the max-product algorithm 
 for a fixed number of iteration ($100$ iterations). 
 In our implementation, we use the \emph{Grante} library~\cite{GranteLibrary} to perform learning and inference over the 
 factor graph.

\myParagraph{Graph-Neural-Network-based Fault Identification}
In \cref{sec:graph_neural_networks} we saw that a graph neural network requires a feature for each node in the graph to perform neural message passing.
We now discuss how we set the feature vector for each node in the graph.
Recall that the GNN uses a pairwise undirected graph, where a node is either a \fmode or a test outcome.
The feature $\vxx_{\test_k}\in\Reals{2}$ for a test $\test_k$ is set as the one-hot encoding of the test outcome (\ie $[1\; 0]$ if the test passed, $[0\;1]$ if it failed).
For the \fmode nodes we do not have any measurable quantity at runtime; we therefore use the training dataset to compute the feature vectors.
In particular the feature vector $\vxx_{\failure_i}\in\Reals{2}$ for a \fmode $\failure_i$ is computed as follow:
let $\rho_i$ be the empirical probability that $\failure_i$ is \factive, \ie 
 $\rho_i=\frac{1}{|\dataset|} \sum_{(\syndrome,\faults)\in \dataset}\ind[\failure_i = \mathfactive]$;
then the feature vector is chosen as $\vxx_{\failure_i} = [1-\rho_i,\rho_i]^\tran$.
Intuitively, the feature describes the prior probability of the \fmode $\failure_i$'s state. 

We now discuss the architecture of the GNN.
Our GNN is composed by a linear layer that embeds the feature vectors in $\Reals{16}$, followed by a ReLU function.
The output is then passed to a stack of graph convolution layers interleaved with ReLU activation functions.
We tested four different graph convolution layers
\begin{itemize}
  \item in the case of GCN, we stack $3$ layers with $16$ hidden channels each;
  \item in the case of GCNII, we stack $64$ layers with $16$ hidden channels each with $\alpha = 0.1$, $\beta = 0.4$;
  \item in the case of GIN, we stack $3$ layers with $16$ hidden channels each with the function $\zeta\at{k}(\blank)$ (\cf~\cref{eq:update_gin}) being a 2-layer perceptron for $k=1,\ldots,3$;
  \item in the case of GraphSAGE, we stack $3$ (and $6$ for temporal \dgraphs) layers with mean aggregator and $16$ hidden channels each.
\end{itemize}
Finally, the readout function that converts the graph embedding to node labels is a linear layers followed by a softmax pooling. We perform an ablation of the different GNN architectures in~\cref{sec:results}.

We implemented the GNNs in PyTorch~\cite{Paszke19neurips-pytorch} and 
trained them on the training dataset for $100$ epochs using the Adam optimizer.
To reduce the amount of guesswork in choosing an initial learning rate, we used the learning rate finder available in the PyTorch Lightning library~\cite{Falcon19-PyTorchLightning}.
The procedure is based on~\cite{Smith2017wacv-cyclicalLearningRate}: the learning rate finder does a small training run where the learning rate is increased after each processed batch and the corresponding loss is logged.
Then, the learning rate is chosen to be the point with the steepest negative gradient.

\myParagraph{Baselines}
We compared the proposed monitors against two simple baselines. 
In the first baseline (label: ``Baseline''), whenever a \dtest returns \fail, all \fmodes in its scope are considered active.
In the second baseline (label: ``Baseline (w/rel. scores)''), modules are ordered by a reliability score defined by the system designer.
In our experiments we considered the radar to be more reliable than the sensor fusion, which  is more reliable than the LiDAR, which in turn is more reliable than the camera.
When a \dtest fails, this second baseline  labels all the \fmodes in the test scope associated to the least reliable module (and its outputs) as \factive.
For example if a \dtest comparing camera and LiDAR obstacles returns \fail, the \fmodes associated with the camera are the ones that are labeled active because the camera is considered less reliable than the LiDAR.
Both baselines label a module' \fmodes as active if at least one of the module's outputs is failing.

%% file: sections/subsections/scenarios.tex

\subsection{Scenarios}\label{sec:scenarios}

We designed a set of challenging scenarios 
to stress-test the Apollo Auto perception system. 
These scenarios were created using the LGSVL Simulator Visual Scenario Editor, 
which allows the user to create scenarios using a drag-and-drop interface.
The vehicle behavior is tested on each scenario in a multitude of situations including different time of day (noon, 6 PM, 9 PM) or weather condition (rain and fog).
The scenarios are described in~\cref{tab:scenarios}.

\input{tables/scenarios_ext2.tex}

\subsubsection{Dataset generation}\label{sec:dataset}

We executed the \dtests described in~\cref{sec:apollo_dgraph} every $0.3$ seconds, and used the corresponding 
test outcomes to perform fault identification. 
Time synchronization of the modules' output is achieved by pairing outputs that are closest in time to each other.
Ground-truth labels for the outputs' \fmodes~are generated using the ground-truth detections provided by the simulator.
In particular, to generate the label for each \fmode of an output, we used the three \dtests described in~\cref{sec:apollo_diagnostic_test}
comparing the set of obstacles to the ground-truth detections.
For a module $\modulevar$ instead, since all modules have only one \fmode, the associated failure mode $\failure_{\modulevar}$ is  labeled as \factive~if and only if any failure mode if its output is \factive.
We collected $1650$ regular \dgraphs, and split them into $1320$ ($80\%$) training samples, $165$ ($10\%$) testing samples, and $165$ validation samples.
We also collected $1590$ temporal \dgraphs, and split them into $1272$ ($80\%$) training samples, $159$ ($10\%$) testing samples, and $159$ validation samples.

%% file: tables/scenarios_ext2.tex

\setlength\LTleft{-1cm}
\begin{longtable}{ p{6.5cm} p{6.5cm} }
  \caption{
    Scenarios. 
    \fcolorbox{black}{FaultFreeColor}{\rule{0pt}{4pt}\rule{4pt}{0pt}}\quad Fault-free.
    \fcolorbox{black}{FaultColor}{\rule{0pt}{4pt}\rule{4pt}{0pt}}\quad Fault-detected.
  }\label{tab:scenarios}\\
  \textbf{Scene} & \textbf{Fault Detection Results} \\
  \midrule 
  \includegraphics[width=6cm]{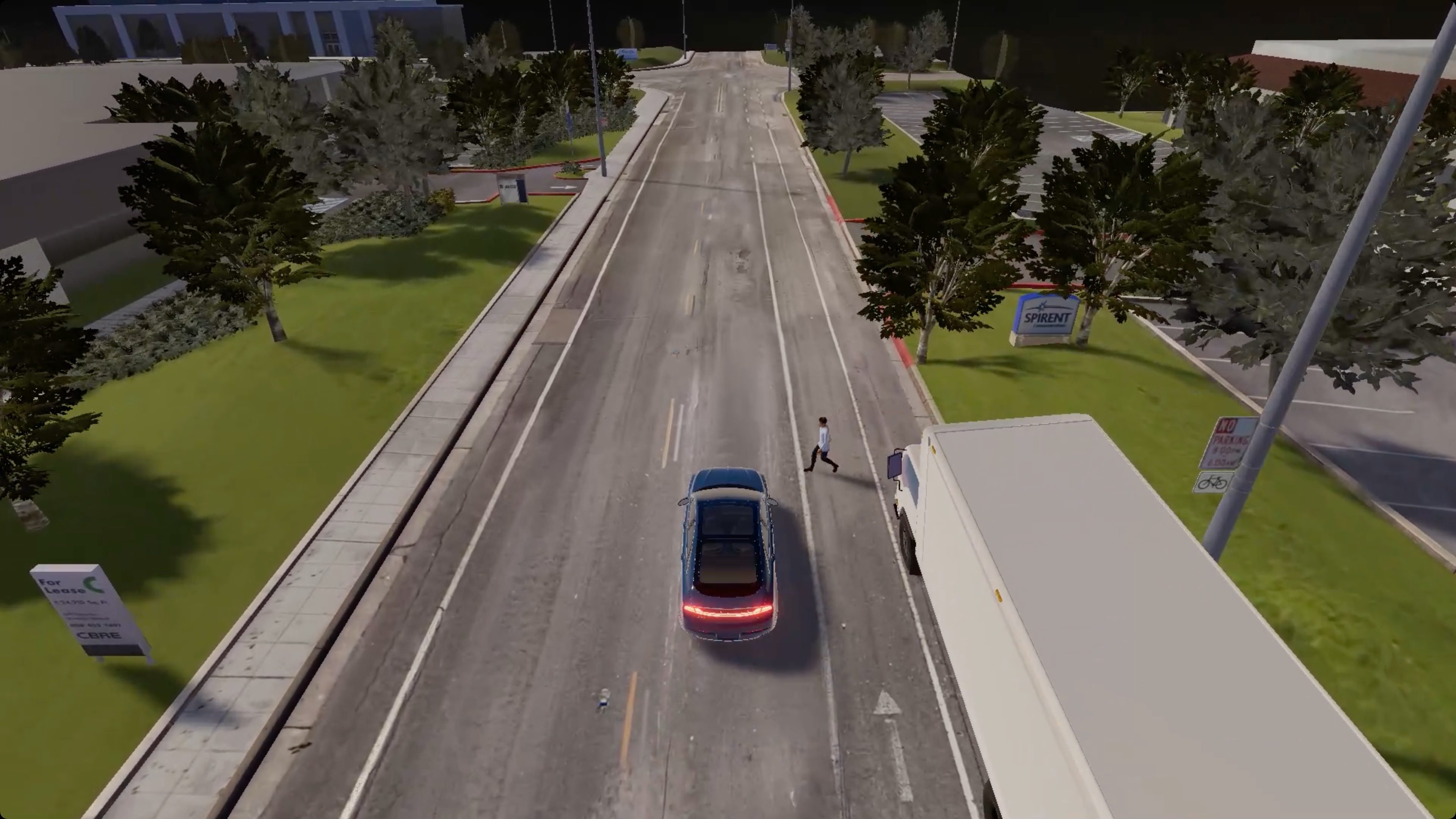}
  &
  \begin{tikzpicture}[
    path image/.style={path picture={
      \node at (1,3) {\includegraphics[scale=0.8, rotate=105]{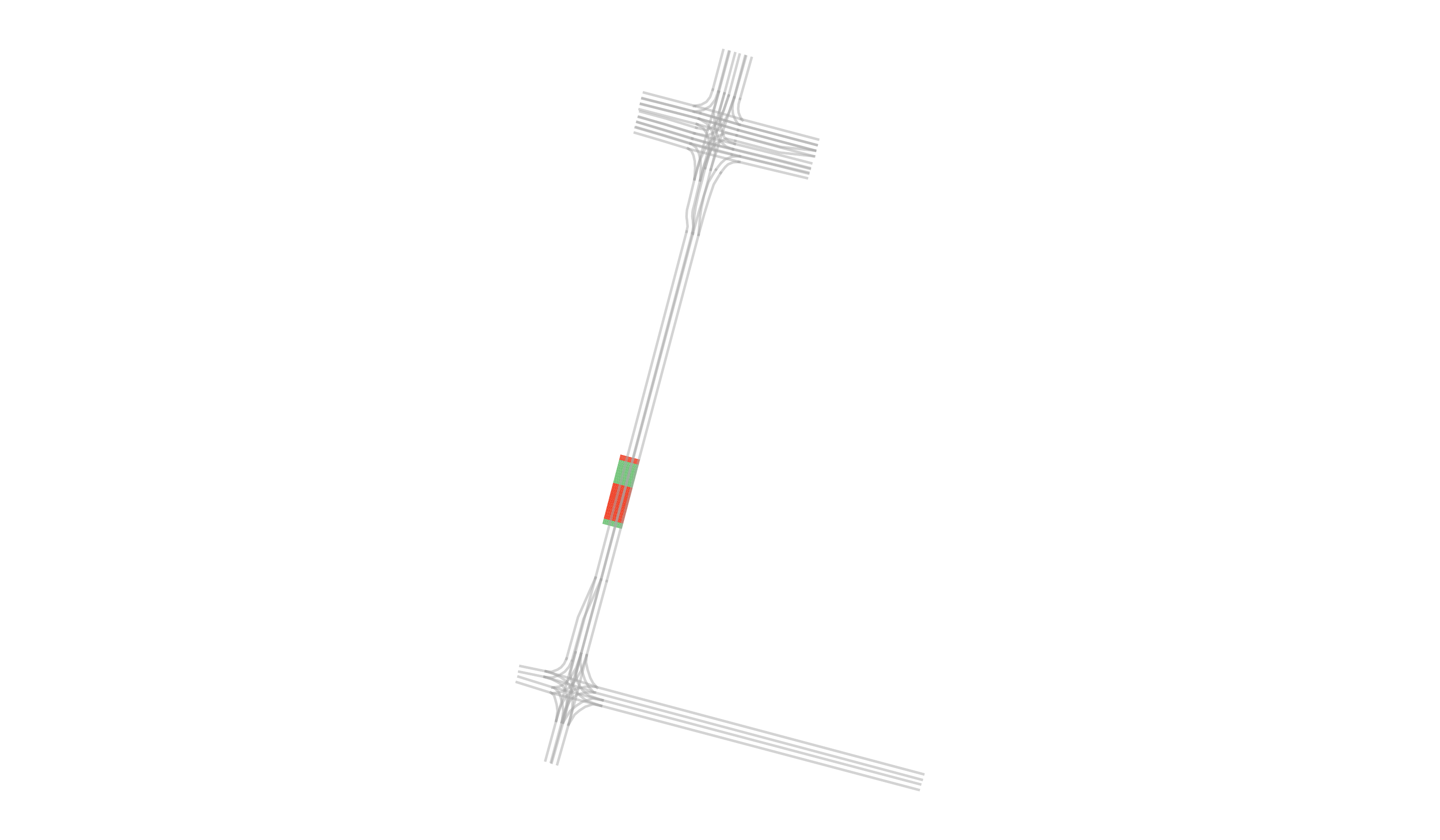}};}
    }]
    \path [path image] (0,0) rectangle (\linewidth, 3cm);
  \end{tikzpicture}  
  \\
  \multicolumn{2}{p{13cm}}{
    \textbf{Hidden Pedestrian}.
    A pedestrian, initially occluded by a track parked on the right-hand side of the street, steps in front of the ego vehicle.
  }\\
  \midrule 
  \includegraphics[width=6cm]{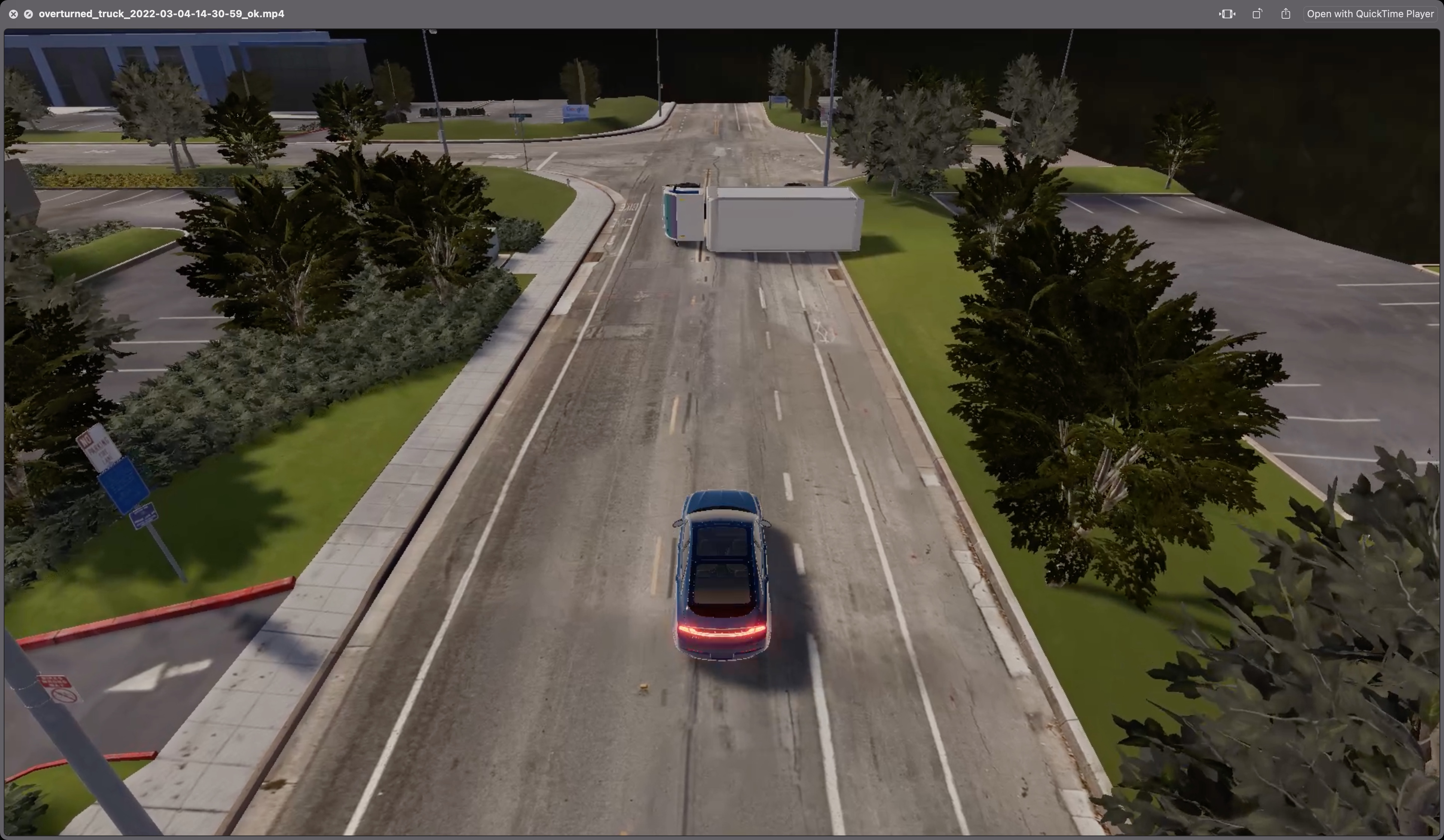}
  &
  \begin{tikzpicture}[
    path image/.style={path picture={
      \node at (2.2,3) {\includegraphics[scale=0.4, rotate=105]{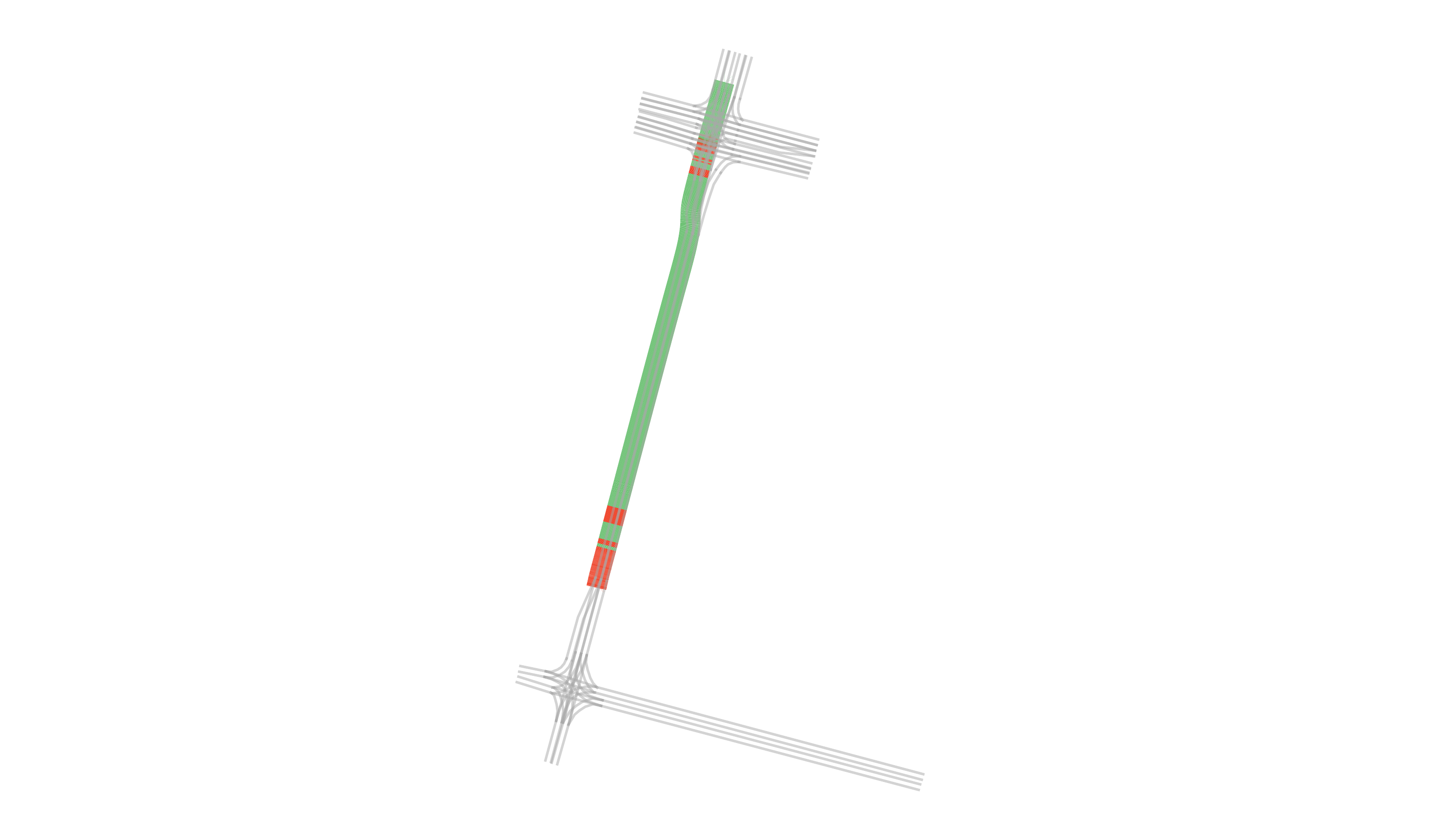}};}
    }]
    \path [path image] (0,0) rectangle (\linewidth, 3cm);
  \end{tikzpicture}
  \\
  \multicolumn{2}{p{13cm}}{
    \textbf{Overturned Truck}.
    The ego vehicle encounters an overturned truck occupying the lane it is driving in.
    The scenario recreates an accident occurred in Taiwan where a Tesla hit an overturned truck on a highway~\cite{teslaAccident}.
  }\\
  \midrule 
  \includegraphics[width=6cm]{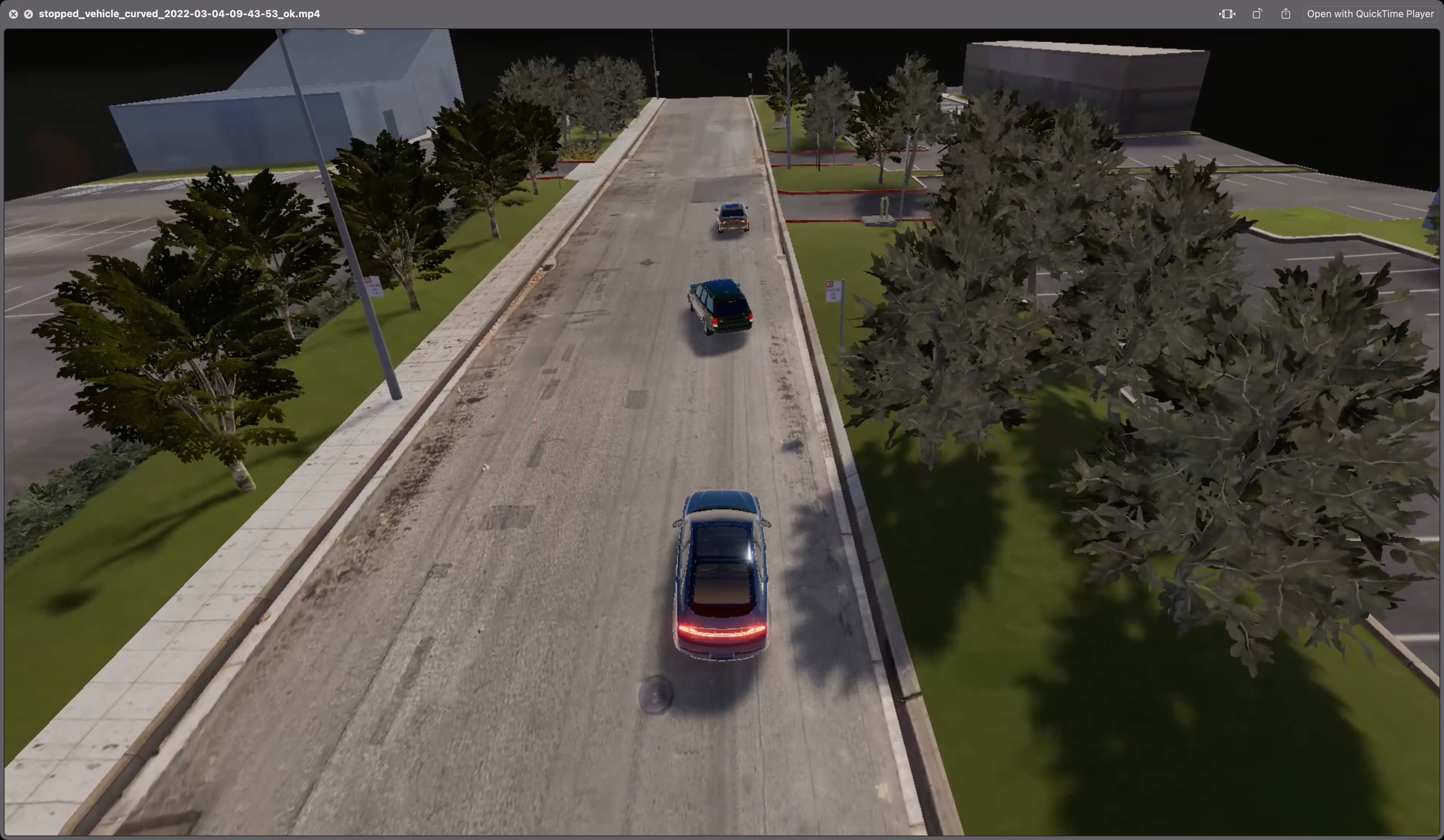}
  &
  \begin{tikzpicture}[
    path image/.style={path picture={
      \node at (2.5,7) {\includegraphics[scale=0.8, rotate=15]{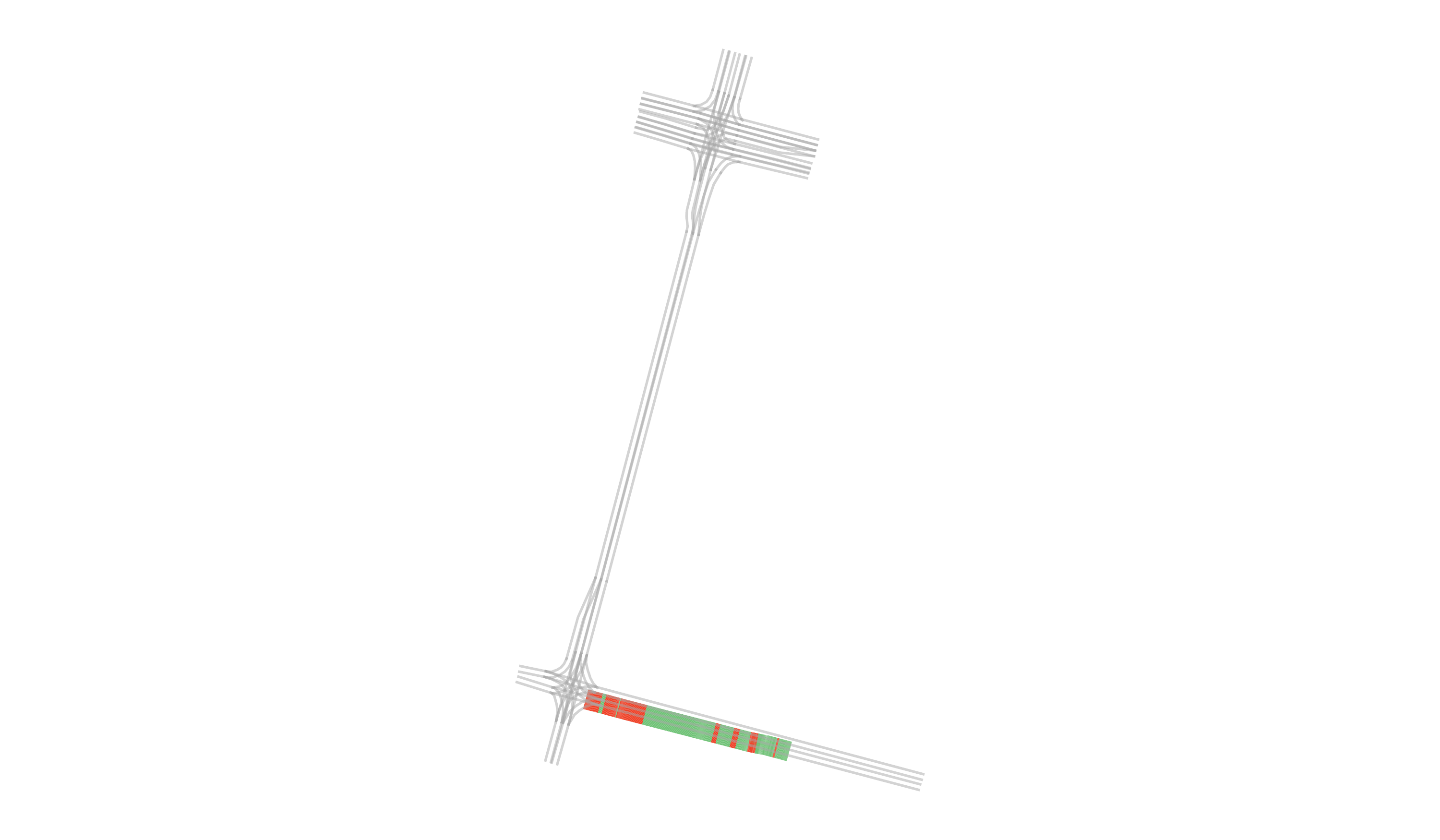}};}
    }]
    \path [path image] (0,0) rectangle (\linewidth, 3cm);
  \end{tikzpicture}
  \\
  \multicolumn{2}{p{13cm}}{
    \textbf{Stopped Vehicle}.
    While driving, the car in front of the ego vehicle makes a lane change to avoid the stationary car that is in their lane. 
    This leaves the ego vehicle with little to no time to react to the stationary car.
  }\\
  \midrule 
  \includegraphics[width=6cm]{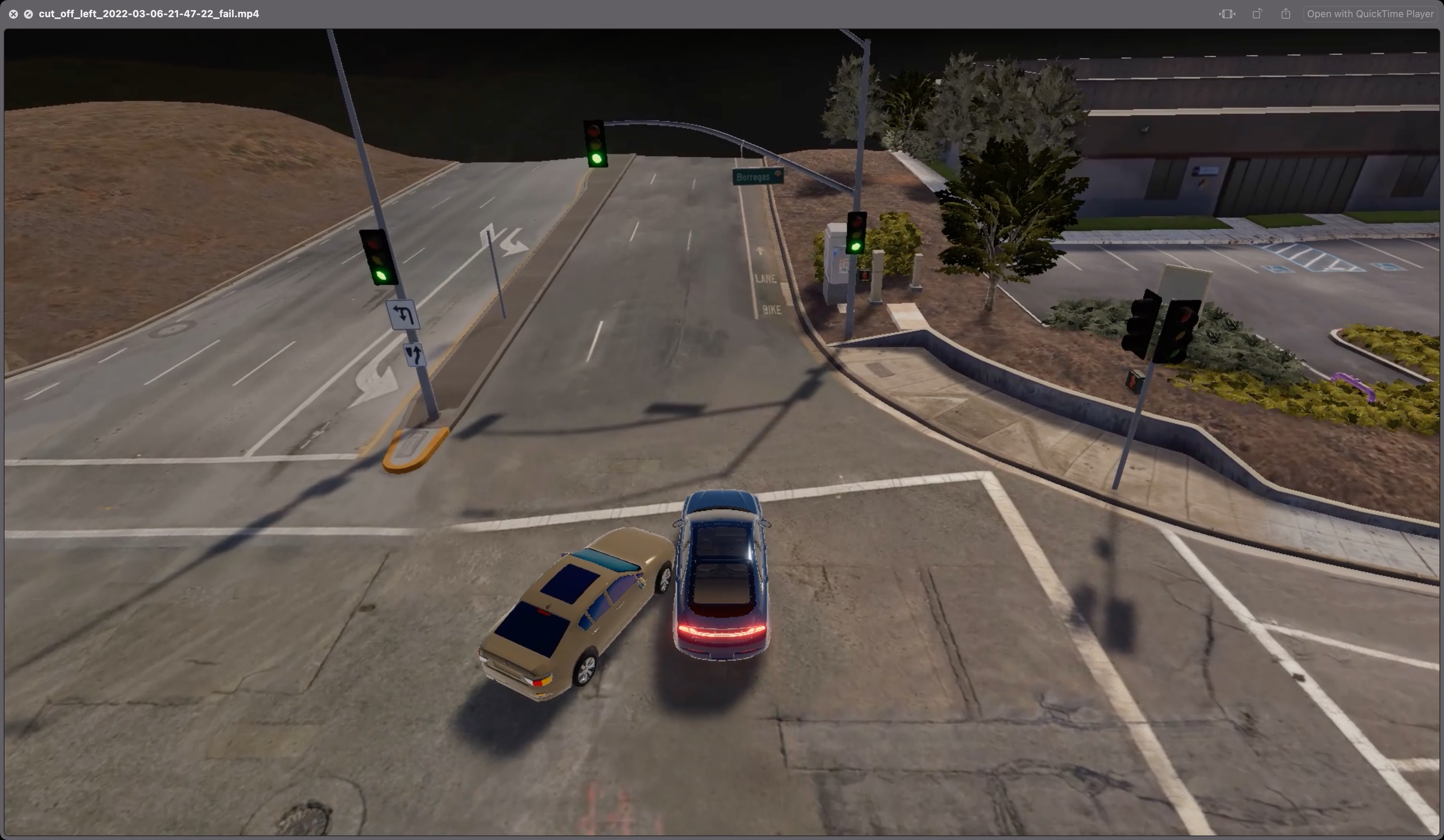}
  &
  \begin{tikzpicture}[
    path image/.style={path picture={
      \node at (6,4) {\includegraphics[scale=0.8, rotate=105]{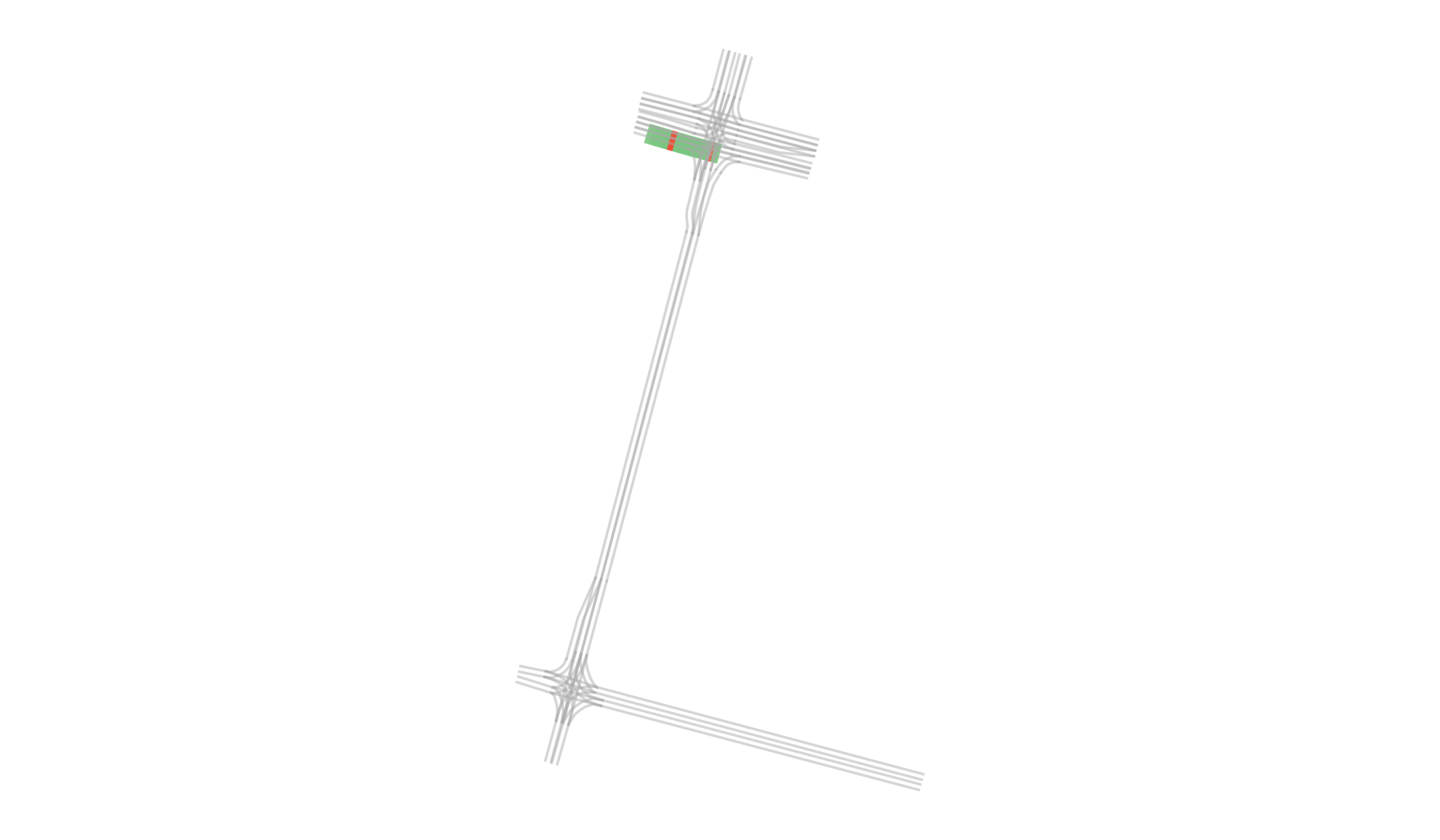}};}
    }]
    \path [path image] (0,0) rectangle (\linewidth, 3cm);
  \end{tikzpicture}
  \\
  \multicolumn{2}{p{13cm}}{
    \textbf{Cut Off Left}.
    While driving in the right lane on a three-lane road, a vehicle from the left lane cuts the ego vehicle off.
  }\\
  \midrule 
  \includegraphics[width=6cm]{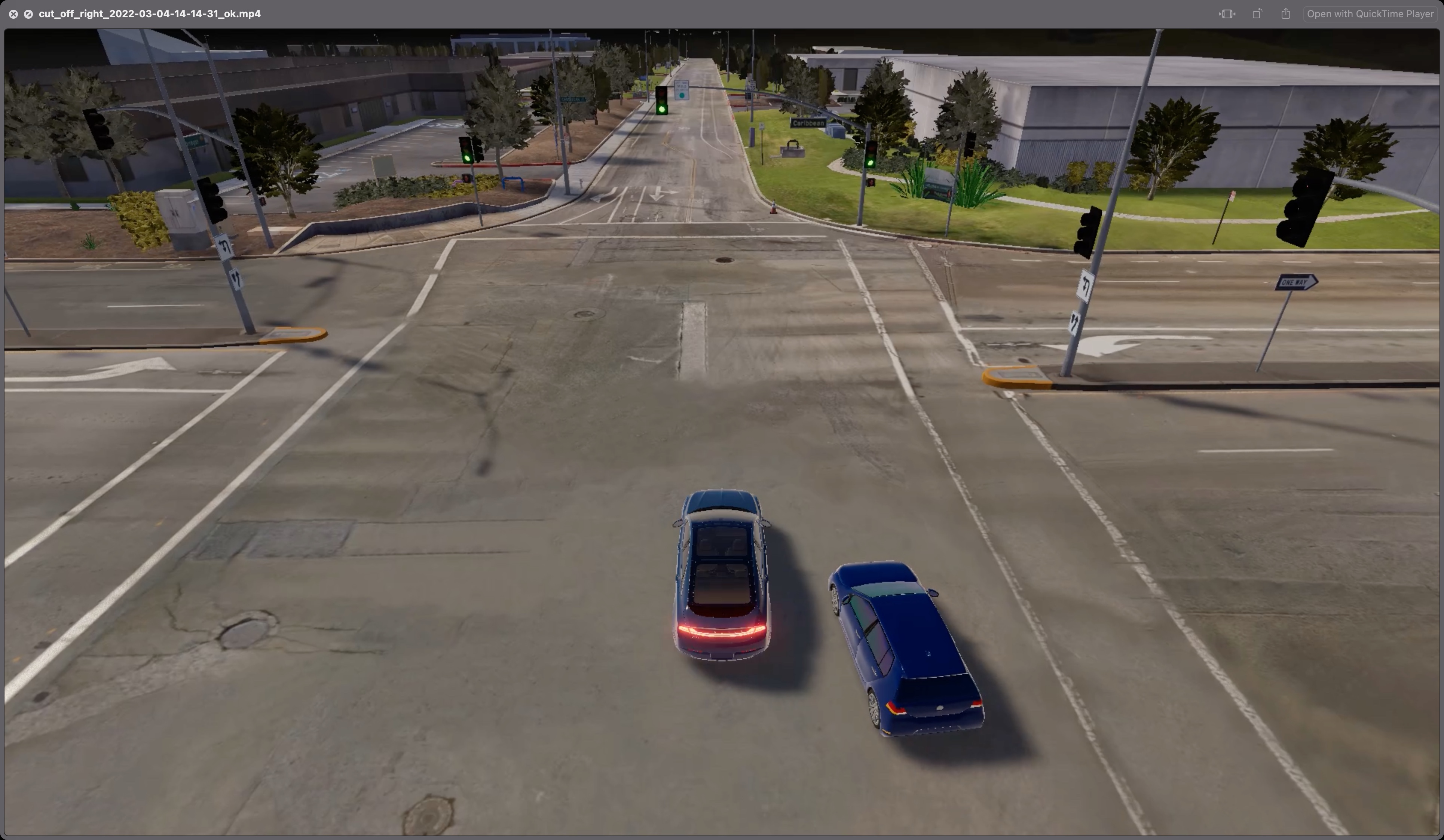}
  &
  \begin{tikzpicture}[
    path image/.style={path picture={
      \node at (6,4) {\includegraphics[scale=0.8, rotate=105]{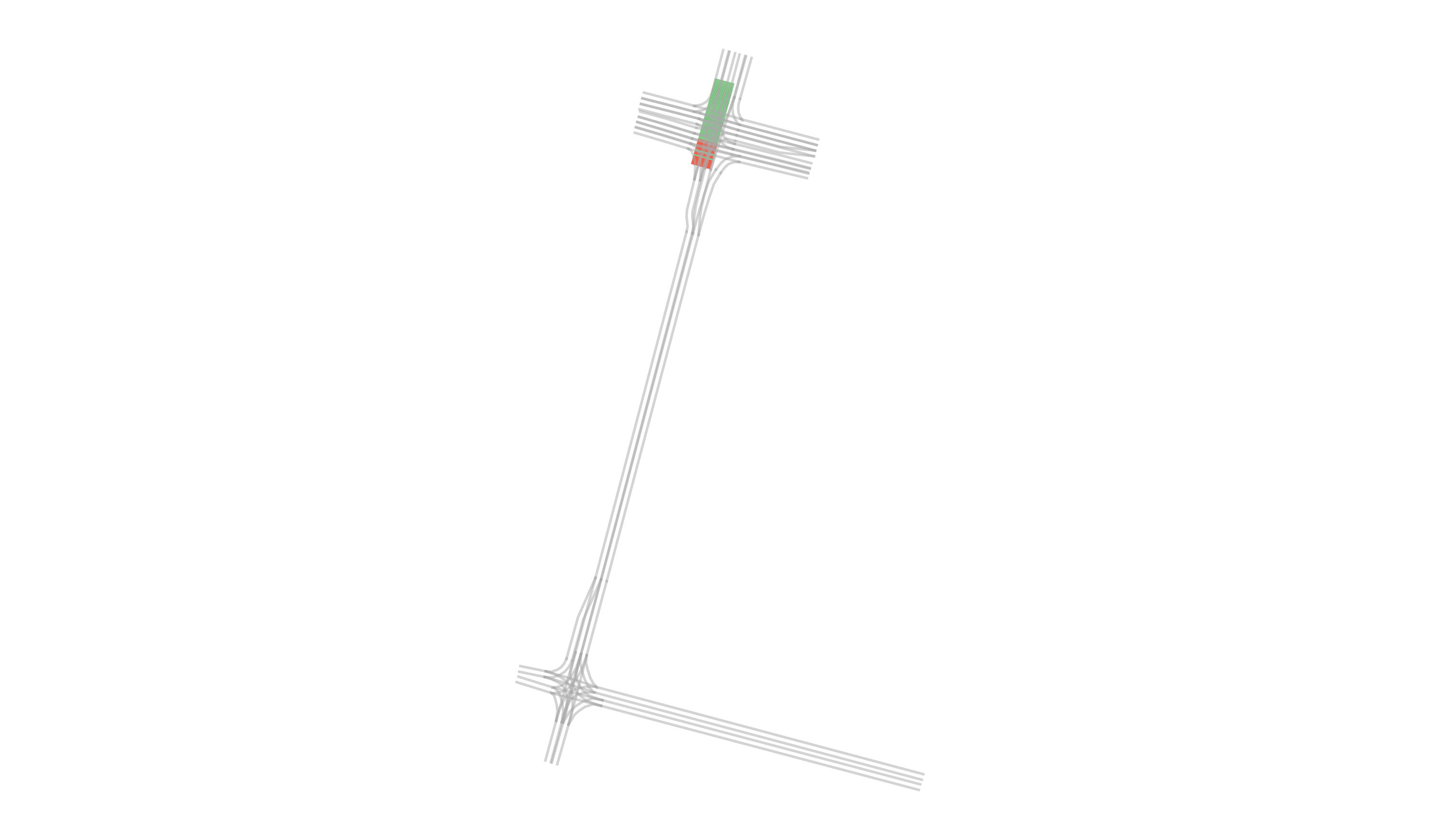}};}
    }]
    \path [path image] (0,0) rectangle (\linewidth, 3cm);
  \end{tikzpicture}
  \\
  \multicolumn{2}{p{13cm}}{
    \textbf{Cut Off Right}.
    While driving in the left lane on a two-lane road, a vehicle from the right lane cuts the ego vehicle off while turning into a parking lot.
  }\\
  \midrule 
  \includegraphics[width=6cm]{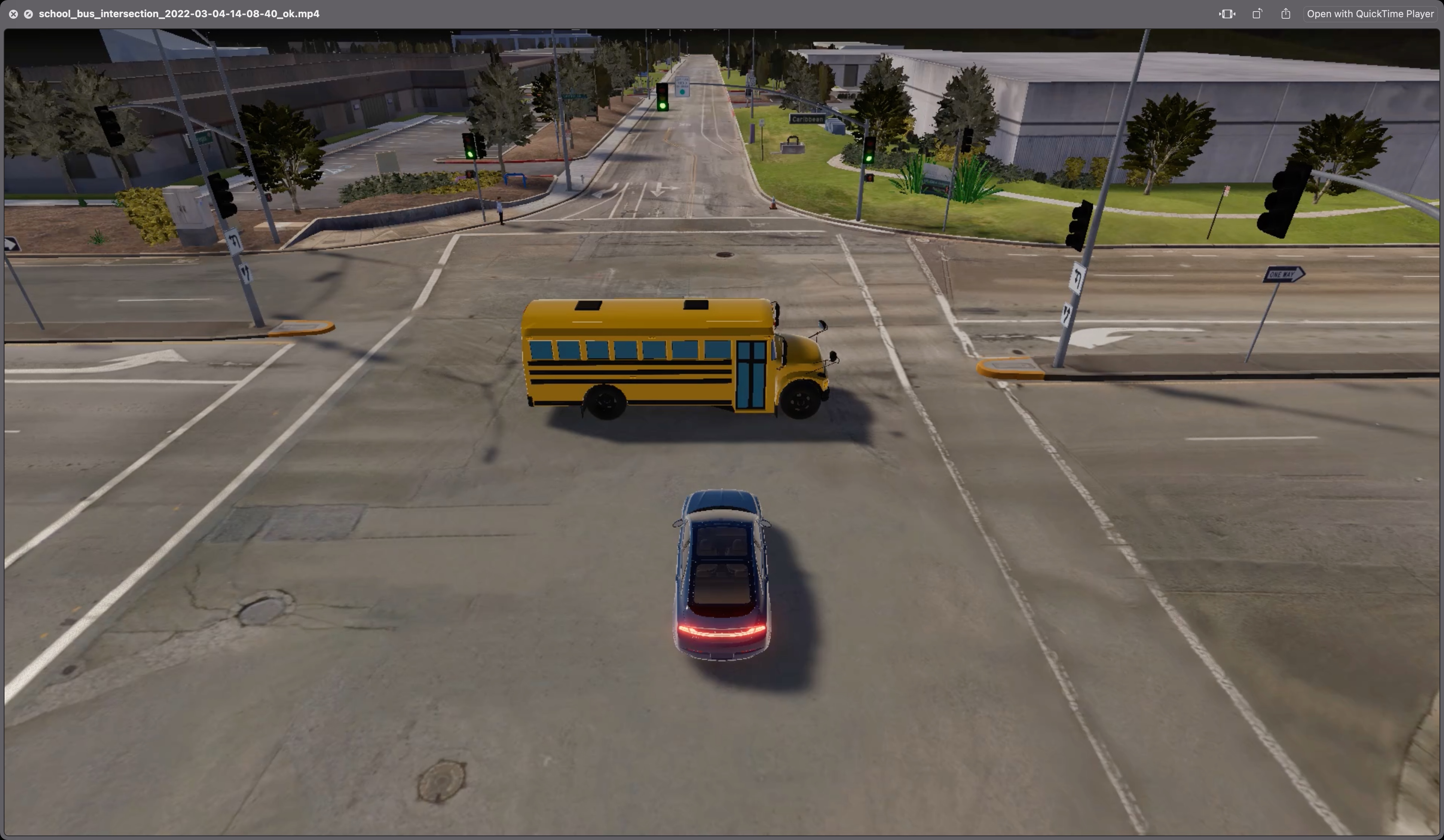}
  &
  \begin{tikzpicture}[
    path image/.style={path picture={
      \node at (6,4) {\includegraphics[scale=0.8, rotate=105]{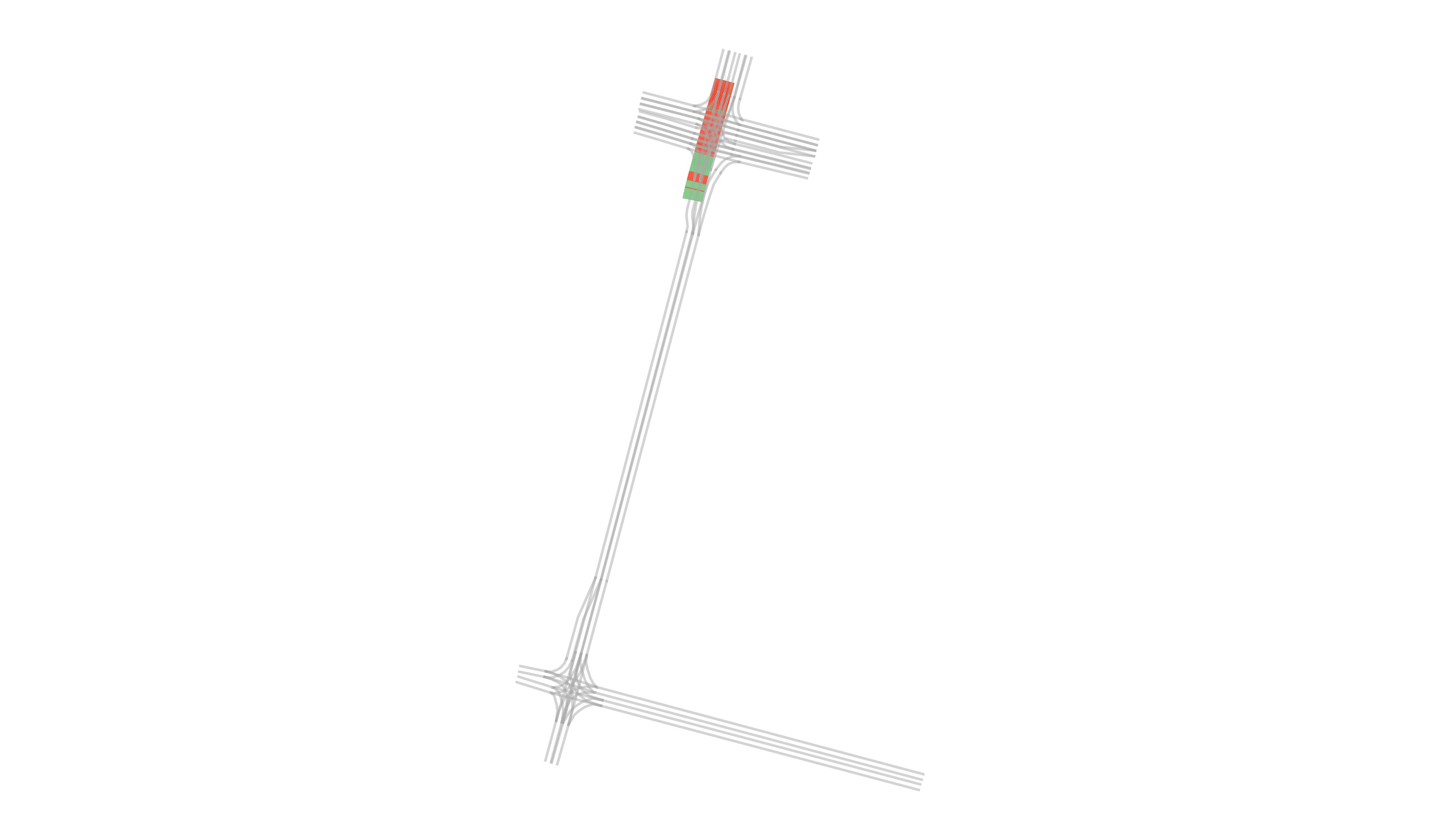}};}
    }]
    \path [path image] (0,0) rectangle (\linewidth, 3cm);
  \end{tikzpicture}
  \\
  \multicolumn{2}{p{13cm}}{
    \textbf{School Bus Intersection}.
    The ego vehicle drives through an intersection. A school bus crosses the intersection coming from the left-hand side.
    As the ego vehicle crosses the intersection, a pedestrian steps into the intersection from the left-hand side.
  }\\
  \midrule 
  \includegraphics[width=6cm]{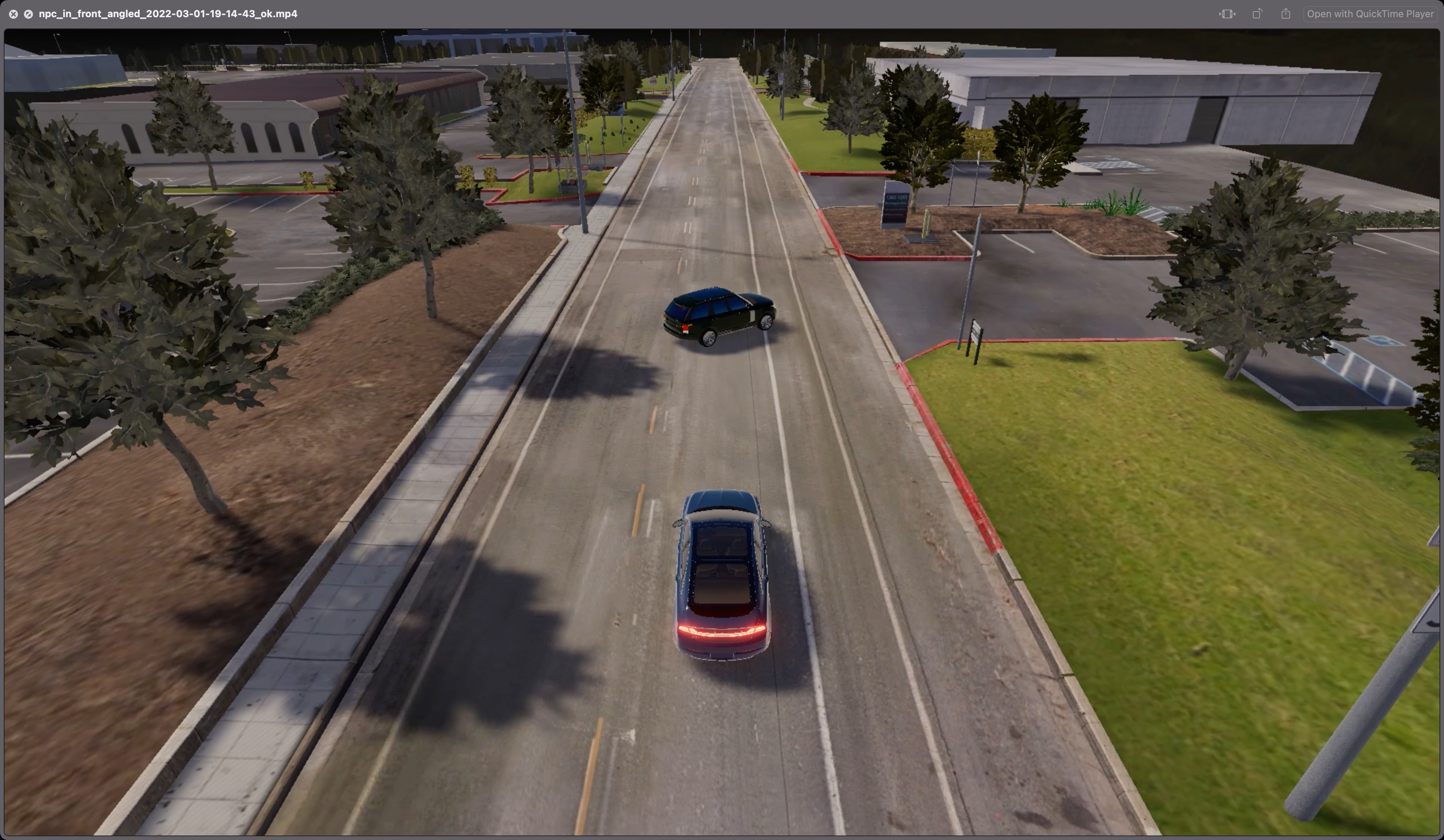}
  &
  \begin{tikzpicture}[
    path image/.style={path picture={
      \node at (6,3) {\includegraphics[scale=0.8, rotate=105]{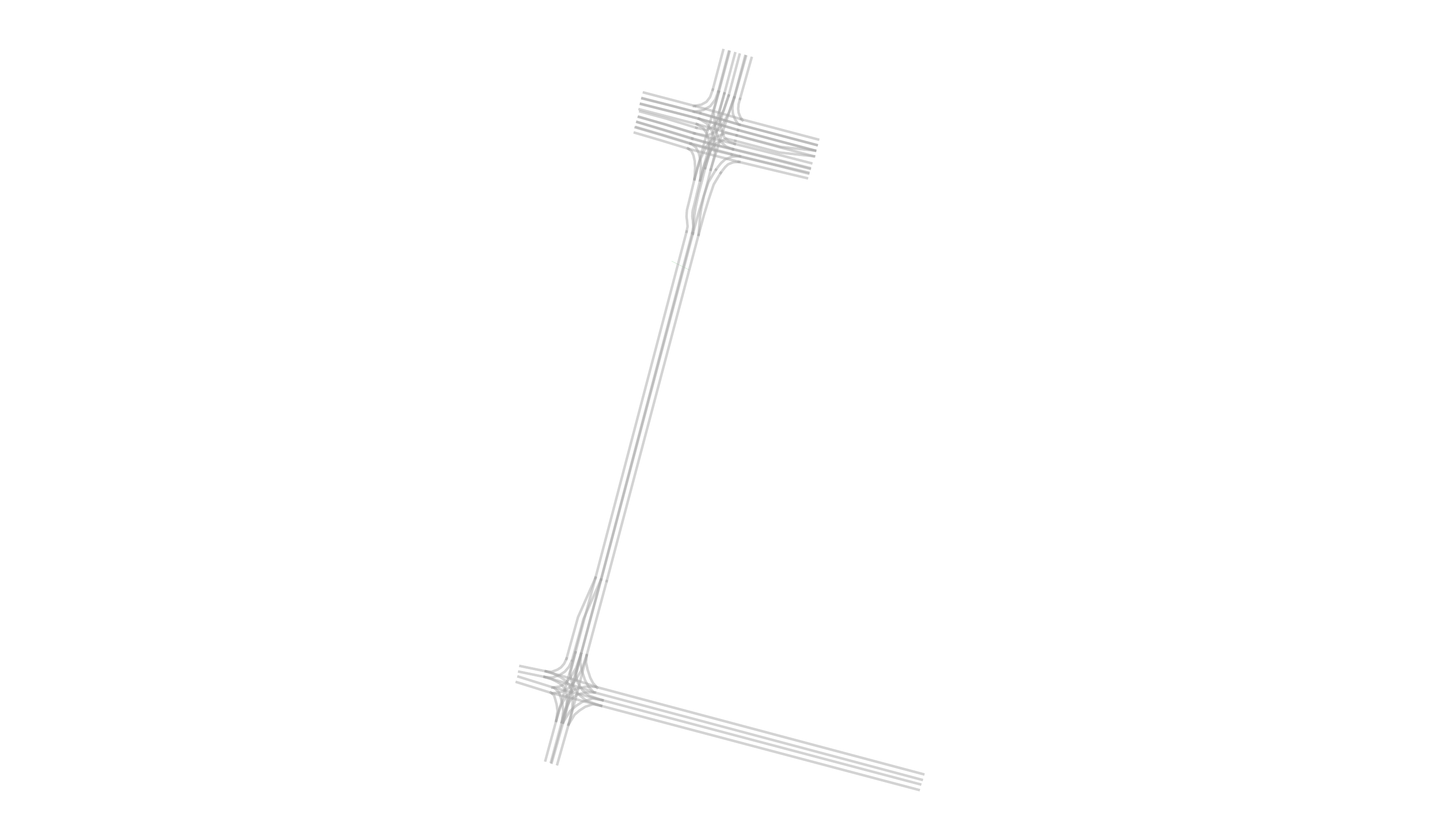}};}
    }]
    \path [path image] (0,0) rectangle (\linewidth, 3cm);
  \end{tikzpicture}
  \\
  \multicolumn{2}{p{13cm}}{
    \textbf{Car in Front}.
    A car is still in front of the ego vehicle preventing it to move forward.
  }\\
  \midrule 
  \includegraphics[width=6cm]{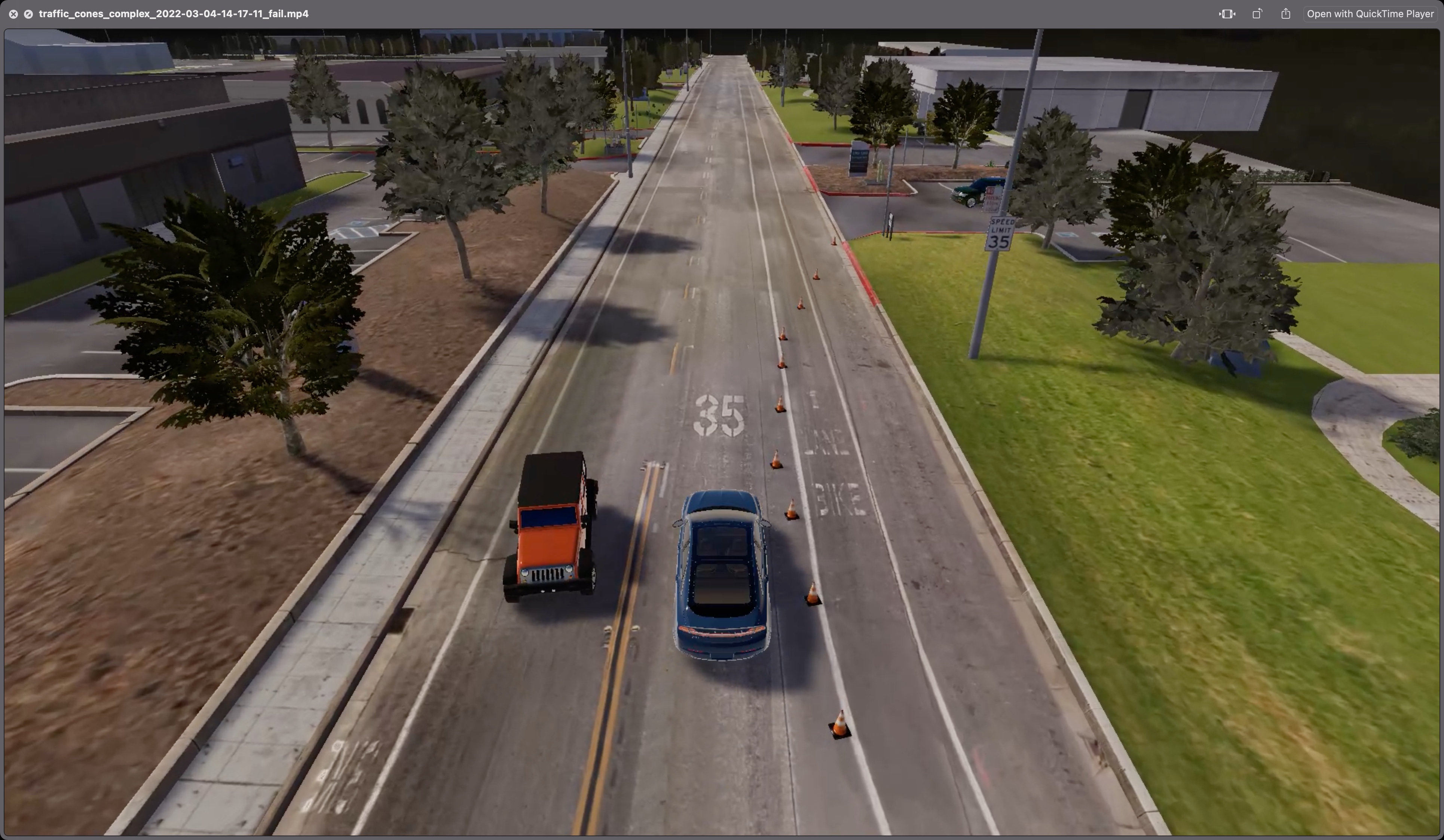}
  &
  \begin{tikzpicture}[
    path image/.style={path picture={
      \node at (4.5,3) {\includegraphics[scale=0.8, rotate=105]{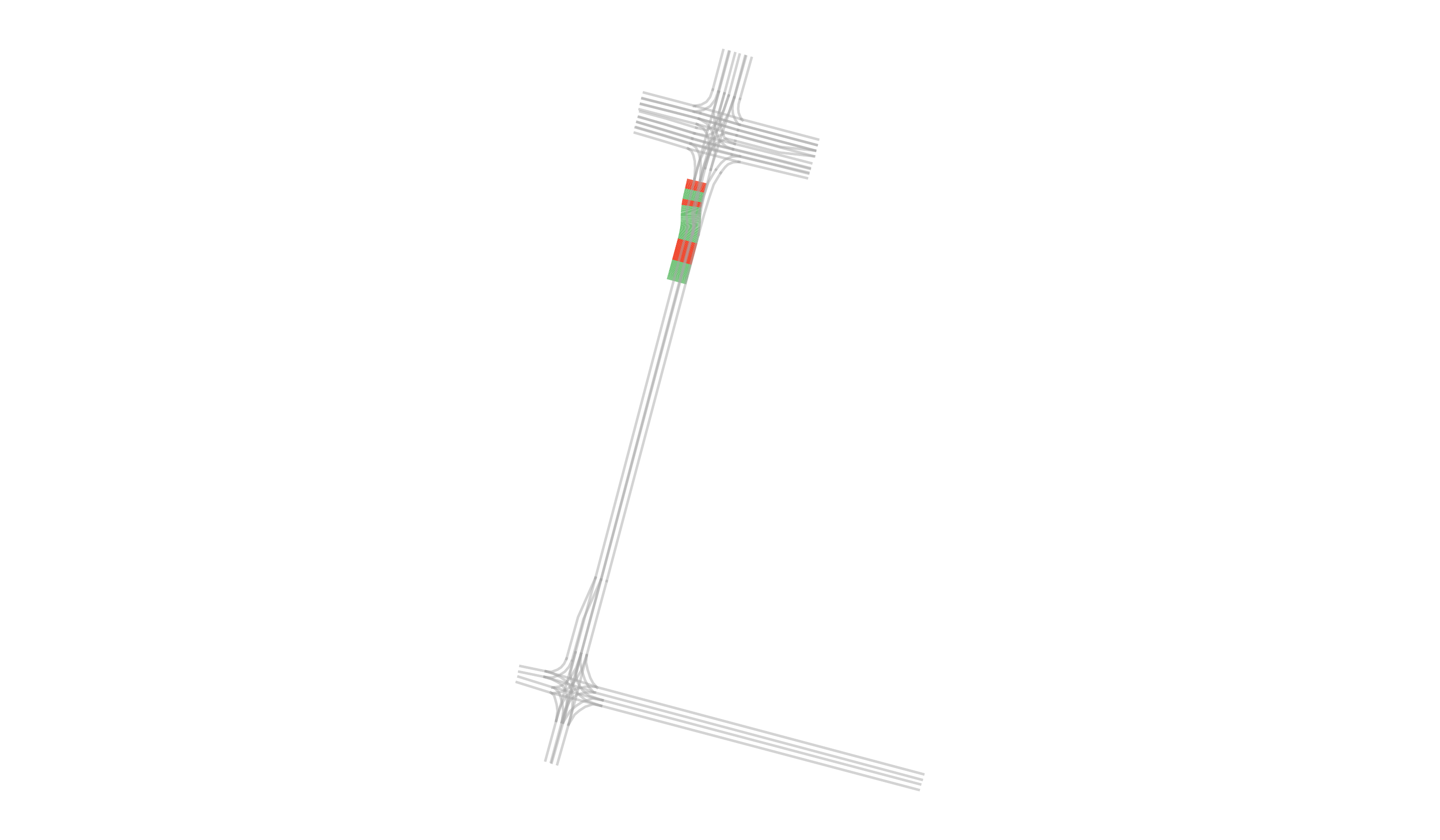}};}
    }]
    \path [path image] (0,0) rectangle (\linewidth, 3cm);
  \end{tikzpicture}
  \\
  \multicolumn{2}{p{13cm}}{
    \textbf{Cones in the Lane}.
    The ego vehicle is driving on a lane partially delimited by traffic cones, while another vehicle is driving in the opposite lane. 
    After passing traffic cones, another vehicle exits a parking lot and merges right in front of the ego vehicle.
  }\\
  \midrule 
  \includegraphics[width=6cm]{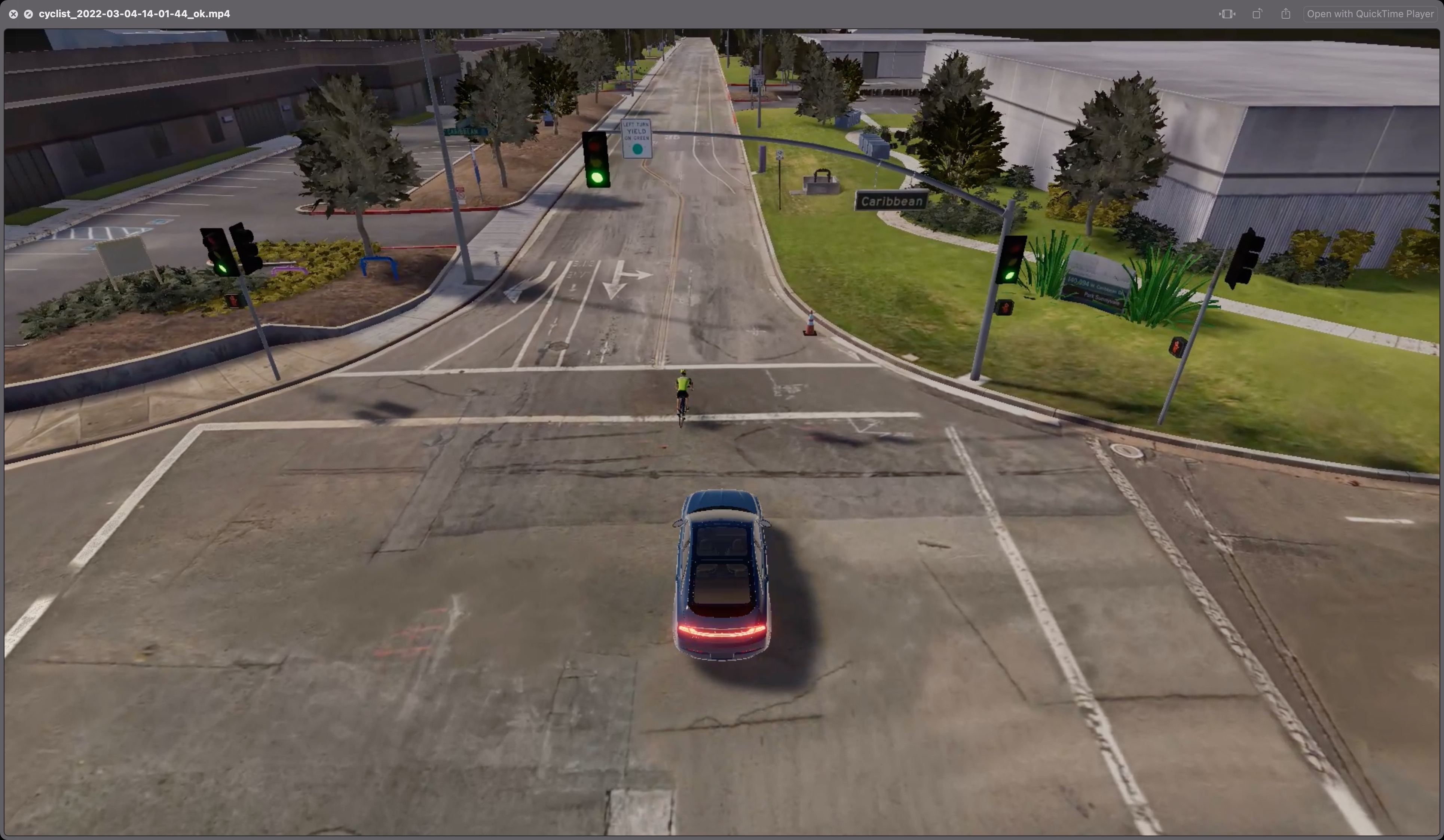}
  &
  \begin{tikzpicture}[
    path image/.style={path picture={
      \node at (9,3) {\includegraphics[scale=1.2, rotate=105]{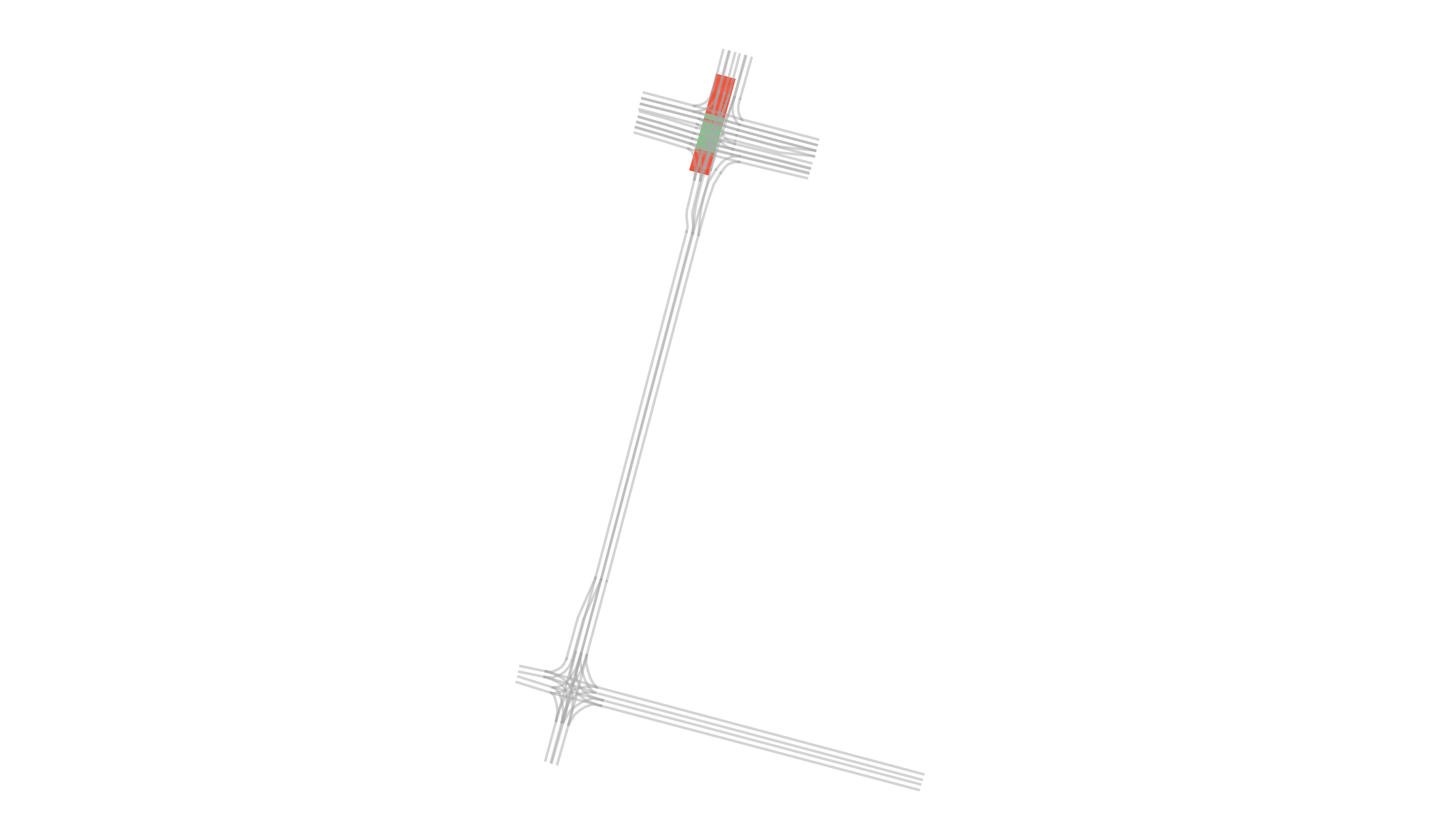}};}
    }]
    \path [path image] (0,0) rectangle (\linewidth, 3cm);
  \end{tikzpicture}
  \\
  \multicolumn{2}{p{13cm}}{
    \textbf{Cyclist}.
    The ego vehicle is stopped at an intersection and as it starts driving through the intersection, a cyclist enters the field of view from the left-hand side of the intersection and rides right in front of the ego vehicle.
  }\\
  \midrule 
  \includegraphics[width=6cm]{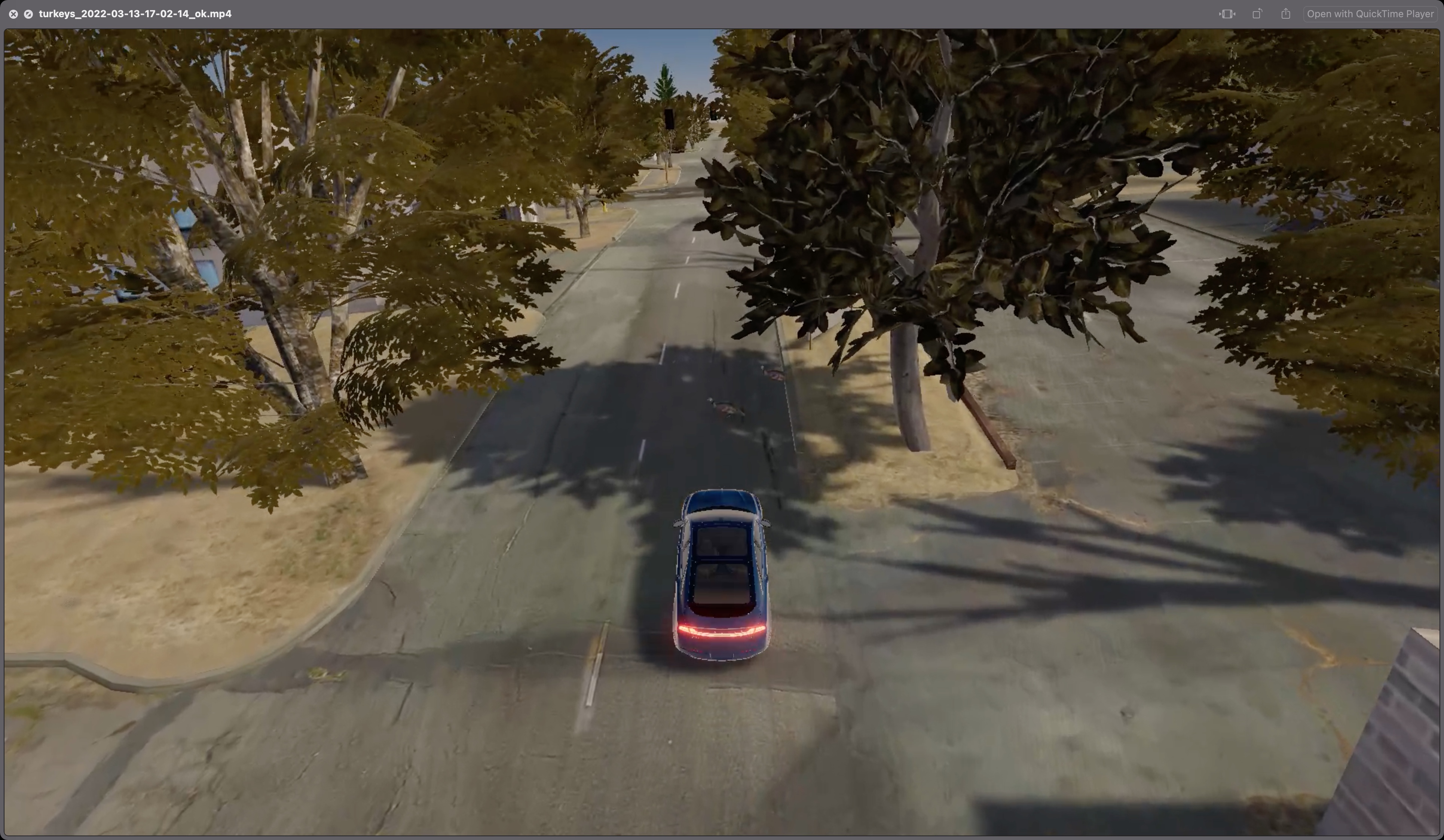}
  &
  \begin{tikzpicture}[
    path image/.style={path picture={
      \node at (-4,1) {\includegraphics[scale=0.8, rotate=28]{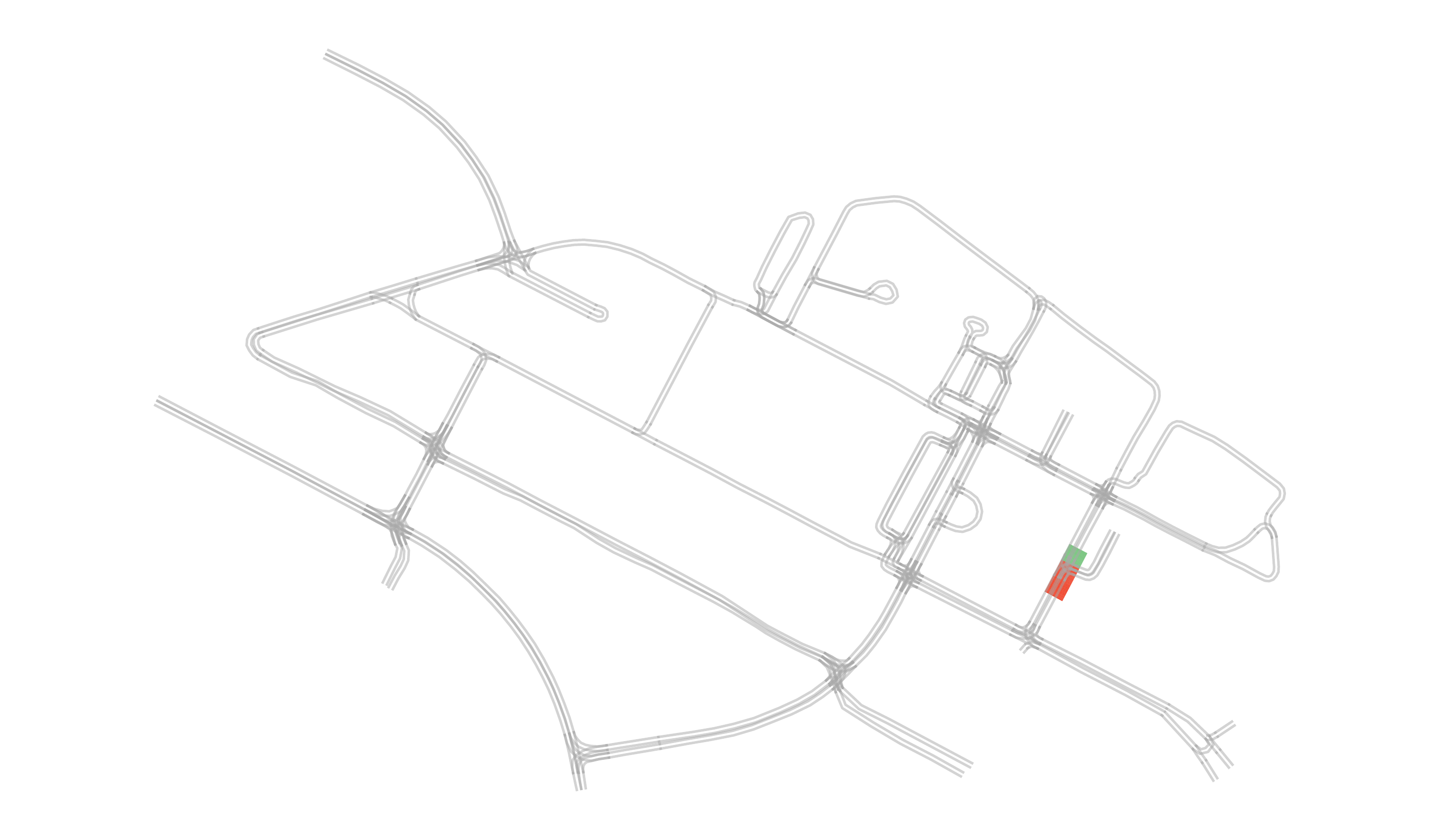}};}
    }]
    \path [path image] (0,0) rectangle (\linewidth, 3cm);
  \end{tikzpicture}
  \\
  \multicolumn{2}{p{13cm}}{
    \textbf{Turkeys}.
    While driving on a straight road, the ego vehicle must avoid a collision with two turkeys that suddenly walk in front of the ego vehicle.
  }
\end{longtable}

%% file: sections/subsections/results.tex

\subsection{Fault Detection and Identification Results}
\label{sec:results}

We used three metrics to evaluate the performance for both the fault detection and identification problems:
\begin{description}
  \item[Accuracy] is the percentage of correctly detected (resp. identified) failures over the total number of samples;
  \item[Precision] measures the percentage of correct identifications over the number of failures the fault identification system reported; a monitor achieves high precision if it has a low rate of false alarms;
  \item[Recall] measures the percentage of correct identifications over the number of failures the system experienced; 
  a monitor has high recall if it is able to catch a large fraction of failures occurring in the perception system.
\end{description}


\input{tables/fault_identification.tex}

\input{sections/figures/fault_identification.tex}

\subsubsection{Fault Identification Results}\label{sec:fault_identification_results}

\cref{tab:fault_identification} reports the accuracy of all compared techniques, averaged across all test scenarios in~\cref{tab:scenarios}. The first and fourth columns report the overall accuracy (``All'') when using regular and temporal 
\dgraphs, respectively. The remaining columns report a breakdown of the accuracy in terms of modules and outputs. 
The overall accuracy results suggest that factor-graph-based probabilistic fault identification 
outperforms all other algorithms and achieves \SI{96.72}{\percent} accuracy when using regular \dgraphs and 
\SI{96.88}{\percent} with temporal \dgraphs. 
GNNs architectures achieve the second-best performance (GraphSAGE in the regular case, GIN in the temporal case).
If we now look at the breakdown of the fault identification results between modules and outputs, we notice 
two trends. 
First, the factor graph still performs the best across the spectrum, but it is slightly slightly inferior than a baseline in the regular case. As we will see shortly, the baselines tend to make quite conservative decisions (\ie they tend to detect more failures than the ones actually present in the system), which increases accuracy (and recall) at the expense of precision.
Second, output fault identification has higher accuracy than module fault identification; this is expected, since most of our tests directly involve outputs, while we can only indirectly infer module failures via the a priori relations.
Note that the two statistics (output fault identification vs. module fault identification) are typically used for different purposes, as discussed in~\cref{rmk:modules_vs_outputs}.

\cref{fig:precision_recall_regular} shows precision-recall trade-offs when using regular \dgraphs.
 Best results are near the top-right corner of each figure, where both precision and recall are high.
The figure confirms that while the baselines have large recall (due to the fact that are conservative 
in detecting failure modes as active), their precision is relatively low (\ie they have a large number of false alarms).
On the other side of the spectrum, GNN architectures (with the exception of GCNII) achieve high prediction 
(\SI{87.25}{\percent} for GraphSAGE) but low recall (\SI{60.96}{\percent} for GraphSAGE).
The deterministic fault identification struggles to mark \fmodes as active, achieving low precision and recall in the output space; this is due to the fact that it disregards \pass results (which do not even appear in the optimization~\cref{eq:optimization_weakerOR}).
Factor graph inference again achieves a reasonable trade-off, with 
 \SI{85.22}{\percent} precision and \SI{67.12}{\percent} recall.
\cref{fig:precision_recall_temporal} shows precision-recall trade-offs when using temporal \dgraphs.
Compared to the regular \dgraph we see a steep increase in precision in the output space.
The best-performing model goes from around \SI{90}{\percent} precision of the regular graph to \SI{97}{\percent} of the temporal \dgraph.

\myParagraph{PAC-Diagnosability}
\cref{fig:pac_bounds_regular} and \cref{fig:pac_bounds_temporal} show the PAC-Diagnosability bound defined in \cref{eq:pac-diagnosability-bound} for each of the compared techniques.
The bound represents the number of fault identification mistakes each algorithm is expected to make with a given confidence ($\delta$ in \cref{eq:pac-diagnosability-bound}).
The plots show that with high probability, most of the algorithms are expected to make less than $1$ mistake in the fault identification (\ie false alarms or false negatives).
The factor graph has the lowest bound of all methods in both the regular and temporal \dgraphs;
the only exception is \cref{fig:pac_bounds_regular}(right), where the baseline with reliability score has the lowest bound for module fault identification. 

\input{sections/figures/pac_bounds.tex}

\myParagraph{\kdiagnosability}
Let us now discuss the deterministic diagnosability of the perception system considered in our experiments 
(\cref{fig:apollo_diagnostic_graph}).
If the tests behave as a Deterministic OR, the \dgraph used in our experiments is $5$-diagnosable.
This means that if there are up to $5$ active \fmodes the deterministic fault identification will be able to correctly identify them.
If we instead assume the tests behave as a Weak-OR, which might fail when all the \fmodes in its scope are active, 
the \dgraph is $3$-diagnosable.
It's worth noticing that this does not mean that if there are more than $3$ (or $5$) active \fmodes the fault identification will surely fail, but rather that we do not have the guarantee that it will not make any mistake. 
When using Deterministic \WeakerOR tests, the diagnosability drops to zero, meaning that the fault identification guarantees vanish. 

\myParagraph{Extra diagnosability results}
To show the effectiveness of the deterministic and probabilistic diagnosability we generated a random $4$-diagnosable \dgraph with \num{10} independent \fmodes and Weak-OR tests and collected the fault identification results (using the deterministic model)  for every syndrome and every possible fault assignment.
The results are shown in \cref{fig:dummy_example}.
The figure reports the average number of incorrect fault identification results (\ie the Hamming distance between the 
estimated and actual vector of active faults) for increasing number of active faults.
The vertical dashed line represents the deterministic diagnosability value: by \cref{def:diagnosability}, the fault identification is guaranteed to correctly identify the active \fmodes provided that there are less than \num{4} active \fmodes.
In fact, from the plot we see that the fault identification algorithm does not make any mistake in the fault identification when there are less than \num{4} faults.
The horizontal dashed line instead represents the probabilistic diagnosability value, in particular it is the ceiling of the bound in \cref{eq:pac-diagnosability-bound}, computed with very high confidence ($1\!-\!1\!\times\!10^{-12}$).
The bound guarantees that with high probability the average number of mistakes (the average Hamming distance)  the  fault identification algorithm is going to make is less that \num{2}; this is again consistent with the numerical results.

\input{sections/figures/dummy.tex}

\myParagraph{Timing}
The runtime of each method is shown in~\cref{tab:timing}.
All algorithms perform inference in less than \SI{4}{\milli\second}, except for GCNII which averages at around \SI{20}{\milli\second}.
This is likely due to the fact that GCNII uses a deep architecture, which incurs an increased computational cost.
The best performing algorithm, \ie the factor graph, can be executed in real-time as its runtime averages around \SI{0.8}{\milli\second} for regular graphs and \SI{3.8}{\milli\second} for temporal graphs.

\input{tables/timing.tex}

\subsubsection{Fault Detection Results}\label{sec:fault_detection_results}
Recall that fault detection is the problem of deciding whether the system is working in normal conditions or whether at least a fault has occurred. 
\cref{tab:fault_detection} and \cref{fig:fault_detection} show accuracy, precision, and recall.
\cref{fig:fault_detection}  shows that most of the algorithms for inference presented in this paper (as well as the baselines) attain similar performance with precision above \SI{90}{\percent} and recall above \SI{80}{\percent}; this confirms that fault detection is a somewhat easier problem compared to fault identification. 
\cref{tab:fault_detection} shows that the deterministic approach and the baselines do particularly well for fault detection: they both detect failure as soon as a single test fails, which makes their accuracy high.
On the other hand, the factor graph approach may prefer explaining a failed test as a false alarm.
Therefore, while factor graphs would be the go-to approach for fault identification, a simpler baseline approach 
suffices for fault detection.

\input{tables/fault_detection.tex}
\input{sections/figures/fault_detection.tex}

%% file: tables/fault_identification.tex

\begin{table}
  \smaller
  \begin{tabular}{ |l||c|c|c|c|c|c| }
    \hline
    \multirow{2}{*}{\textbf{Algorithm}} & \multicolumn{3}{c|}{\textbf{Regular}} & \multicolumn{3}{c|}{\textbf{Temporal}}\\
    & All & Outputs & Modules & All & Outputs & Modules \\    \hline
    Factor Graph              & \fst{93.30} & \fst{96.72} & 83.03       & \fst{93.60} & \fst{96.88} & \fst{83.74} \\
    Deterministic             & 91.06       & 93.69       & \snd{83.18} & 89.26       & 92.33       & 80.06 \\
    Baseline (w/rel. scores)  & 92.39       & 94.65       & \fst{85.61} & 90.18       & 92.69       & 82.67 \\
    Baseline                  & 84.85       & 89.09       & 72.12       & 83.90       & 87.73       & 72.39 \\
    GCN                       & 92.27       & 96.01       & 81.06       & 91.79       & 96.06       & 78.99 \\
    GCNII                     & 87.61       & 93.94       & 68.64       & 92.60       & 96.01       & 82.36 \\
    GIN                       & 91.89       & 96.06       & 79.39       & \snd{93.21} & \snd{96.47} & \snd{83.44} \\
    GraphSage                 & \snd{92.84} & \snd{96.46} & 81.97       & 92.71       & 96.42       & 81.60 \\
   \hline
  \end{tabular}
  \caption{Fault identification accuracy. Best accuracy if highlighted in green, second-best is highlighted in yellow.}
  \label{tab:fault_identification}
\end{table}

%% file: sections/figures/fault_identification.tex

\begin{figure}[!ht]
  \centering
  \hspace*{-1.2cm}
  \includegraphics[width=1.2\textwidth]{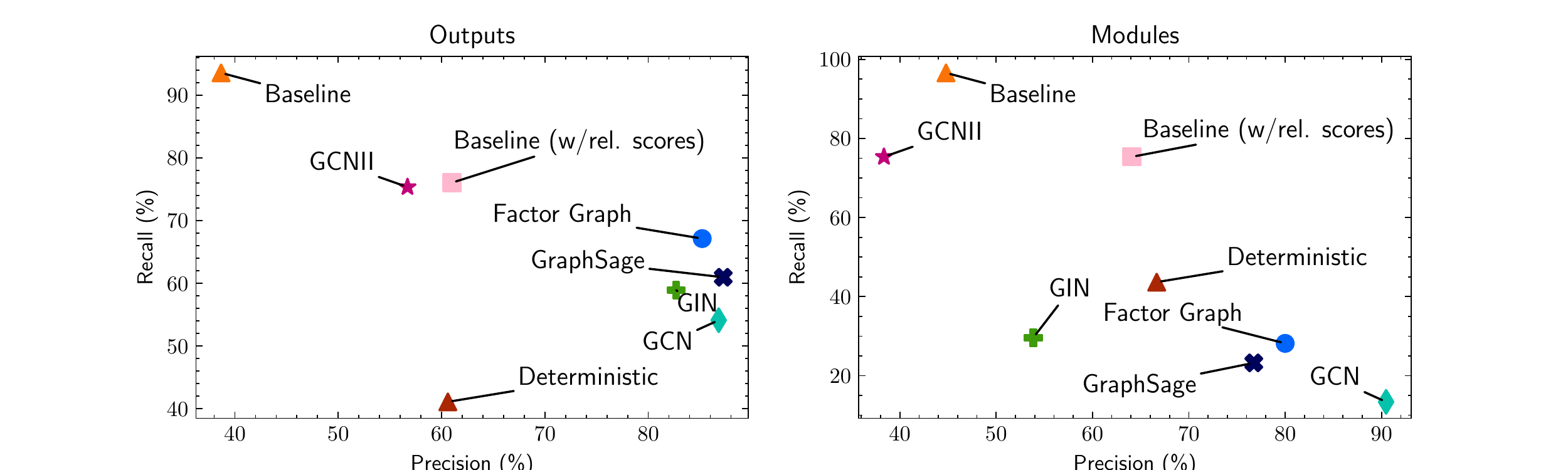}
  \caption{Precision/Recall for regular \dgraphs. (Left) Modules, (Right) Outputs. 
  }
  \label{fig:precision_recall_regular}
\end{figure}

\begin{figure}[!ht]
  \centering
  \hspace*{-1.2cm}
  \includegraphics[width=1.2\textwidth]{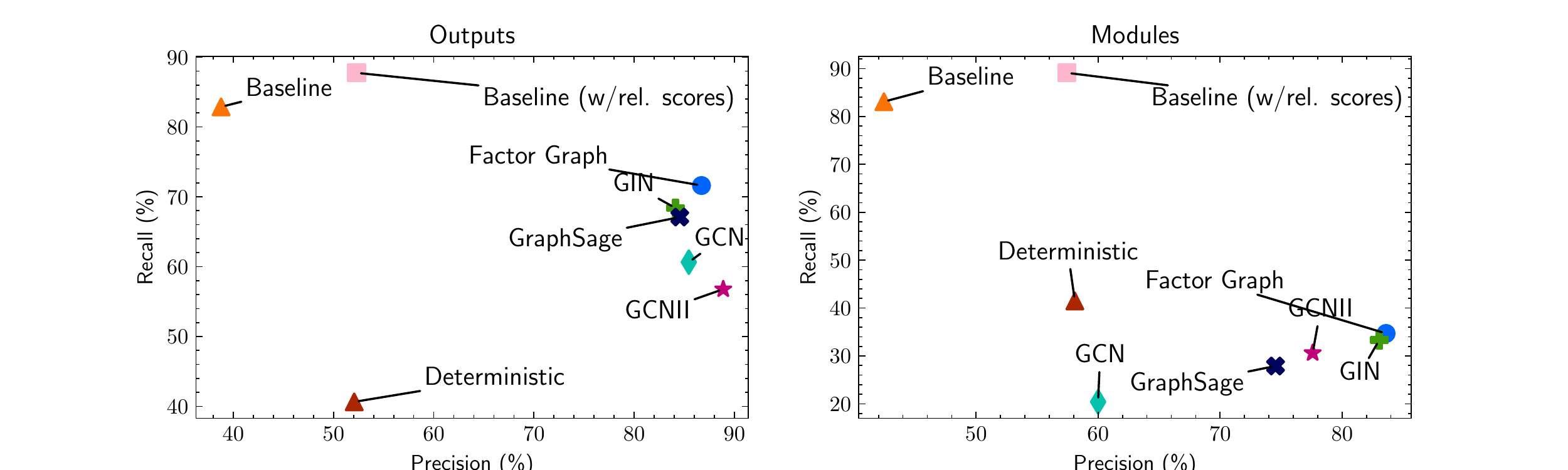}
  \caption{Precision/Recall for temporal \dgraphs. (Left) Modules, (Right) Outputs.}
  \label{fig:precision_recall_temporal}
\end{figure}

%% file: sections/figures/pac_bounds.tex

\begin{figure}[!ht]
  \centering
  \hspace*{-1.2cm}
  \includegraphics[width=1.2\textwidth]{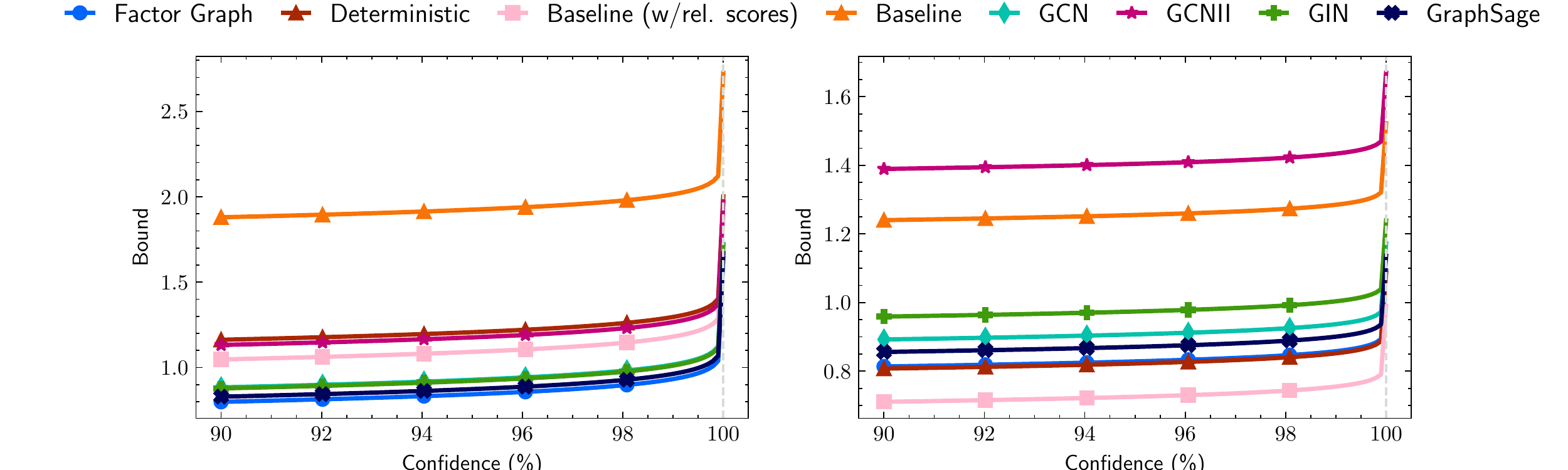}
  \caption{PAC-diagnosability bounds for regular \dgraphs. (Left) Modules, (Right) Outputs. Lower is better.\label{fig:pac_bounds_regular}}
\end{figure}

\begin{figure}[!ht]
  \centering
  \hspace*{-1.2cm}
  \includegraphics[width=1.2\textwidth]{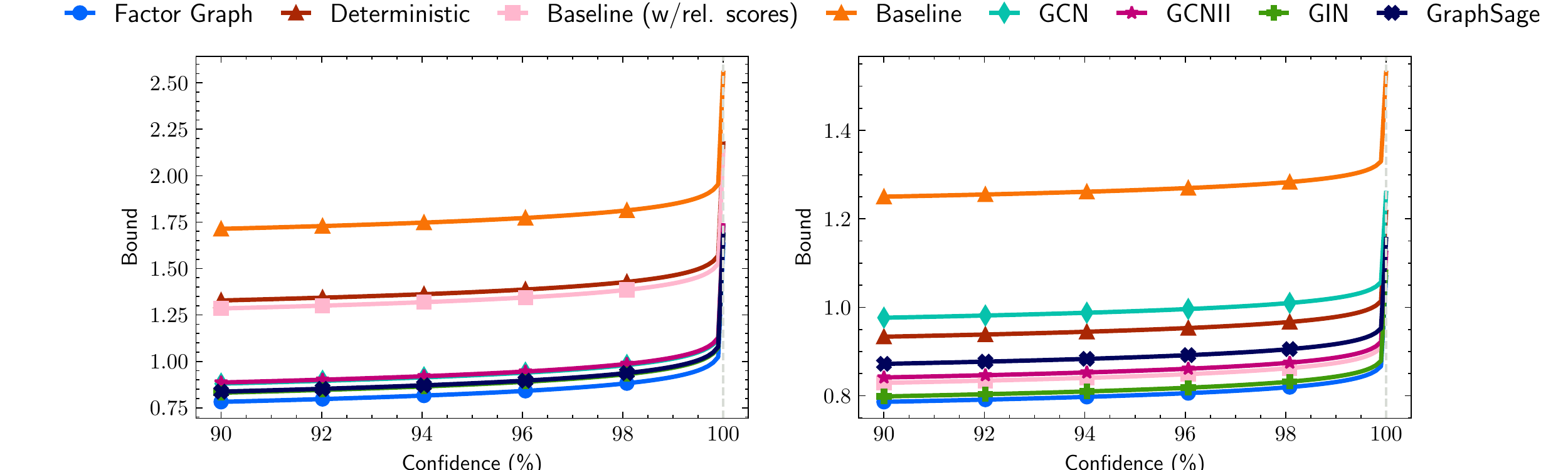}
  \caption{PAC-diagnosability bounds for temporal \dgraphs. (Left) Modules, (Right) Outputs. Lower is better.\label{fig:pac_bounds_temporal}}
\end{figure}

%% file: sections/figures/dummy.tex

\begin{figure}[!ht]
  \centering
  \includegraphics[width=1\textwidth]{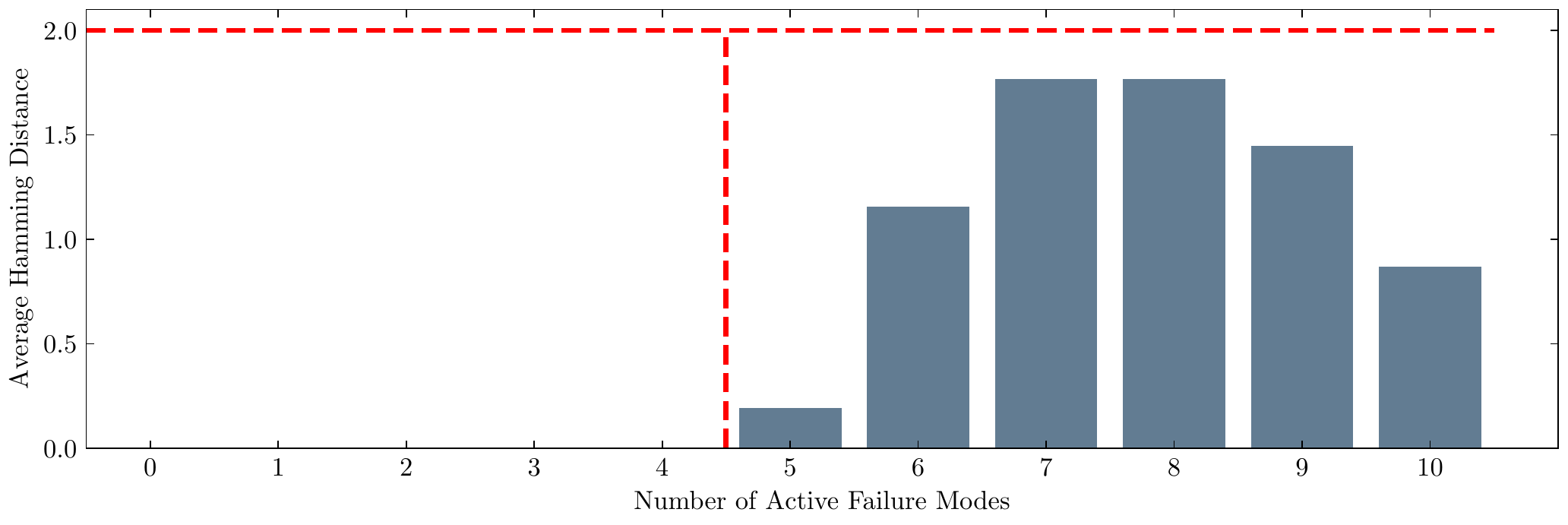}
  \caption{
    Average Hamming distance between the estimated and actual vector $\faults$ of fault states in a randomly generated $4$-diagnosable \dgraph with  $10$ independent \fmodes and Weak-OR tests.
    The vertical dashed line represents the deterministic diagnosability bound: if the system is experiencing less than $4$ active failure modes, the fault identification is guaranteed to be correct ($0$ Hamming distance).
    The horizontal dashed line represents the ceiling of the PAC-diagnosability bound in \cref{eq:pac-diagnosability-bound}: with very high probability the average number of mistakes (average Hamming distance) is less than the PAC-diagnosability bound.
  }
  \label{fig:dummy_example}
\end{figure}

%% file: tables/timing.tex

\begin{table}
  \hspace*{-0.0cm}
  \smaller{
  \begin{tabular}{ llcccccccc }
    & & \rotatebox{90}{\thead{Factor\\Graph}} & \rotatebox{90}{\thead{Deterministic}} & 
    \rotatebox{90}{\thead{Baseline\\(w/rel. scores)}} & \rotatebox{90}{\thead{Baseline}} & 
    \rotatebox{90}{\thead{GCN}} & \rotatebox{90}{\thead{GCNII}} & \rotatebox{90}{\thead{GIN}} & 
    \rotatebox{90}{\thead{GraphSage}} \\
    \midrule
    \parbox[t]{2mm}{\multirow{2}{*}{\rotatebox[origin=c]{90}{\smallskip Regular \smallskip}}}
      & Avg. & 0.79  & 3.25  & 0.10  & 0.10  & 0.63  & 19.88  & 0.48  & 0.59  \\
      & Std. & (0.17) & (0.14) & (0.06) & (0.06) & (0.01) & (0.10) & (0.01) & (0.02) \\[10pt]
    \parbox[t]{2mm}{\multirow{2}{*}{\rotatebox[origin=c]{90}{\smallskip Temporal \smallskip}}} 
      & Avg. & 2.53  & 3.68  & 0.27  & 0.26  & 0.68  & 24.56  & 0.50  & 0.85  \\
      & Std. & (0.04) & (0.46) & (0.17) & (0.16) & (0.01) & (0.33) & (0.01) & (0.01) \\
  \end{tabular}
  }
  \vspace{3mm}
  \caption{Average  runtime (``Avg.'') and standard deviation (``Std.'') for fault identification, in milliseconds.}
  \label{tab:timing}
\end{table}

%% file: tables/fault_detection.tex

\begin{table}
  \smaller
  \begin{tabular}{ |l||c|c|c|c|c|c| }
    \hline
    \multirow{2}{*}{\textbf{Algorithm}} & \multicolumn{3}{c|}{\textbf{Regular}} & \multicolumn{3}{c|}{\textbf{Temporal}}\\
    & All & Outputs & Modules & All & Outputs & Modules \\    \hline
    Factor Graph              & \snd{76.67} & \snd{88.48} & 64.85       & 81.60       & 91.41       & 71.78 \\
    Deterministic             & \fst{89.09} & \fst{89.09} & \fst{89.09} & \fst{93.25} & \fst{93.25} & \fst{93.25} \\
    Baseline (w/rel. scores)  & \fst{89.09} & \fst{89.09} & \fst{89.09} & \snd{85.28} & 85.28       & \snd{85.28} \\
    Baseline                  & \fst{89.09} & \fst{89.09} & \fst{89.09} & \snd{85.28} & 85.28       & \snd{85.28} \\
    GCN                       & 71.82       & 86.06       & 57.58       & 80.06       & 90.18       & 69.94 \\
    GCNII                     & 68.48       & 87.88       & 49.09       & 78.83       & 85.89       & 71.78 \\
    GIN                       & 83.94       & 86.06       & \snd{81.82} & 83.13       & \snd{92.64} & 73.62 \\
    GraphSage                 & \snd{76.67} & \fst{89.09} & 64.24       & 79.14       & 89.57       & 68.71 \\
   \hline
  \end{tabular}
  \caption{Fault detection accuracy. Best accuracy if highlighted in green, second-best is highlighted in yellow.\label{tab:fault_detection}}
\end{table}

%% file: sections/figures/fault_detection.tex

\begin{figure}[!ht]
  \centering
  \hspace*{-1.2cm}
  \includegraphics[width=1.2\textwidth]{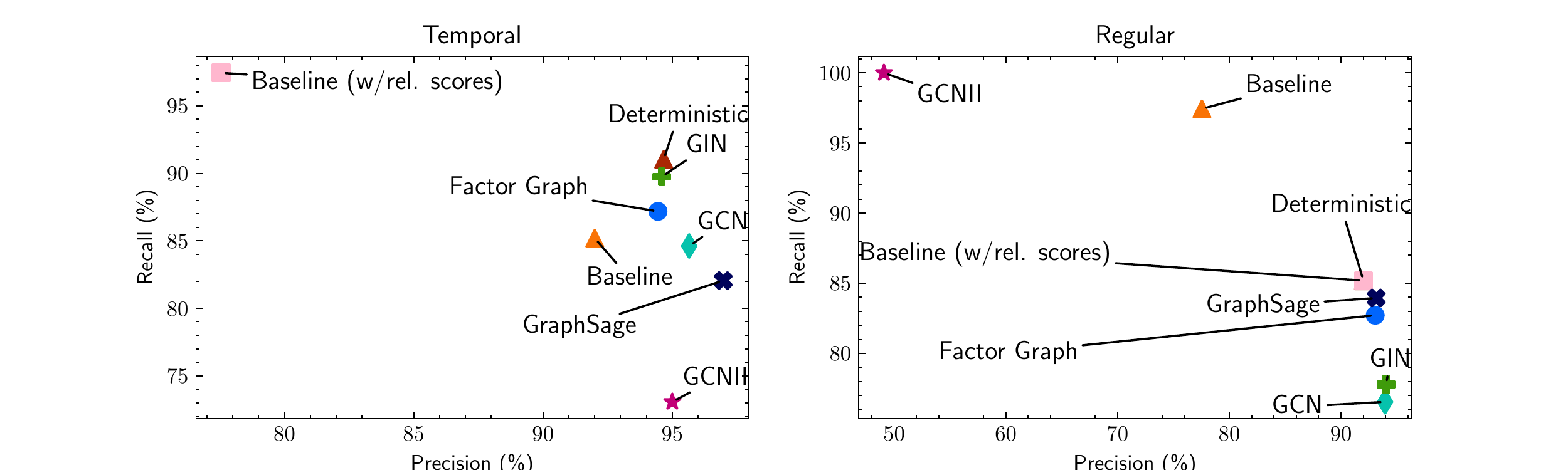}
  \caption{Fault detection in \dgraphs. (Left) Regular, (Right) Temporal. \label{fig:fault_detection}}
\end{figure}

%% file: sections/subsections/example_scenario.tex

\subsection{Example Scenario: Using Monitoring to Prevent Accidents}

\label{sec:example_scenario}

\begin{figure}[htbp]
  \centering
  \includegraphics[width=\textwidth]{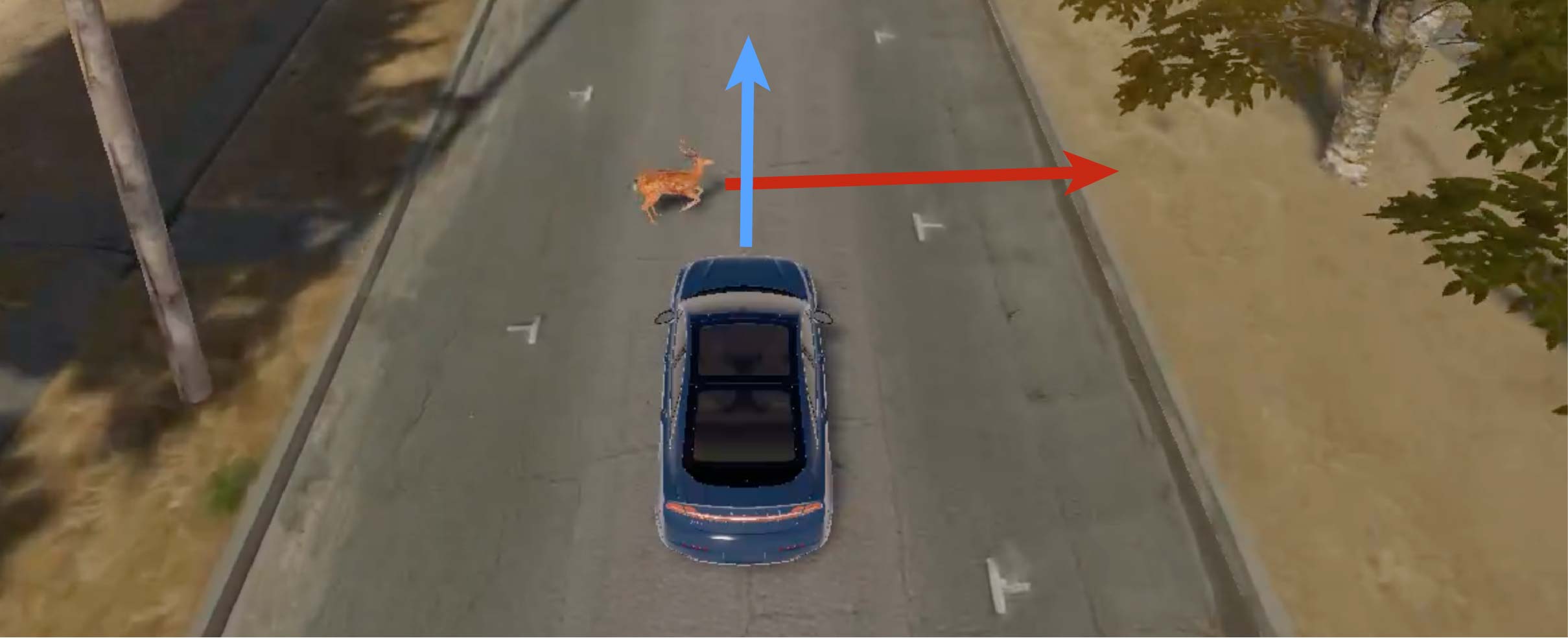}
  \caption{Example scenario involving a deer crossing the road in front of the ego vehicle.}
  \label{fig:deer_scenario}
\end{figure}

We conclude the experimental section by showing how fault detection and identification can be effectively used to prevent dangerous situations.
To this aim, we developed an additional scenario (not included in \cref{tab:scenarios}) where a deer crosses the road while the ego vehicle cruises on a straight road (\cref{fig:deer_scenario}).

\begin{figure}[ht]
  \centering
  \begin{tikzpicture}[
    path image/.style={path picture={
      \node at (2.7,3) {\includegraphics[scale=1.2, rotate=28]{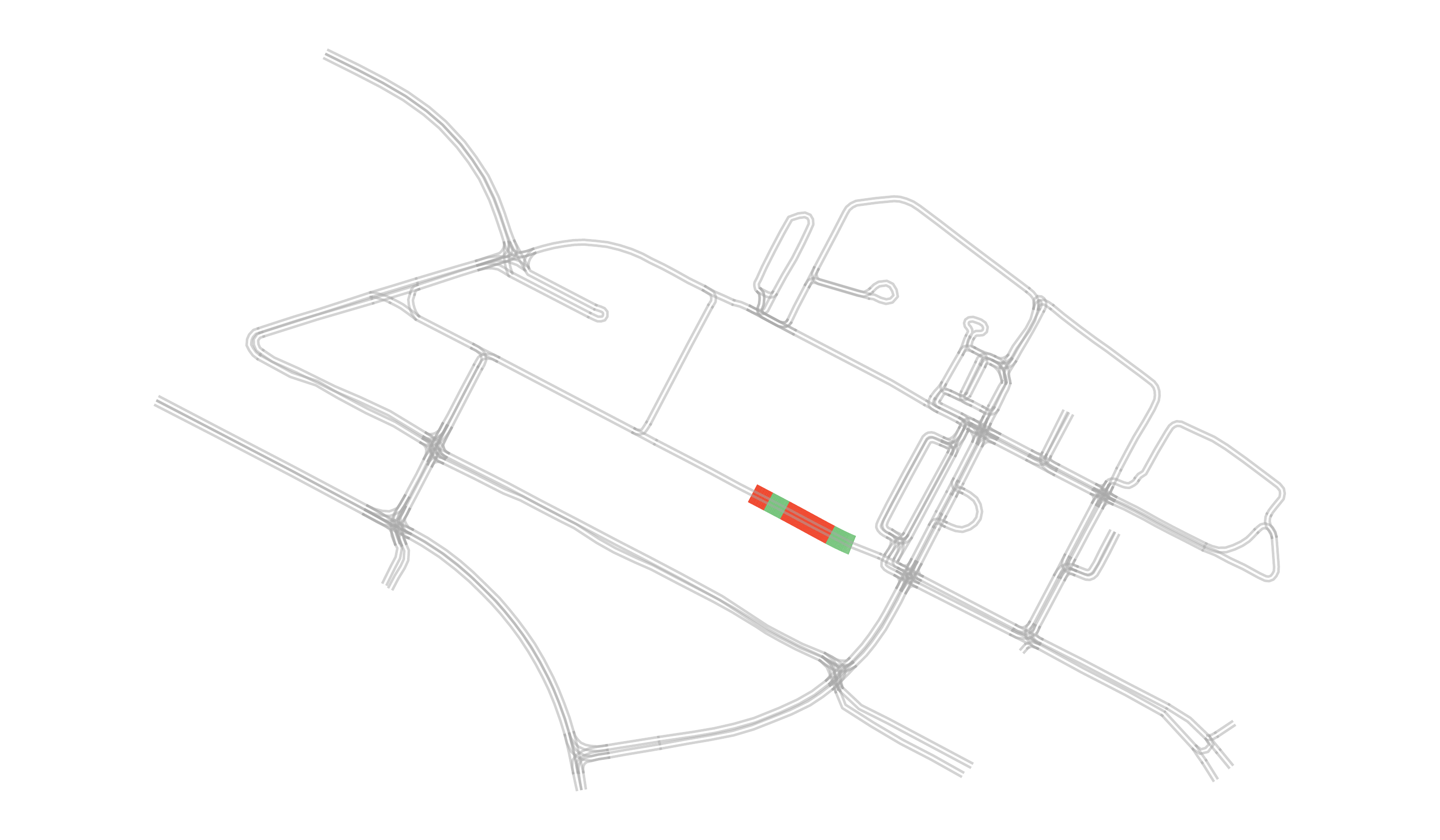}};}
    }]
    \definecolor{tempcolor}{RGB}{72,142,255}
    \path [path image] (0,0) rectangle (\linewidth, 44mm);
    \node at (4,1.8) {\includegraphics[scale=0.5, rotate=28]{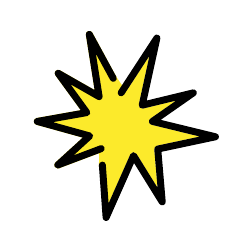}};
    \draw [line width=1mm,draw=tempcolor,stealth-](4.5,1) -- (7,1);
\end{tikzpicture}
\caption{
  Fault identification results for the example scenario in~\cref{fig:deer_scenario}.
  The car travels from right to left. 
  Initially, the monitor detects no failure (rightmost, green section). 
  As the ego vehicle gets closer to the obstacle, the LiDAR-based and camera-based obstacle detectors fail to detect the deer while the radar-based obstacle detector correctly locates the obstacle; as a result the fault 
  identification/detection triggers an alarm (red sections).
}
\label{fig:deer_traj}
\end{figure}

\begin{figure}[htbp]
  \centering
  \includegraphics[width=\textwidth]{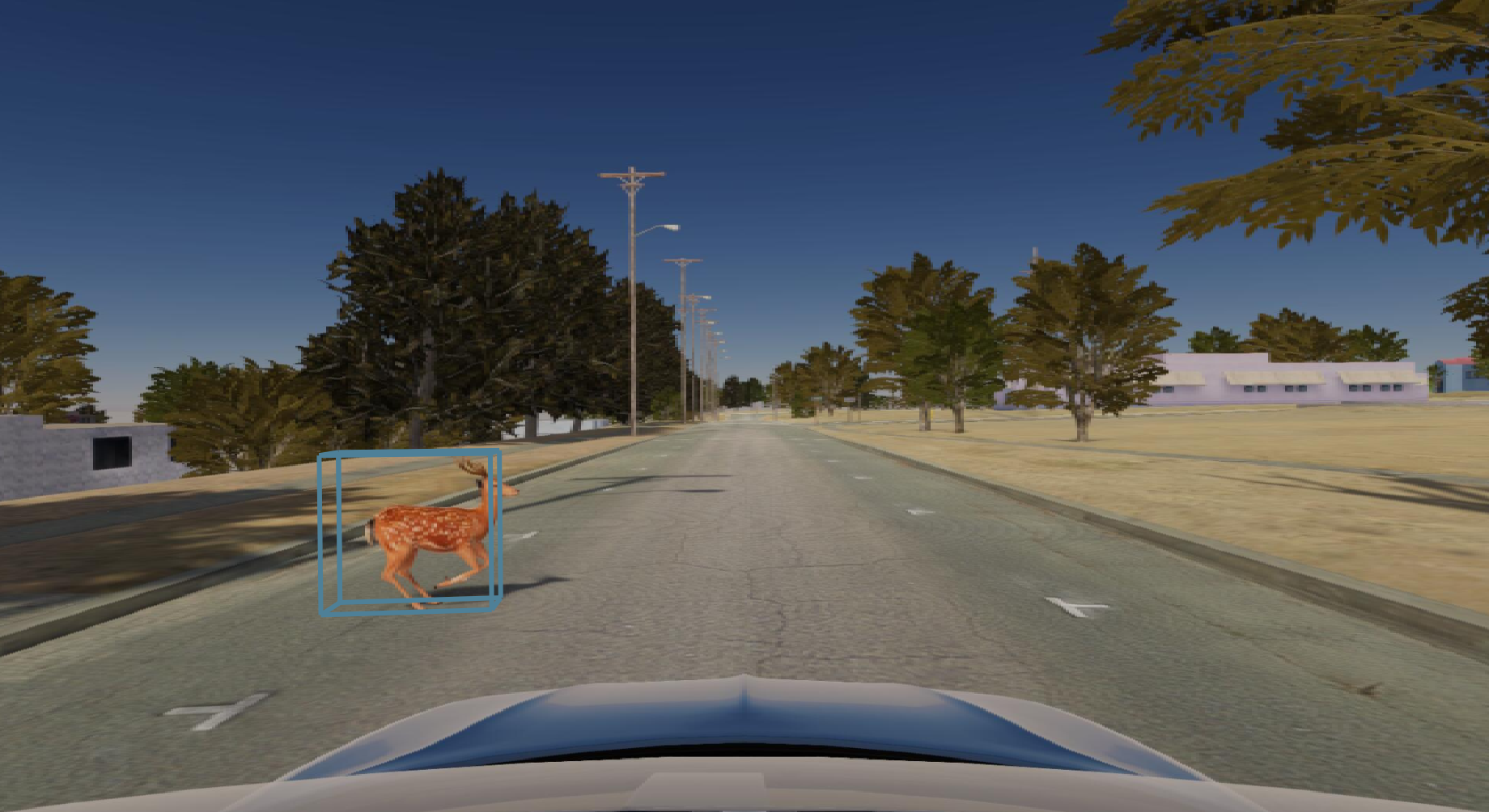}
  \caption{Camera Image for the scenario in~\cref{fig:deer_scenario}. Blue bounding box is the ground truth detection. The camera fails to detect the deer crossing the road (misdetection failure).}
  \label{fig:deer_camera}
\end{figure}

The scenario is novel to the identification algorithm, \ie not used for training, test, or validation.
The results of the failure identification are shown in~\cref{fig:deer_traj}, where we used the probabilistic 
fault identification.
%
Initially, the monitor detects no failure (rightmost green section).
As the ego vehicle gets closer to the undetected obstacle, the radar detects the obstacle but the camera does not. 
The inconsistency between the two sets of obstacles causes the test between camera and radar to return \fail.
Given the test's outcomes, the factor graph correctly detects and identifies the failure, triggering an alarm 
(rightmost red section).
As the ego vehicle gets even closer, the deer goes out of the field-of-view of the radar while entering the LiDAR field-of-view.
For a few meters, both camera and LiDAR fail to detect the deer \cref{fig:deer_camera}, but since it is out of the field-of-view of the radar, the \dtest fails to report the failure\footnote{This could be solved by improving the logic of the \dtest; for instance, it could predict that ---while the obstacle moved outside the field-of-view--- it is unlikely it disappeared.}.
As the obstacle re-enters the field-of-view of the radar, the \dtest again returns \fail, signaling the presence of a failure.

\begin{figure}[!tbp]
  \begin{center}
  \hspace*{-1cm}
  \subfloat 
  {
    \begin{tikzpicture}
      \node at (0,0) {\includegraphics[width=0.5\textwidth]{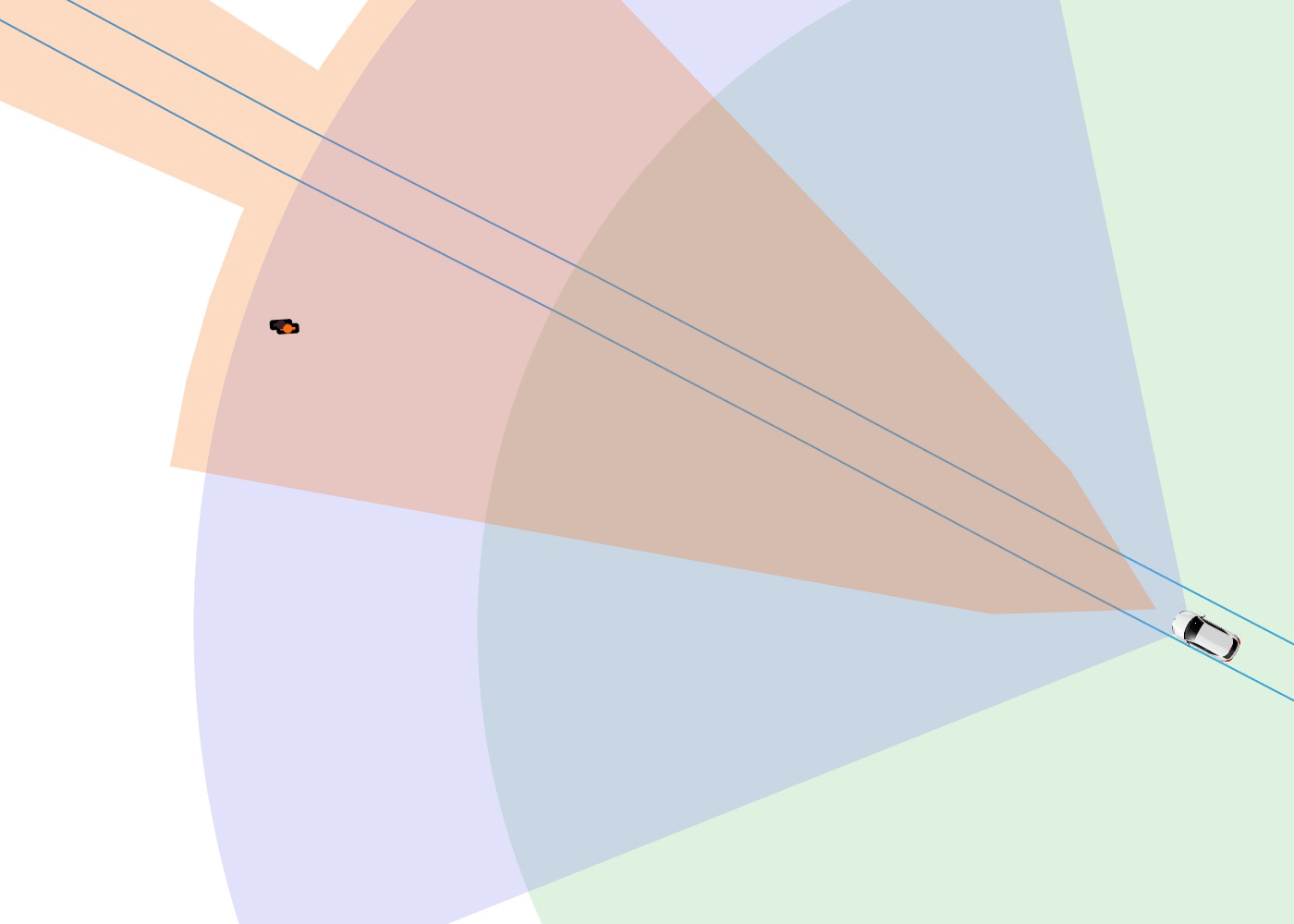}};
      \draw[color=red!60, very thick, text=black](-1.68,0.64) circle (0.3) +(0,-0.3) 
        node[anchor=north, text width=4cm, execute at begin node=\setlength{\baselineskip}{2pt}]{
          \scriptsize{The radar detects the obstacle, but the camera fails to do so}
        };
    \end{tikzpicture}
  }~
  \subfloat 
  {
    \begin{tikzpicture}
      \node at (0,0) {\includegraphics[width=0.5\textwidth]{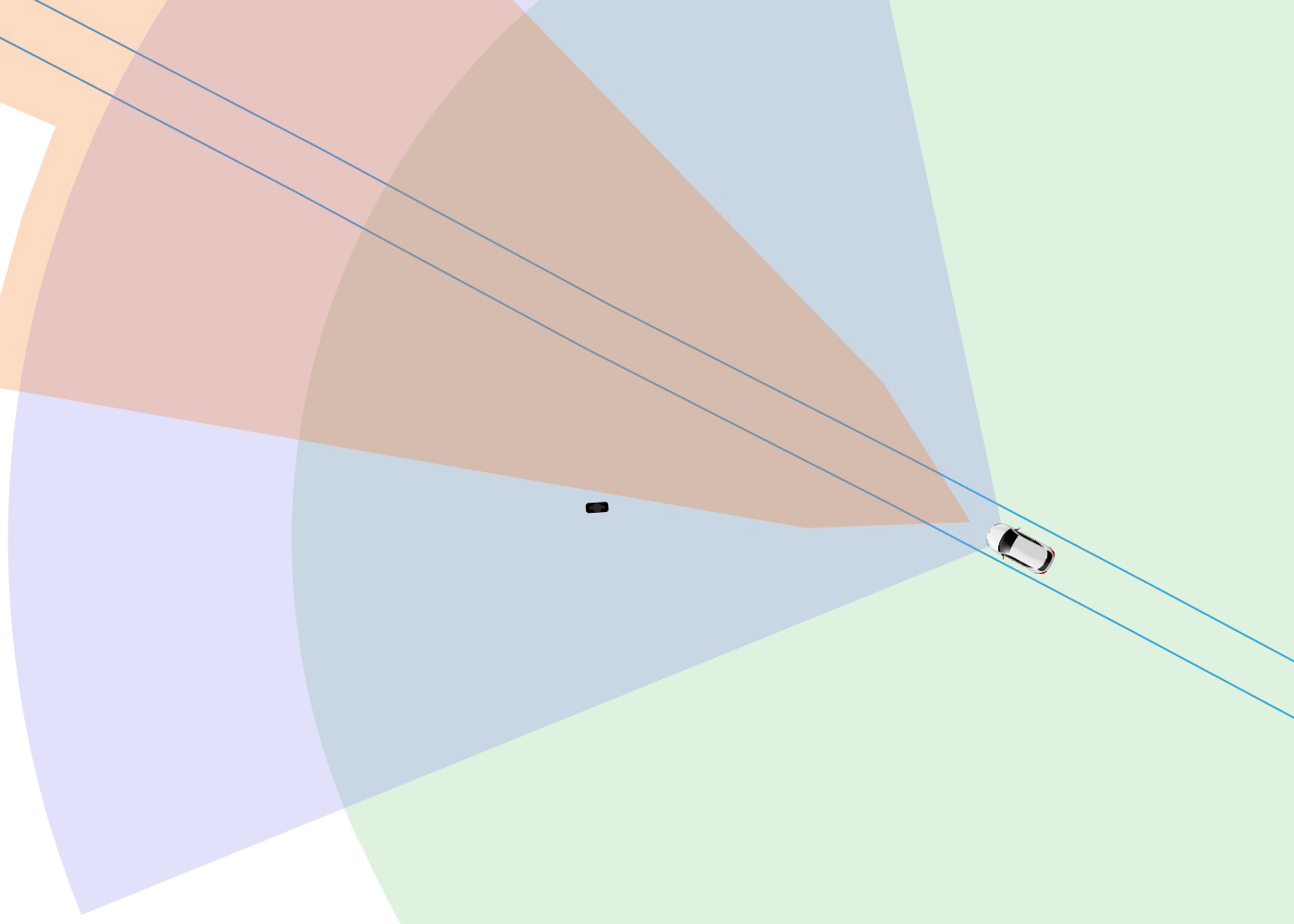}};
      \draw[color=red!60, very thick, text=black](-0.22,-0.2) circle (0.3) +(0,-0.3) 
        node[anchor=north, text width=4cm, execute at begin node=\setlength{\baselineskip}{2pt}]{
          \scriptsize{Camera and LiDAR fail to detect the obstacle while it is outside the radar field-of-view}
        };
    \end{tikzpicture}
  }
  \caption{
    Two snapshots from the example scenario of~\cref{fig:deer_scenario}.
    Shaded areas represent the sensor field-of-view (FOV):
    green, blue, and orange represent the LiDAR, camera, and radar FOVs, respectively. 
    On the left, the deer is outside the LiDAR FOV (so the LiDAR obstacle detector is not supposed to detect the obstacle); the radar detects the obstacle, while
    the camera fails to detect it even if it is inside its FOV. 
    Since the corresponding \dtest fails, our monitors can detect the failure.
    On the right, the deer is outside the radar FOV; in this case, both the camera and the LiDAR fail to  detect the obstacles (even though it is within their FOVs), hence no \dtest fails and our monitor fails to detect the fault. 
  \label{fig:cervoInvisibile}}
  \end{center}
\end{figure}

The first alarm is raised \SI{7.19}{\second} before the collision, flagging the camera misdetection as an active failure mode.
Before the collision, the AV has a speed of \SI{8.43}{\meter/\second}. 
{The car can reach a maximum deceleration of \SI{6}{\meter\per\second\squared}.}
As result, the car would need \SI{1.4}{\second} to come to a complete stop.
We note that after detecting the fault, for a short interval of time the monitor detects no failure: 
this is due to the fact that the deer goes out of the radar field-of-view, and no other obstacle detector is capable of detecting it, thus lacking redundancy to diagnose the failure; see the visualization and explanation in~\cref{fig:cervoInvisibile}.

To gather statistical evidence of the effectiveness of the fault detection, we run the same scenario 10 times at different times of the day (sun, twilight, and night) and different weather conditions (including fog and rain).
The probabilistic fault detection approach never raised false alarms in these tests, and the average time between the alarm and the collision was \SI{7.54}{\second}.
The car traveled at an average speed of \SI{6.16}{\meter/\second}, requiring \SI{1.03}{\second} to come to a complete stop.
The fault identification exhibited an average accuracy of \SI{93.75}{\percent}.

%% file: sections/conclusions.tex

\section{Conclusions}
This paper investigated runtime monitoring of complex perception systems and
 presented a novel framework to collect and organize diagnostic information for fault detection and identification in perception systems.
Toward this goal, we formalized the concept of \emph{\dtests}, a generalization of runtime monitors, that return diagnostic information about the presence of \fmodes. 
 We then introduced the concept of \emph{\dgraph}, as a structure to organize diagnostic information and its relations with the monitored perception system.
We then provided a set of deterministic, probabilistic, and learning-based algorithms that use \dgraphs to perform fault detection and identification.  
In addition to the algorithms, we investigated fundamental limits and provided deterministic and probabilistic guarantees 
on the fault detection and identification results. 
These include results about the maximum number of faults that can be correctly identified in a given perception system
as well as PAC-bounds on the number of mistakes our fault identification algorithms are expected to make.  
We conclude the paper with an extensive experimental evaluation, which recreates several realistic failure modes in 
the \lgsvl open-source autonomous driving simulator, and applies the proposed system monitors to a state-of-the-art autonomous driving software stack 
(Baidu's Apollo Auto). The results show that the proposed system monitors outperform baselines in terms of fault identification accuracy, have the potential of preventing accidents in realistic scenarios, and incur a negligible computational overhead.

This work opens a number of avenues for future work.
First, we plan to test our monitors on real-world datasets (rather than realistic simulations) and to provide more examples of the proposed approach applied to other perception subsystems (\eg localization, lane segmentation).
Second, we plan to add a risk metric to the fault identification process that could help the decision layer to make more informed decisions.
Finally, in this paper, we used simple \dtests.
Moving forward, it would be desirable to use more advanced \dtests available in the literature. 